\documentclass{article}

\usepackage{PRIMEarxiv}

\usepackage{newclude}
\usepackage[utf8]{inputenc} 
\usepackage[T1]{fontenc}    
\usepackage{hyperref}       
\usepackage{url}            
\usepackage{amsfonts}       
\usepackage{nicefrac}       
\usepackage{microtype}      
\usepackage{lipsum}
\usepackage{fancyhdr}       
\usepackage{graphicx}       
\usepackage{svg}

\usepackage{enumitem}  

\usepackage{longtable}
\usepackage{array, boldline, makecell}  
\usepackage{textcomp}
\usepackage{tabularx}
\usepackage{threeparttable}
\usepackage{multirow}
\usepackage{booktabs}
\usepackage{rotating}
\usepackage{colortbl}
\usepackage{arydshln}
\usepackage{caption}
\usepackage{subcaption}
\usepackage{adjustbox}
\usepackage{wasysym}
\usepackage{here}
\usepackage{wrapfig}

\newcommand*\nuh{-}
\newcommand*\tsa{$\dagger$}
\newcommand*\tsb{$\ddagger$}
\newcommand*\tpm{\,\textpm\,}

\newcommand*\MARK{\textbullet}
\newcommand*\CHECK{\textcolor{black}{\checked}}
\newcommand*\EMPTY{---}

\graphicspath{{media/}}     

\usepackage{pifont}
\usepackage{pgfplots}
\pgfplotsset{width=10cm, compat=1.18}
\usepackage{pgfplotstable}

\usepackage{xcolor}
\definecolor{colorclust}{HTML}{E84C3D}
\definecolor{colorloc}{HTML}{F08705}
\definecolor{colorseg}{HTML}{12AAB5}
\definecolor{colordec}{HTML}{1699D3}

\definecolor{colorseg1}{HTML}{33B7BD}
\definecolor{colorseg2}{HTML}{4DC7D1}
\definecolor{colorseg3}{HTML}{7DD6DF}
\definecolor{colorseg4}{HTML}{A2E4EC}
\definecolor{colorseg5}{HTML}{C5F1F7}
\definecolor{colorseg6}{HTML}{E1F8FC}

\definecolor{colordec1}{HTML}{80C4ED}
\definecolor{colordec2}{HTML}{C3E0FF}

\definecolor{softyellow}{HTML}{FFF2CC}
\definecolor{softorange}{HTML}{FFE6CC}
\definecolor{softred}{HTML}{F8CECC}
\definecolor{softblue}{HTML}{DAE8FC}
\definecolor{softgreen}{HTML}{D5E8D4}
\definecolor{softpurple}{HTML}{E1D5E7}
\definecolor{softgray}{HTML}{F5F5F5}
\definecolor{teal}{HTML}{12A5B0}
\definecolor{graydivider}{HTML}{CCCCCC}
\definecolor{rowbackground}{HTML}{F1F1F1}
\definecolor{citecolor}{HTML}{006eb8}
\definecolor{linkcolor}{HTML}{006eb8}

\hypersetup{
  colorlinks=true,        
  linkcolor=linkcolor,         
  citecolor=green,        
  urlcolor=purple,        
  linkbordercolor=red,    
  pdfborder={0 0 0},      
}

\usepackage{tikz}
\usetikzlibrary{positioning,shapes,shadows,arrows}

\title{Unsupervised Object Discovery:\\A Comprehensive Survey and Unified Taxonomy}

\author{
  Jos\'{e}-Fabian Villa-V\'{a}squez \\
  \'{E}TS Montr\'{e}al \\
  Montr\'{e}al, Canada\\
  \texttt{jose.villa-vasquez.1@ens.etsmtl.ca} \\
   \And
  Marco Pedersoli \\
  \'{E}TS Montr\'{e}al \\
  Montr\'{e}al, Canada\\
  \texttt{marco.pedersoli@etsmtl.ca} \\
}

\begin{document}
\maketitle

\begin{abstract}
Unsupervised object discovery is commonly interpreted as the task of localizing and/or categorizing objects in visual data without the need for labeled examples. While current object recognition methods have proven highly effective for practical applications, the ongoing demand for annotated data in real-world scenarios drives research into unsupervised approaches. Furthermore, existing literature in object discovery is both extensive and diverse, posing a significant challenge for researchers that aim to navigate and synthesize this knowledge. Motivated by the evidenced interest in this avenue of research, and the lack of comprehensive studies that could facilitate a holistic understanding of unsupervised object discovery, this survey conducts an in-depth exploration of the existing approaches and systematically categorizes this compendium based on the tasks addressed and the families of techniques employed. Additionally, we present an overview of common datasets and metrics, highlighting the challenges of comparing methods due to varying evaluation protocols. This work intends to provide practitioners with an insightful perspective on the domain, with the hope of inspiring new ideas and fostering a deeper understanding of object discovery approaches.
\end{abstract}


\section{Introduction}
The ability to discover objects in visually rich data such as images and videos, stands as a fundamental challenge in the field of computer vision \cite{greff2020binding}. Achieving this capacity constitutes a crucial step in allowing intelligent systems and machines to emulate a human-level understanding of the visual world. In humans, the capacity to extract meaningful entities from various modalities of sensory inputs is a paramount aspect of the process of perception, and is essential for engaging in higher-level cognitive tasks such as reasoning and planning \cite{greff2020binding}. Consequently, in order to build technologies that replicate these sophisticated abilities in machiness—--enabling them to perceive and interact with objects in a manner similar to many animals, it becomes imperative to develop methods and algorithms that can automatically localize and recognize objects. However, prior to machines acquiring the capacity to autonomously \textit{recognize} unfamiliar objects, a foundational step involves training them to \textit{discover} these entities of interest without reliance on human supervision. The importance of developing such algorithms lies in their vast potential to facilitate higher-level processing of visual information, thereby catalyzing the advancement of real-world practical AI applications across various sectors including robotics, autonomous vehicles, manufacturing, and beyond.

The majority of the current vision technologies, sufficiently advanced for object recognition in practical engineering and scientific applications, rely heavily on meticulously annotated datasets. However, these systems often encounter limitations when confronted with objects absent from their training data. Adding to the complexity is the considerable expense involved in acquiring ground-truth data for specific applications, which demands significant human resources. These challenges collectively hinder the ability of current systems to generalize effectively to data beyond their training sets. One strategy that addresses these challenges involves employing approaches to few-shot learning. These techniques utilize a limited number of examples to refine supervised models, aiming to broaden their capacity for generalization to \emph{unseen} data \cite{huangSurveySelfSupervisedFewShot2021}. Nonetheless, despite the advancements in few-shot learning, there remains a persistent reliance on annotated data and the assumption that models can only identify pre-defined categories. \footnote{This is known as the \textit{closed-set assumption} and implies that a model can only identify pre-defined categories that are present in the training set.} A more recent strategy seeks to localize and recognize object categories beyond the annotated label space, in open vocabulary settings. The rapid advancement of vision language pre-training models, such as CLIP \cite{Radford2021LearningTV}, coupled with the easier acquisition of image-text pairs, is positioning the open vocabulary approach as a more general and effective alternative compared to zero-shot learning and weakly supervised methods \cite{Wu2023TowardsOV}. Nevertheless, these strategies do not fully eliminate the requirement for annotated data. Even when the dependency on labeled data is mitigated, it continues to pose a significant practical challenge. This issue is further underscored by the human capability to effortlessly identify approximately 30,000 different object classes \cite{tuytelaarsUnsupervisedObjectDiscovery2010}, highlighting the substantial disparity that still remains between machine and human cognitive capabilities.

Over the years, this reality has spurred extensive investigation into methods that harness unlabeled data to address various well-known tasks, such as: image classification, object detection, co-localization and co-segmentation, saliency detection, semantic segmentation, and more recently, object-centric structured decomposition. This proliferation of research avenues has sparked numerous endeavors that seek to enhance the computational ability to discover objects within image collections; giving rise to a plethora of relevant studies that enrich the scientific literature. While these studies may vary significantly in their approaches to object discovery---from traditional clustering techniques \cite{tuytelaarsUnsupervisedObjectDiscovery2010} from the pre-deep learning era to more modern methods leveraging Vision Language Models (\textit{VLMs}) \cite{Gu2021OpenvocabularyOD, Zareian2020OpenVocabularyOD}, the overarching objective remains consistent: to localize and differentiate objects within images. To be more specific, methods may cluster images based on object classes, identify object locations \footnote{A further subdivision includes class-aware vs class-agnostic localization, single-object vs multiple objects settings, and instance-agnostic vs instance-aware mask-based localization, i.e. \emph{segmentation}.} in images using either bounding boxes or masks, or decompose images into meaningful entities—\textit{i.e.} objects—by inferring properties such as appearance, location, and relationships \footnote{Some methods seek to infer latent factors of variation that can be manipulated to synthesis novel images}. Moreover, a wide range of techniques has been employed in these pursuits, spanning from classical methods such as normalized cuts and spectral clustering to advanced deep learning approaches, including Variational Autoencoders (VAEs), Generative Adversarial Networks (GANs), and Transformers.

Consequently, to foster greater awareness and understanding of the multitude of methodologies employed in the object discovery literature, this survey seeks to provide a comprehensive study of these approaches. We propose a framework for categorizing the extensive literature on \emph{unsupervised object discovery} methods, based on the different end goals and the diversity of families of techniques utilized. As previously stated, numerous distinct research directions have addressed the challenge of object discovery over the years, and to the best of our knowledge, no existing survey thoroughly explores and unifies this range of perspectives on unsupervised object discovery. We start this survey, by examining the primary contributions of previous related surveys in \hyperref[sec:related]{Sec.~\ref{sec:related}}.  Next, in \hyperref[sec:perspectives]{Sec.~\ref{sec:perspectives}}, we introduce the framework for categorizing the broad spectrum of methods identified in the literature. Specifically, we organize these methods into categories that pertain to well-known tasks such as image clustering, bounding-box localization, object segmentation, and more recently object-centric scene decomposition. Then, in \hyperref[sec:considerations]{Sec.~\ref{sec:considerations}} we review several important considerations that emerge when addressing the unsupervised regime, and scrutinize the different strategies employed to overcome the absence of annotated data. Moving forward to \hyperref[sec:families]{Sec.~\ref{sec:families}}, we delve into the actual methodologies that have been adopted by various works to address the unsupervised object discovery problem. There, we examine the common methodological choices undertaken in the formulation of models and subsequently present a taxonomy of methods based on the families of techniques. In \hyperref[sec:experimental]{Sec.~\ref{sec:experimental}}, we examine common experimental configurations, including benchmark datasets and evaluation metrics. Finally, in \hyperref[sec:discussion]{Sec.~\ref{sec:discussion}} we discuss relevant and critical insights, and in \hyperref[sec:conclusion]{Sec.~\ref{sec:conclusion}} we present our concluding remarks.

\section{Related Previous Surveys}
\label{sec:related}

Among the existing surveys related to our work, one of the earliest studies on the topic of unsupervised object discovery introduced a framework for evaluating different methodologies. In this work, Tuytelaars \textit{et al.} \cite{tuytelaarsUnsupervisedObjectDiscovery2010}, conducted a comparative analysis of various baseline approaches, methods based on latent variable models, and spectral clustering techniques, all of which were designed to discover the objects present in the images in an unsupervised setting. Although the survey explicitly focused on unsupervised object discovery, it becomes evident that it was more accurately centered on unsupervised object \textit{category} discovery. At the time of its publication, there was no consensus on the precise definition of the unsupervised object discovery task, which is a challenge that continues to persist today. This ongoing ambiguity surrounding the definition of the task has served as a strong motivator for our study.

In a recent work, Siméoni \textit{et al.} \cite{Simoni2023UnsupervisedOL} reviewed methods with a focus on unsupervised object localization, emphasizing the role of Vision Transformers (ViTs) and self-supervised learning. The survey presents relevant tasks and metrics, and reviews common techniques employed by methods
that directly use self-supervised representations both without training or with further refining strategies. In fact, this focus on ViTs reflects a broader trend in the field, where self-supervised techniques are increasingly being integrated into object discovery pipelines---encompassing clustering approaches, box and mask localization methods, and object-centric learning frameworks, allowing for more robust and generalizable models. Like this, there exist several works that have explored diverse areas related to object discovery. 

Nonetheless, most of these surveys have been narrowed to particular methodologies and approaches to object discovery, and are not necessarily centered around the unsupervised regime. Among these, Wang \textit{et al.} \cite{Wang2019SalientOD} reviewed salient object detection, Schmarje \textit{et al.} \cite{Schmarje2020ASO} presented a survey on image classification, Shehzadi \textit{et al.} \cite{Shehzadi2023ObjectDW} review the use of transformers in object detection, Oza \textit{et al.} \cite{Oza2021UnsupervisedDA} presented a comprehensive comparison of unsupervised domain adaptive object detection methods, Sharma \textit{et al.} \cite{Sharma2022ASO} explored the instance segmentation literature based on reinforcement learning and transformers, Huang \textit{et al.} \cite{huangSurveySelfSupervisedFewShot2021} reviewed few-shot and self-supervised object detection methods, and Wu \textit{et al.} \cite{Wu2023TowardsOV} recently reviewed open vocabulary learning.

While these reviews provide valuable insights into specific methodologies, a comprehensive framework that integrates the diverse approaches across these areas has been missing. To our knowledge, our study is the first to address this gap by presenting a unified perspective on unsupervised object discovery, providing a broader and more comprehensive analysis of this research avenue within the field of visual scene understanding.

\section{Understanding Object Discovery: The different perspectives}
\label{sec:perspectives}

Throughout the years, numerous approaches have tackled the problem of object discovery, progressively aiming to reduce the strong dependency on carefully annotated data inherent in supervised learning methods. Some methods have focused on object discovery through localization tasks, accomplished either through pixel-wise masks or rectangular regions known as \textit{bounding-boxes}. Conversely, other approaches prioritize object discovery by determining the object's class. \footnote{We use the terms class and category interchangeably.}, regardless of its spatial location. Recently, a novel perspective on object discovery has emerged, aiming to localize and categorize objects by learning symbolic representations of them. Typically, these methods seek to enhance generalization to unseen data, with some capable of synthesizing novel data \cite{greff2020binding, locatello2020objectcentric}. Subsequent paragraphs succinctly describe these distinct methodological paradigms.

\begin{figure}
    \centering
    \includegraphics[width=0.99\textwidth]{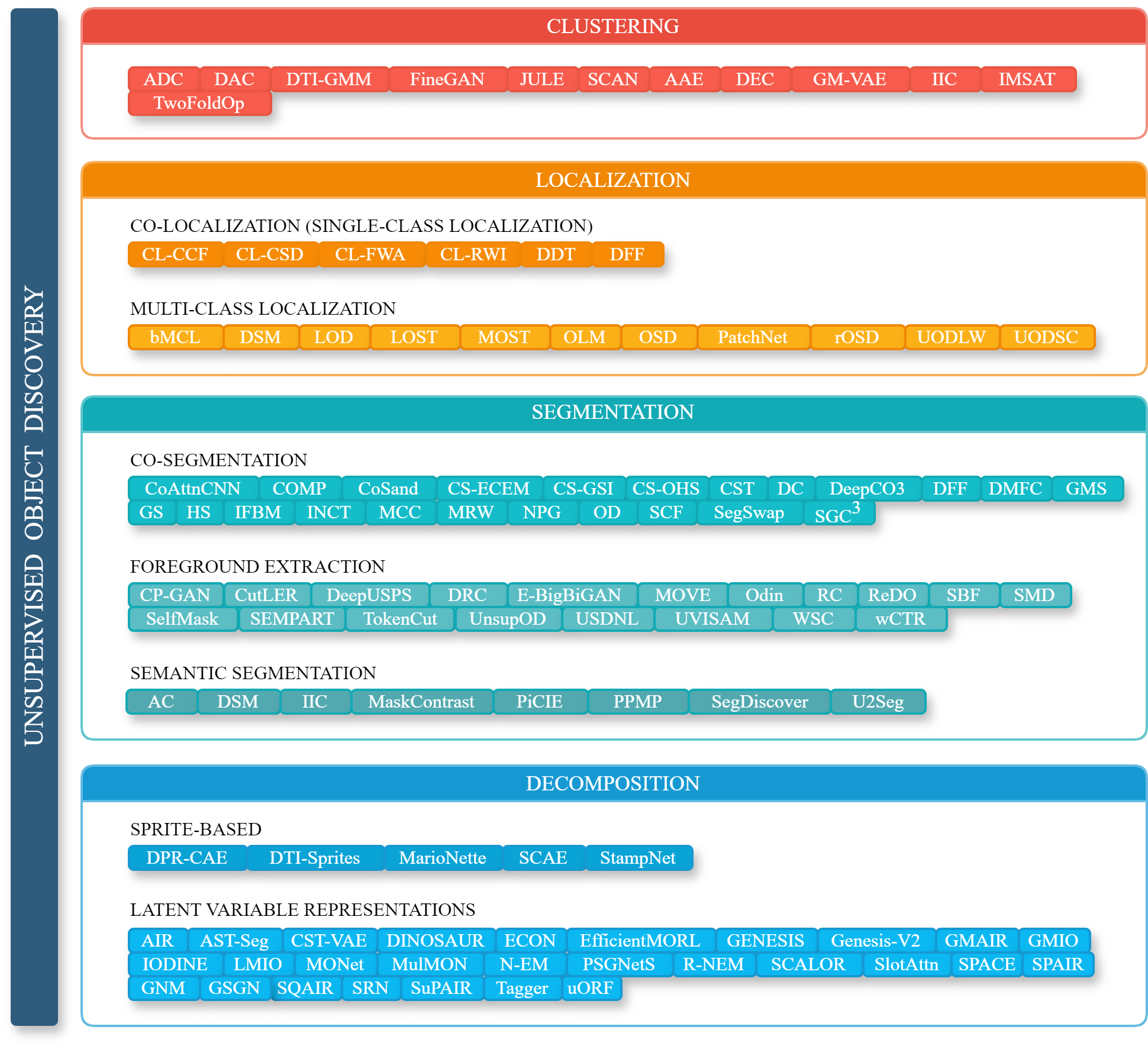}
    \captionsetup{width=0.99\textwidth}
    \caption{Taxonomy of unsupervised object discovery methods based on the common tasks tackled in the literature.}
    \label{fig:taxonomy_what}
\end{figure}

As illustrated in \hyperref[fig:taxonomy_what]{Fig.~\ref{fig:taxonomy_what}}, the broad spectrum of methods addressing the object discovery problem in various ways can be systematically categorized into four primary groups: clustering, localization, segmentation and decomposition. Each of these groups encompasses methodologies that align well to these four general approaches to the discovery task. For instance, clustering methods aim to categorize images based on the object class depicted in each. Localization methods entail the generation of bounding boxes around individual objects, whereas  segmentation methods aim to produce masks that delineate objects at the pixel level. Decomposition methods, on the other hand, target objects as integral components of a visual scene, considering the background region as a scene component as well. Furthermore, localization, segmentation, and decomposition methods may also strive to classify objects based on their respective categories. \newline

\textbf{\emph{Clustering}} methods aim to group images based on the object category present in the image, regardless of the spatial location of the objects. Consequently, these methods often presuppose the presence of a single object per image, with each resulting cluster representing a distinct object category. However, there is no inherent limitation preventing images from being associated with multiple classes. In such scenarios, the clustering algorithm must utilize local image representations rather than global ones. Furthermore, it is important to note that some methods within this category are explicitly designed for image clustering, while others are general data clustering algorithms or models that have been adapted for this purpose \cite{makhzani2016adversarial}. \newline

\textbf{\emph{Localization}} methods aim to generate bounding-boxes that accurately encompass objects within an image. Typically, these methods are instance-aware, treating each bounding-box as an individual object instance regardless of its specific class. While most of these approaches are class-agnostic, prioritizing the geometric positioning of objects, some also incorporate classification capabilities to discover object classes. Additionally, a notable distinction among localization methods lies in their experimental setups. Some are run separately for each object class, with the independent outcomes subsequently averaged to yield dataset-wide results. Conversely, others operate on the entire image collection simultaneously, with the results reported accordingly. Lastly, certain localization methods possess the capacity to detect multiple objects within a single image. \newline

\textbf{\emph{Segmentation}} methods also aim to discover the location of objects within an image. However, instead of employing bounding-boxes, they generate masks to pinpoint objects at the pixel level. Some of these methods assume the presence of a predominant object in each image, utilizing saliency cues and foreground extraction techniques to unveil it. Conversely, other approaches do not presuppose a single object but still rely on prominence to infer that the foreground region of an image corresponds to either a single object or multiple objects. Another avenue within this category of methods involves semantically segmenting the entire image into spatial regions corresponding to objects. As with localization methods, a key distinction among segmentation methods lies in their experimental setup. One particular line of research focuses on applying segmentation techniques to subsets of images known to contain objects of the same category. In addition, segmentation methods that do not distinguish objects of the same class with distinct masks are instance-agnostic, irrespective of their class-awareness capability. \newline

\textbf{\emph{Decomposition}} methods seek to learn object-centric representations that ideally represent individual objects within an image. These representations can then be subsequently decoded into separate components in pixel space. The prevailing approach in this category of methods involves constructing latent representations through Bayesian learning and utilizing an encoder-decoder framework. Another related line of work, seeks to find object-centric representations by learning prototypes of objects alongside transformation parameters that allow to reconstruct individual objects and, ultimately, the entire scene. Methods in this category are inherently instance-aware as they strive to isolate each individual object. However, not all methods withing this class of methods aim to discover the category of objects. \newline

In the following subsections, we provide detailed explanations of the various types of methods corresponding to each of the four categories outlined in the taxonomy presented in the above figure. We reaffirm that, within this framework, these methods indeed perform a form of object discovery that aligns with more established tasks, as illustrated in \hyperref[fig:task_ilustration]{Fig.~\ref{fig:task_ilustration}}. \newline  


\textcolor{colorclust}{\subsection{Discovery by Clustering}}

Clustering methods aim to discover groups of data points within unstructured datasets, with the objective of delineating clusters that emulate class structures. These algorithms need to learn to establish clusters and, at the same time, to decide the membership of the data points among the modeled clusters. Therefore, these methods seek to find similarities among data points, assigning similar data points to specific clusters, while segregating dissimilar data points into separate clusters. As such, the concept of similarity is prominent in clustering algorithms and is usually defined according to the specific applications of interest. In this regard, a common interpretation of unsupervised clustering draws inspiration from the task of unsupervised discrete representation learning and considers this task as pertaining to the problem of latent variable disentanglement, in which an algorithm aims to find structure in latent space, by learning a mapping function without supervision \cite{dilokthanakul2017deep}.

There exist numerous domains of applications in which clustering can be applied. Accordingly, a diversity of clustering methods have been proposed over the years, which explore different approaches to modeling clusters and determining cluster membership. A review of such approaches is presented in \hyperref[sec:families]{Sec.~\ref{sec:families}}. Within this category of methods that perform clustering, a first group of methods \cite{makhzani2016adversarial,xie2016unsupervised,dilokthanakul2017deep,ji2019invariant,hu2017learning,tissera2022neural} has been designed---or in certain cases adapted \cite{makhzani2016adversarial}---to perform data clustering without being restricted to a specific data type. In contrast, another group emerges from methods that have been specifically developed  for clustering images. These methods draw inspiration from the human visual ability to discern similarities among images in a collection, which enables them to identify which images should be grouped together. This process, of course, implies an abstract understanding of the content of images. As such, these methods consider image clustering as an image classification task, which seeks to assign a semantic label from a predefined set of classes to an image \cite{vangansbeke2020scan}. Some methods that fall in this group are \cite{haeusser2019associative,chang2017deep,monnier2020deep,singh2019finegan,yang2016joint,vangansbeke2020scan}.


Consequently, although some methods are formulated as data clustering algorithms while others are specifically tailored for clustering image, their ultimate objective remains consistent: class discovery. This shared goal defines the task on which these methods are evaluated. Thus, in this study, these types of methods that correspond to a form of unsupervised object discovery are categorized as belonging to the group: \textit{Clustering}.

\textcolor{colorloc}{\subsection{Discovery by Box-Localization}}

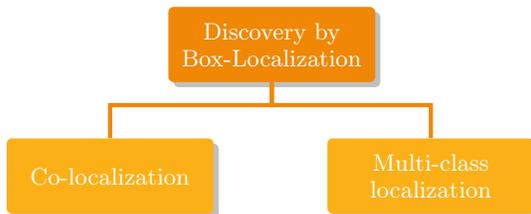
\begin{wrapfigure}{r}{0.5\textwidth}
    \definecolor{colorloc_co}{HTML}{F78B05}
\definecolor{colorloc_mul}{HTML}{FCB017}
\begin{center}
\begin{tikzpicture}[
    node/.style={rectangle, draw=none, rounded corners=2pt, align=center, fill=#1, drop shadow, text=white, minimum width=2.75cm, minimum height=1cm},
    line/.style={-, ultra thick, color=colorloc},
    node distance=1.5cm
]
    \node[node=colorloc] (Root) at (1.9,0) {Discovery by \\ Box-Localization};
    
    \node[node=colorloc_mul] (Leaf1) at (-0.25,-1.75) {Co-localization};
    \node[node=colorloc_mul, right=of Leaf1] (Leaf2)  {Multi-class \\ localization};
    
    \draw[line] (Root.south) -- ++(0,-0.3) -| (Leaf1.north);
    \draw[line] (Root.south) -- ++(0,-0.3) -| (Leaf2.north);
\end{tikzpicture}
\end{center}
    \caption{Localization tasks in UOD}
    \label{fig:chart_task_loc}
\end{wrapfigure}
\vspace{-\abovecaptionskip}

A second direction of research in object discovery encompasses methods that focus on localizing objects in images through the use of rectangular regions or bounding-boxes. While all these methods aim to output bounding-boxes to localize regions in images, there are nuanced distinctions that must be acknowledged to gain a deeper understanding of their varied approaches to solving object discovery. In \hyperref[fig:chart_task_loc]{Fig.~\ref{fig:chart_task_loc}} a further subdivision is observed.

The primary factor distinguishing localization methods is found in their experimental settings. Within this context, there exists a group of methods designed for single-class settings. In such scenarios, each of the distinct images in the training dataset contain a single object of the same category, i.e., the common object.  Even though the datasets used by these methods must contain the common object, a few are designed to deal with noisy images that do not belong to the common class. In the literature, this particular setting is generally called \textit{Co-Localization} and shares a similar goal with two other popular problems: Co-Segmentation, which produces segmentation masks rather than bounding boxes, and Weakly Supervised Localization, which additionally requires negative images that do not contain the common object, as referred by \cite{joulin2014efficient}. In this study, this group includes \cite{le2017colocalization, liCLCSD2016image, joulin2014efficient, tang2014colocalization, wei2017unsupervised, collins2018deep}.

Another group of methods are designed to work in a mixed-class experimental setting, and therefore, they do not require a single-class dataset. This type of methods is generally referred to as \textit{Multi-class Discovery} or simply \textit{Object Discovery}. A subtle further distinction among these methods arises when considering that certain methods assume the presence of a single object per image, while others are able to discover multiple object instances. Moreover, within mixed-class methods, some are capable of categorization, while others are not. It is important to note, however, that these nuanced differences are not unique to localization methods, as will be discussed in the following subsections. Among the mixed-class localization methods are: \cite{zhu2014bMCL, melas-kyriazi2022deep, vo2021large, simeoni2021localizing, zhang2020object, vo2019unsupervised, moon2021patchnet, vo2020toward, cho2015unsupervised, murasaki2019paper}.  \newline 

Furthermore, our investigation reveals that the term \textit{Object Discovery} is also utilized in various other lines of research where the explicit goal is to discover objects, and thus, its usage is not limited to the context of bounding box localization methods. This study acknowledges and leverages this circumstance by adopting the term \textbf{Object Discovery} as an umbrella term to denote a general task that is specialized into four main groups of methods, as presented in this section. Consequently, the taxonomies outlined in this study encapsulate this diversity of perspectives, suggesting that the remaining tasks analyzed within the other perspectives or lines of research, also contribute to the formation of groups and subdivisions within our classification scheme. \newline \newline \newline

\captionsetup[table]{skip=2em}
\begin{figure}[!ht]
    \centering
    \begin{tabular}{cc}
        \begin{subfigure}[b]{0.45\textwidth}
            \centering
            \includegraphics[width=\textwidth]{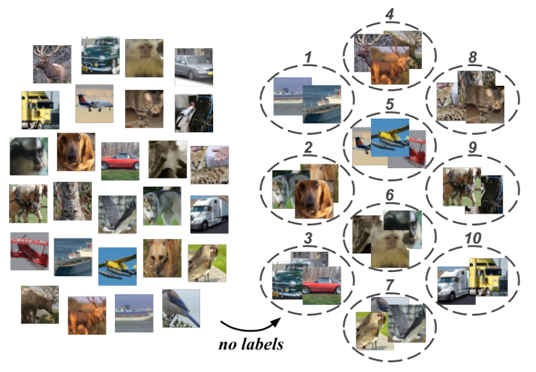}
            \caption{\textcolor{colorclust}{\MakeUppercase{Clustering}}: Images grouped according to the object class. No localization.}
        \end{subfigure} &
        \begin{subfigure}[b]{0.45\textwidth}
            \centering
            \includegraphics[width=\textwidth]{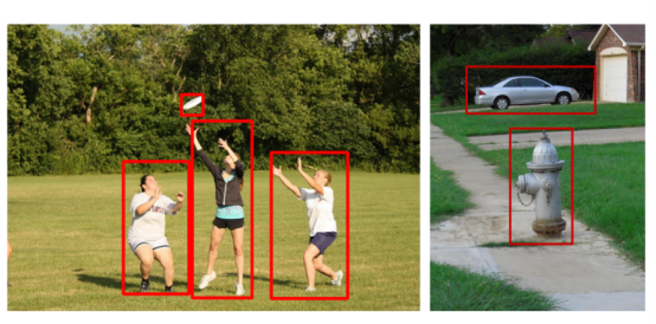}
            \caption{\textcolor{colorloc}{\MakeUppercase{Localization}}: Objects localized with bounding-boxes. Methods can seek to localize single or multiple objects per image. A further division separates methods that perform localization in multi-class settings.}
        \end{subfigure} \\
        
        \begin{subfigure}[b]{0.45\textwidth}
            \centering
            \includegraphics[width=\textwidth]{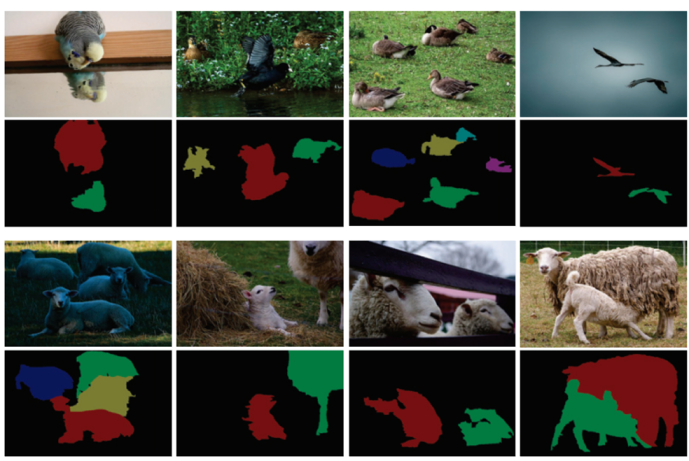}
            \caption{\textcolor{colorseg}{\MakeUppercase{Segmentation}}: Objects localized with pixel-level masks. Methods can seek to segment single objects, multiple objects per image or the entire image (semantic segmentation). Methods can be class-agnostic, class-aware and instance-aware.}
        \end{subfigure} &
        \begin{subfigure}[b]{0.45\textwidth}
            \centering
            \includegraphics[width=\textwidth]{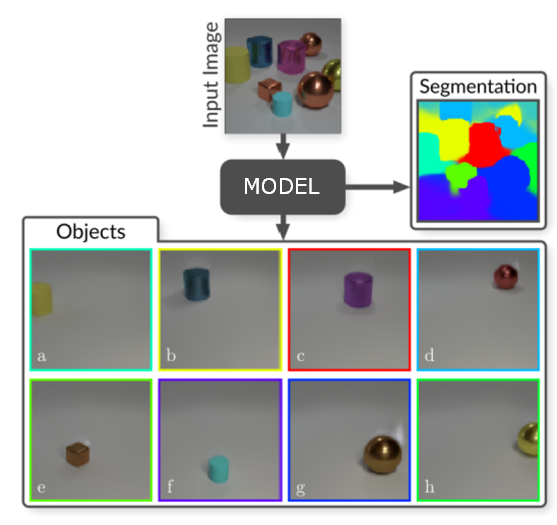}
            \caption{\textcolor{colordec}{\MakeUppercase{Decomposition}}: Objects are localized similar to Segmentation methods. However, these methods seek to learn a structured latent representation of each object which serve to generate image components containing the discovered objects.}
        \end{subfigure} \\
    \end{tabular}

    \captionsetup{width=0.92\textwidth}
    \caption{Various perspectives to the task of unsupervised object discovery. The illustrations are adapted from: a) Invariant Information Clustering for Unsupervised Image Classification and Segmentation IIC \cite{ji2019invariant}, b) Localizing Objects with Self-Supervised Transformers and No Labels LOST \cite{simeoni2021localizing}, c) Deep Instance Co-Segmentation by Co-Peak Search and Co-Saliency Detection DeepCO3 \cite{hsu2019deepco3}, d) Multi-object representation learning with iterative variational inference IODINE \cite{greff2019multi}
    }
\end{figure}
\label{fig:task_ilustration}

\vspace{2cm}

\textcolor{colorseg}{\subsection{Discovery by Segmentation}}

\begin{wrapfigure}{r}{0.55\textwidth}
    \definecolor{colorseg_co}{HTML}{14BDC9}
\definecolor{colorseg_for}{HTML}{5CBDC4}
\definecolor{colorseg_sem}{HTML}{53A9B0}
\begin{center}
\begin{tikzpicture}[
    node/.style={rectangle, draw=none, rounded corners=2pt, align=center, fill=#1, drop shadow, text=white, minimum width=2.75cm, minimum height=1cm},
    line/.style={-, ultra thick, color=colorseg},
    node distance=0.25cm
]
    \node[node=colorseg] (Root) at (0.01,0) {Discovery by \\ Segmentation};
    
    \node[node=colorseg_for] (Leaf1) at (-3,-1.75) {Co-segmentation};
    \node[node=colorseg_for, right=of Leaf1] (Leaf2)  {Foreground \\ extraction};
    \node[node=colorseg_for, right=of Leaf2] (Leaf3) {Semantic \\ segmentation};
    
    \draw[line] (Root.south) -- ++(0,-0.3) -| (Leaf1.north);
    \draw[line] (Root.south) -- ++(0,-0.3) -| (Leaf2.north);
    \draw[line] (Root.south) -- ++(0,-0.3) -| (Leaf3.north);
\end{tikzpicture}
\end{center}
    \caption{Segmentation tasks in UOD}
    \label{fig:chart_task_seg}
\end{wrapfigure}
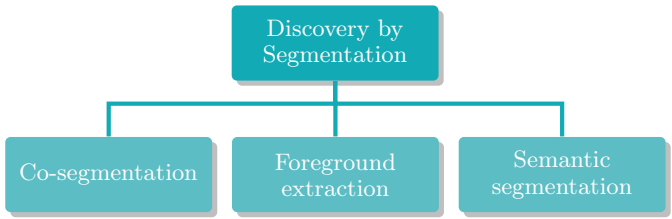
\vspace{-\abovecaptionskip}

In the third area of research for unsupervised object discovery, the focus is on finding the location of objects by segmenting out parts of an image at pixel level. As observed in \hyperref[fig:chart_task_seg]{Fig.~\ref{fig:chart_task_seg}}, this task has been traditionally tackled in three broad ways: co-segmentation, foreground extraction, and semantic segmentation. All these tasks share the ultimate goal of localizing objects via pixel-wise assignments. Nonetheless, there are subtle but important differences that must be taken into consideration when framing the object discovery problem in one of these ways. We will describe such differences in the rest of this section. It should also be noted that the terminology used to refer to these tasks is not exclusive to unsupervised approaches. In fact, with the exception of co-segmentation, which is sometimes referred to as a weakly supervised task, both semantic segmentation and foreground extraction are well-studied tasks in the supervised learning literature. Throughout this survey, we focus on these tasks only from the point of view of unsupervised learning. 

The first way in which the object discovery problem has been tackled from a segmentation standpoint is through the task of co-segmentation, which emerged as a natural extension of another task known as co-localization (OD). Co-segmentation methods seek to identify subsets of pixels in images that share semantic meaning across an image collection. Naturally, these shared semantics must correspond to the common object being targeted, and of course, this requirement implies that all images must contain the same and only one object class. To distinguish this task from the weakly supervised regime, co-segmentation methods do not require negative examples and typically handle noisy or distracting images that do not contain the common object. Even so, we note that it is still possible to argue that co-segmentation requires a subtle level of supervision because most of the images in the single-class dataset must contain at least one instance of the common object. Nonetheless, considering that no ground-truth location information and no image level labels are required to distinguish positive from negative examples, co-segmentation can be deemed an unsupervised learning task. 

A further subdivision among co-segmentation methods can be devised when considering the number of objects per image that they are able to discover. There are methods such as \cite{dai2013cosegmentation, shoitan2021unsupervised, joulin2010discriminative, collins2018deep, jerripothula2014automatic, weigeodesic, yan2013hierarchical, rubio2012unsupervised, rubinstein2013unsupervised, shen2022learning, tao2017image} that assume the presence of only one instance of the common object per image and are thus only capable of performing single-object segmentation. Other methods \cite{hsu2018coattention, faktor2013cosegmentation, gunheekim2011distributed, li2018unsupervised, li2016unsupervised, joulin2012multiclass, lee2015multiple, wang2017multiple, jerripothula2016image} can effectively extract separate masks for each non-overlapping object instance, while others \cite{hsu2019deepco3} have attempted to discover individual object instances. On the other hand, some works \cite{hongliangli2014unsupervised, chang2015optimizing} have attempted to combine co-segmentation and categorization in order to discover multiple objects in datasets that contain multiple common objects from different classes. 

A second group comprises methods that have attempted object discovery without supervision by framing the problem as foreground extraction. These methods make the problem of discovery somewhat easier by assuming that every pixel in an image belongs to either the foreground or the background. It is interesting to note that this assumption essentially turns the problem of semantic segmentation into a binary one. Furthermore, methods that attempt foreground extraction are not necessarily restricted to discovering only one object per image, like in \cite{voynov2021object, chengglobal, chen2019unsupervised, shin2022unsupervised, wang2022self, zhang2018deep, li2015weighted}. Indeed, there exist plenty of methods that perform multi-object discovery through foreground extraction \cite{arandjelovic2019object, nguyen2021deepusps, yu2021DRC, henaff2022object, zhang2017supervision, peng2017salient, zhao2020unsupervised, choudhury2022guess, zhu2014saliency}. It is worth noting that many of the methods that fall into this second group usually refer to this task as \textit{unsupervised saliency detection} and therefore are evaluated in benchmarks traditionally used for saliency detection (DUTS, ECSSD, and DUT-OMRON). Another consideration to keep in mind is that none of these approaches attempted to perform categorization of the inferred objects.

\textcolor{colordec}{\subsection{Discovery by Scene Decomposition}}

The fourth area of research for unsupervised object discovery involves methods whose main objective is to leverage unlabeled images that learn latent representations of the individual objects in the images. The basic idea consists of framing an image as a composition of individual components that can be learned individually. Furthermore, these object-centric representations can be designed to infer different object properties such as location, scale, and appearance, which can be used to build a reconstruction of the original image by combining the different inferred components. This perspective on the problem of object discovery enables the extraction of information from high-level features of the image data and represents them in a compact form that can be used to characterise the inferred objects. Within the context of unsupervised object discovery, we refer to these kinds of methods as object-centric decomposition methods.

\begin{wrapfigure}{r}{0.5\textwidth}
    \definecolor{colordec_sprite}{HTML}{0AA3D6}
\definecolor{colordec_latvar}{HTML}{0BB8F2}
\begin{center}
\begin{tikzpicture}[
    node/.style={rectangle, draw=none, rounded corners=2pt, align=center, fill=#1, drop shadow, text=white, minimum width=2.75cm, minimum height=1cm},
    line/.style={-, ultra thick, color=colordec},
    node distance=1.5cm
]
    \node[node=colordec] (Root) at (1.7,0) {Discovery by \\ Decomposition};
    
    \node[node=colordec_latvar] (Leaf1) at (-0.5,-1.75) {Sprite-based};
    \node[node=colordec_latvar, right=of Leaf1] (Leaf2)  {Latent-variable \\ representations};
    
    \draw[line] (Root.south) -- ++(0,-0.3) -| (Leaf1.north);
    \draw[line] (Root.south) -- ++(0,-0.3) -| (Leaf2.north);
\end{tikzpicture}
\end{center}
    \caption{Scene Decomposition tasks in UOD}
    \label{fig:chart_task_dec}
\end{wrapfigure}
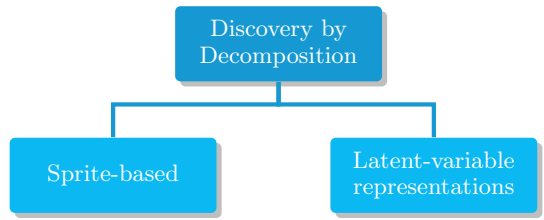


In object-centric decomposition methods, object localization can be performed via bounding-boxes \cite{eslami2016attend, lin2020space} or via pixel-wise mask assignments \cite{locatello2020objectcentric, engelcke2022genesisv2}. As observed in \hyperref[fig:chart_task_dec]{Fig.~\ref{fig:chart_task_dec}}, researchers have currently framed this problem in essentially two distinct ways. In the following paragraphs, we will describe these different ways in more detail. It is worth noting that this area of research is a relatively recent development, and therefore, precise terminology is yet to be agreed upon and more standard benchmarks are yet to emerge. In this work, we have adopted naming protocols and terminology that take into consideration the vast diversity of ways in which different authors have expressed their ideas while tackling this task.

The first category of methods in which the object discovery problem has been tackled from an object-centric decomposition perspective is through \textit{structured latent representation learning}. In these methods, object properties are captured by routing information from the images to the latent variables defined for this purpose. The backbone design element in the approaches formulated in this category is composed by encoder-decoder networks that are jointly trained. For example, a typical architectural setup can be composed of a convolutional network along with a variational auto encoder \cite{greff2019multi} that is trained to capture the shape, appearance, and spatial context of objects in an image. There is a further division among these methods based on how they approach localization. The first group uses spatial attention, implemented using bounding boxes, and includes the following methods: \cite{eslami2016attend, zhu2021gmair, kosiorek2018sequential, crawford2019spatially, lin2020space, jiang2020scalor, deng2021generative, jiang2021generative, stelznerfaster}. This group relies on localization mechanisms that identify interesting regions or "glimpses" in order to extract object information. These mechanisms are usually implemented via spatial transformer networks \cite{jaderberg2016spatial}. The second group of methods has captured much more attention in recent years and includes a broader collection of works, including \cite{emami2021efficient, engelcke2020genesis, engelcke2022genesisv2, greff2019multi, burgess2019monet, greff2016tagger, greff2017neural, locatello2020objectcentric, huang2016efficient, yuangenerative, yang2020learning, yu2021DRC, vonkugelgen2020causal}. These methods are capable of discovering multiple objects per image and are even able to distinguish object instances by inferring non-rectangular regions of an image. Given an image containing objects in different locations---including overlapping objects---, these approaches design the localization mechanisms such that each object is captured as a separate component. As this is clearly a difficult task, most of these object-centric representation methods have been limited to small datasets containing synthetic images with easily distinguishable objects rendered against non-cluttered backgrounds. However, more recently the use of transformer features---based on self-supervised learning strategies such as DINO \cite{caron2021emerging}---has allowed methods \cite{seitzer2022bridging} to tackle more difficult datasets containing natural images such as PASCAL VOC and MS COCO.

The second category of methods, \textit{prototype-based}, leverages the idea of learning object prototypes. In these methods, the aim is to discover objects in images by decomposing them into transformed instances of recurring visual elements that are inferred from an image collection. These recurrent elements are usually referred to as prototypes or \textit{sprites}. These prototypes form a dictionary learned as a collection of latent variables that represent prototypical objects in the dataset, and this dictionary is used to model images as a combination of transformed prototypes. The transformations learned are differentiable and typically include spatial and color transformations. Similar to the first category of decomposition methods, there are prototype-based methods that localize objects via bounding-boxes, such as \cite{visser2019stampnet}, as well as methods that localize objects at the pixel level, such as \cite{smirnov2021marionette, monnier2021unsupervised, xiang2021dprcae, kosiorek2019stacked}.

\vspace{15pt}
\section{Key Considerations in Object Discovery}
\label{sec:considerations}
Unsupervised learning involves uncovering interesting structures within data by identifying patterns and relationships without relying on associated labels or category information. As unsupervised methods must operate devoid of human supervision,  the patterns they generate may lack semantic interpretation. Hence, discovering meaningful entities within the data first requires understanding a notion of meaningfulness. These entities, commonly referred to as objects, serve to encapsulate this notion.

Over the years, the computer vision research community has put considerable effort into building methods capable of recognizing objects in images. In general, the task known as object recognition seeks to identify objects present in images and has been primarily explored in the supervised learning regime. Here, objects can be identified in various ways: by detecting their presence in an image, localizing instances of objects, or even localizing instances of different categories of objects. The latter is usually known as object detection. In the supervised learning regime, it is easier to understand the term object recognition because the information of what should be considered an object or how objects from particular categories look like is already given. However, what does it mean to recognize objects in the unsupervised learning regime, where no information is given about their appearance, shape, or location? Not surprisingly, a diversity of terms have emerged in the literature to refer to such a task. In this survey, we refer to this task as unsupervised object discovery, which involves localizing and/or categorizing unknown objects in visual data.

\pgfplotstableread{ 
Label     Clust Loc Seg Dec
2010    0 0 1 0
2011    0 0 1 0
2012    0 0 2 0
2013    0 0 4 0
2014    0 3 3 0
2015    1 1 3 0
2016    2 1 2 3
2017    3 1 3 1
2018    1 1 4 2
2019    2 1 5 6
2020    2 2 1 9
2021    0 3 5 6
2022    1 1 7 1
2023    0 5 4 4
    }\testdata

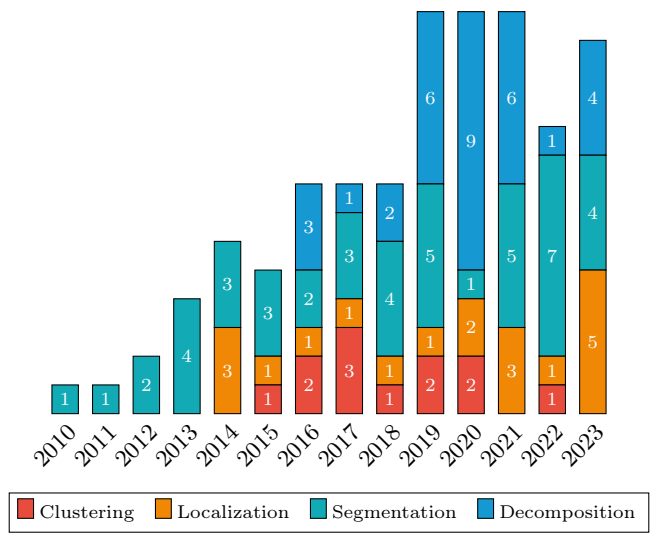
\begin{figure}
\centering
\begin{tikzpicture}

\begin{axis}[
ybar stacked, 
nodes near coords,
nodes near coords style={font=\scriptsize},
hide y axis,
x axis line style={opacity=0}, 
xtick style={draw=none}, 
height=8cm,
xtick=data,     
legend style={at={(0.5,-0.08)}, anchor=north,legend columns=-1, font=\scriptsize, 
/tikz/every even column/.append style={column sep=0.2cm}},
yticklabels={},
xticklabels from table={\testdata}{Label},  
xticklabel style={rotate=45, anchor=east, xshift=12pt, yshift=5pt},
]

\addplot [fill=colorclust, text=white] table [y=Clust, x expr=\coordindex] {\testdata};
\addplot [fill=colorloc, text=white] table [y=Loc, x expr=\coordindex] {\testdata};
\addplot [fill=colorseg, text=white] table [y=Seg, x expr=\coordindex] {\testdata}; 
\addplot [fill=colordec, text=white] table [y=Dec, x expr=\coordindex] {\testdata}; 

\legend{Clustering,Localization,Segmentation, Decomposition}

\end{axis}
\end{tikzpicture}

\caption{\textbf{Evolution of the quantity of papers} (considered in this survey) on each of the four groups of methods that tackle unsupervised object discovery. Papers included range from 2010 to 2023.}
\label{fig:chrono}
\end{figure}

\subsection{Evolution of approaches to object discovery}
As presented earlier in this survey, different interpretations of the task of discovering objects in images have emerged in the research literature over the years. These interpretations range from classifying images into categories to recognizing individual objects. Presumably \cite{tuytelaarsUnsupervisedObjectDiscovery2010}, the work of \cite{weber2000} was the first to explore the unsupervised discovery of object categories, and it employed a probabilistic formulation to represent objects as constellations of features and used expectation maximization for learning.

This idea of discovering object categories by separating images from a dataset into groups that correspond to different classes has been the main driver of unsupervised image clustering methods. These methods learn global representations of images, which are then used to predict cluster assignments. Some of these methods are specifically designed for images. In contrast, others have been proposed as general clustering methods which do not require images but can be adapted to perform clustering of images. Evidently, these methods do not perform localization.
 
On the other hand, an entirely different line of methods has focused on learning to localize objects in images. Since no annotated data is available for unsupervised methods, various settings have been explored to generate cues that can enable and guide learning. Earlier works that perform object localization require single-class datasets, that is, images that contain the same objects of one particular category. Such a setting is known as object co-localization and was later extended to include noisy or distractor images, which are images present in the dataset but that do not contain the relevant object. The same idea fostered a separate setting known as co-segmentation, which performs localization by predicting masks instead of bounding boxes.

These settings were extended to include images with different categories of objects. In the case of bounding box localization methods, this task started to be called multi-class object discovery, which includes methods that only perform localization and methods that also predict categories for the localized objects. In the case of pixel-wise localization, the task is known as unsupervised semantic segmentation, where methods seek to label the segmented objects, and foreground extraction, where methods localize objects with masks but do not perform categorization.

In recent years, these settings have been extended even further by also seeking to learn latent representation of objects that allow performing decomposition of images into independent objects---\hyperref[fig:chrono]{Fig.~\ref{fig:chrono}} shows a chronological progression of these emerging methods. This task is usually called image (scene) decomposition and object-centric learning. Two main variants have arisen in this direction: methods that perform structured latent representation of objects, using the idea of slots to capture objects and latent variables to capture their properties, and prototype-based methods that uncover objects as transformations of learned prototypes.

\subsection{Levels of localization}
There exist particular ways in which methods have approached the task of discovering objects in images without resorting to ground-truth information for supervision. For instance, as we have already described, some methods seek to categorize images without worrying about the location of objects and such task can be seen as unsupervised image classification. On the other hand, there exist other types of methods that do aim to localize objects; these can perform the localization task by either predicting bounding-box locations or pixel-wise masks. Consequently, a natural way to understand this diversity of methods is found by considering the level of localization being pursued, Therefore, there are three different categories depending on the level of localization: methods that do not localize and thus only categorize, methods that localize using bounding-Boxes and methods that generate pixel-wise masks for localization. 

\begin{figure}
    \centering
    \includegraphics[width=1.0\textwidth]{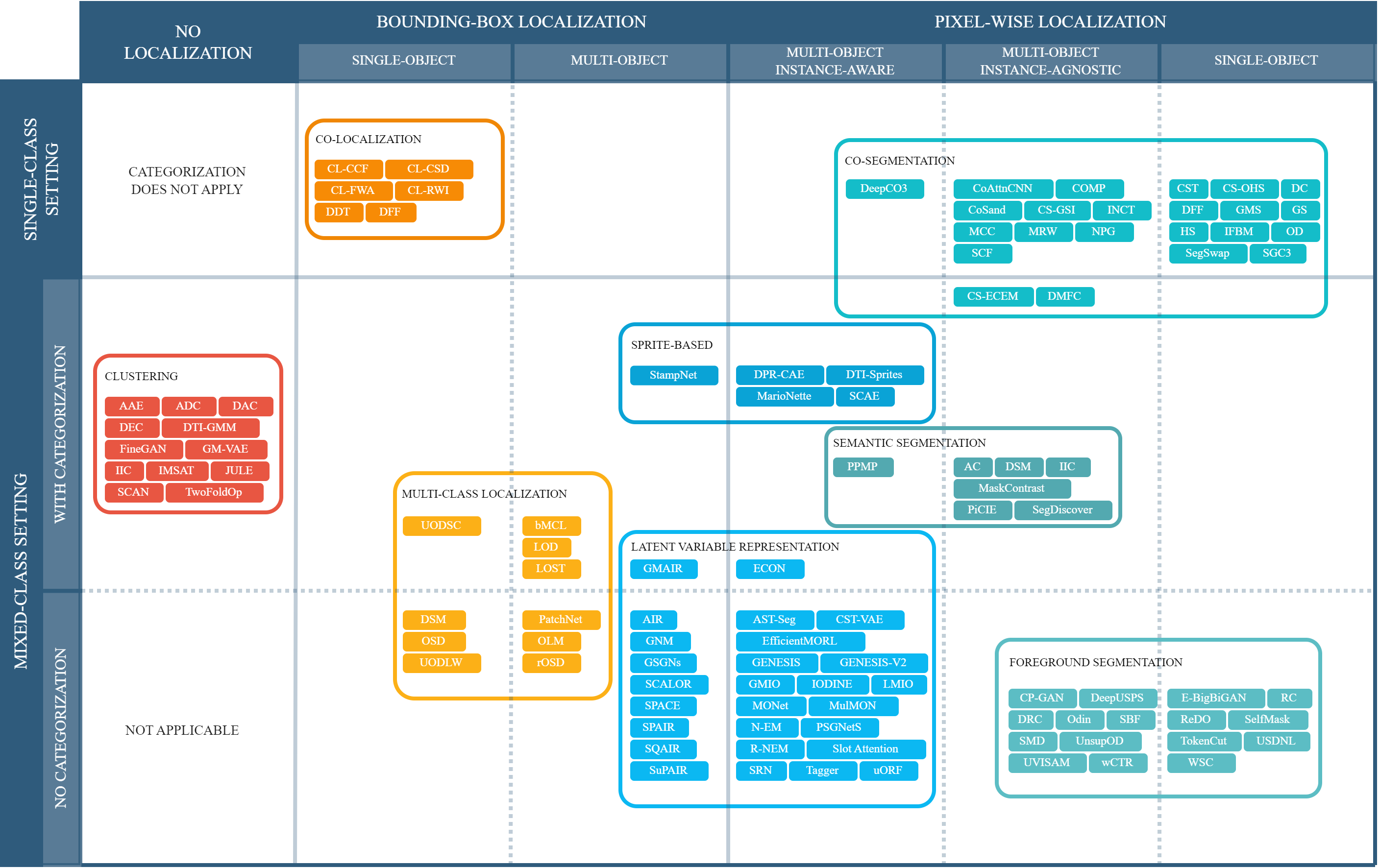}
    \caption{Levels of localization vs Quantity of objects per image}
    \label{fig:levels}
\end{figure}

As we have already described, methods for unsupervised object discovery that do not perform localization, essentially perform unsupervised image classification, a task that is usually tackled by implementing clustering mechanisms. The problem of clustering seeks to identify similarities among a set of data points and then to group together similar data points. Considering that Image classification is the task of assigning a semantic label from a predefined set of classes to an image in the supervised learning regime, clustering methods can be designed or adapted for unsupervised image classification and therefore essentially perform class discovery of objects in images. For instance, the Adversarial Autoencoder \cite{makhzani2016adversarial} is a general approach that uses the generative adversarial networks to disentangle the style and content of images and can be adapted to perform unsupervised image clustering, On the other hand, \cite{monnier2020deep} proposed a framework that enables performing transformation-invariant clustering using complex transformations. As well, the method proposed in IIC \cite{ji2019invariant} directly turns a neural network into a classification function by maximizing mutual information between assignments for paired data samples obtained via random transformations.

A further level of localization is concerned with generating predictions for object locations using tight bounding-boxes around objects. One particular way in which the methods in this group differ from the rest, considers that they can be generally formulated in two different settings. These two ways of framing the bounding-box localization problem include single-class settings where a common object---of a particular class, is found in only positive sets of example images; and the multi-class settings, in which methods seek to discover objects in collections of images that contain multiple categories of objects. However, when considering the quantity of objects that these methods seek to localize,  A a further distinction arises. Indeed, some methods assume a single object per image while others can deal with multiple locations of objects per image. Some  works in this level of localization include \cite{liCLCSD2016image} that tackles the case of a single-class setting by learning to emulate a good object detector’s score distributions of object proposals and refining the location via graph-cut segmentation. As well, \cite{vo2021large} proposed a method that turned the graph-based formulation of \cite{vo2020toward, vo2019unsupervised} into a ranking problem to discover multiple objects per image.

Finally, the third level of localization is the most detailed level and seeks to find locations of objects by predicting pixel-wise assignments of pixels to objects. There are two main groups of methods that correspond to this level. The first group comprises methods that are geared towards performing image segmentation tasks such as co-segmentation \cite{hsu2018coattention, dai2013cosegmentation, rubinstein2013unsupervised, hsu2019deepco3, shen2022learning, tao2017image}, foreground/background extraction \cite{arandjelovic2019object, voynov2021object, chen2019unsupervised, wang2022self, henaff2022object} and semantic segmentation \cite{cho2021picie, vangansbeke2021unsupervised, melas-kyriazi2022deep, karazija2022unsupervised}. In this third level of localization, the second group of methods corresponds to a relatively recent development that is usually referred to as object-centric representation learning or structured decomposition. Methods that perform such tasks not only seek to predict segmentation masks in order to locate objects in images but they mainly aim to learn an interpretable representations of objects in latent space, that allows for scene reconstruction and novel view synthesis. One of the first methods proposed in this direction is \cite{eslami2016attend}, which utilized a structured variational auto-encoder \cite{rezende2014stochastic, kingma2013auto} to learn a set of latent variables through a sequential attention model to generate independent components corresponding to objects. An overview of this classification is observed in \hyperref[fig:levels]{Fig.~\ref{fig:levels}}.

\subsection{Tackling the Unsupervised Regime}
Designing models to infer object locations and categories in an unsupervised regime is challenging. The absence of ground truth information during training requires the incorporation of certain mechanisms into the model’s architecture and pipelines to allow them to generate, capture and leverage learning signals. Indeed, there are well-defined ways in which different methods tackle the lack of annotated data. For instance, a group of methods require pret-rained neural networks to first extract deep features for images and then perform a subsequent mechanism over the externally learned feature space to infer object locations. On the other hand, other methods rely on techniques such as mutual information maximization, graph optimization, image reconstruction or pseudo-label generation to guide the learning process unsupervisedly.

To this date, a clear distinction about what exactly is an unsupervised machine learning model is yet to be agreed upon. Many of the works studied as part of this survey, deem themselves as unsupervised models even though they may make use of deep learning models pre-trained on ImageNet with supervision, use pret-rained models to infer saliency maps to generate pseudo-labels for segmentation networks, require images with well-defined prominent objects mostly centered within the image, or even require a single-class image dataset containing a common category of object. On the other hand, it is certain that unsupervised methods need to establish some kind of guidance to enforce learning under a regime with no access to labeled data. Taking this into consideration, in this survey we consider that a method is unsupervised if it does not require any sort of direct supervision for the specific task to which it is aimed.

\subsubsection{Fully-Unsupervised Settings}
Certain methods are specifically designed to utilize unlabeled data as the only input to perform object discovery tasks. These methods do not rely on any additional guidance from external mechanisms, which could typically include pre-trained networks---usually trained on ImageNet with supervision---for extracting deep features, saliency maps or object proposals. We refer to this kind of methods as fully-unsupervised methods.

On the contrary, fully-unsupervised methods often rely on the intrinsic structure and patterns that can be inferred directly from the data itself to identify and localize objects. These methods can rely on techniques such as clustering, self-supervised learning, and generative modeling, which are commonly employed to discover meaningful and semantic representations that can be exploited for discovering objects in different ways. 

The importance of leveraging and harnessing the intrinsic relationships and patterns in the data to predict meaningful information such as the spatial location or the class of objects, relies in the huge advantage of bypassing the need for any form of labeled supervision which is typically costly to obtain. Moreover, it enables the capacity to achieve a high level of generalization and adaptability across diverse and unseen datasets.

\subsubsection{Leveraging Pre-trained Models}
The well-known success of deep learning on a variety of visual recognition tasks in addition to the usual need for huge amounts of labeled data in long training processes, have motivated a line of research commonly known as model reuse. In essence, the idea is to foster reuse of a pre-trained model in domains that are different from what it was originally trained for, but without requiring further retraining or fine-tuning. This strategy becomes quite an interesting idea when we consider that models pre-trained on ImageNet ---a dataset comprising over a thousand categories of objects in a collection of over 1.2 million images, are capable of learning useful feature representations that can be used for object localization.

Consequently, there exist an important group of methods that share the similarity of relying in the power of pre-trained networks in order to extract and leverage deep features. In such manner, once an image has been translated into feature space, different strategies can be employed to infer object locations from those features. For instance, \cite{zhang2020object} leverage this idea by proposing a pattern mining-based technique for localizing object regions from a single unlabeled image, \cite{simeoni2021localizing} extracts features from a self-supervised pre-trained vision transformer and applies a heuristic seed expansion strategy using patch correlations within an image, \cite{wang2022self} introduces a graph formulation with tokens from a vision transformer as nodes and obtains a segmentation using Normalized Cut, while \cite{wei2017unsupervised} proposes to utilize PCA over activations obtained from a pre-trained CNN to cluster image regions into foreground and background and infers a bounding-box from the largest connected component of foreground pixels. Although, it is worth noting that some methods like \cite{wei2017unsupervised} and \cite{zhang2020object} uses CNN pre-trained with supervision for a different task, while methods like \cite{wang2022self} and \cite{simeoni2021localizing} rely on pre-trained models that do not make use of supervised data whatsoever.

\subsubsection{Leveraging Different Modalities and Additional Cues}
As the main challenge in an unsupervised regime is to stablish a mechanism for generating learning signals, some methods have explored incorporating additional data modalities and other kinds of cues to help guide the learning process of the models. These types of methods certainly seek to perform the object discovery in images in an unsupervised fashion, and to that end, they require some extra information about the data that they still try to leverage without supervision. This extra information can come in the form of multi-view images, single-class image collections containing only one category of object, temporal information from subsequent video frames and so on. For instance, \cite{yu2022unsupervised} is trained on multi-view RGB images to perform scene decomposition and obtain object segmentation, \cite{choudhury2022guess} proposes a model that learns to segment objects by predicting motion of image regions using optical flow as a source of supervision, and \cite{jiang2020scalor} proposes a probabilistic temporal generative model within the VAE framework to perform object-centric decomposition using video sequences. The latter is also similar to a line of methods that seek to perform object-centric structured representation learning in order to decompose an image into separate components where each belongs to a different object. Although these methods work in an unsupervised regime, most of them \cite{greff2019multi, engelcke2020genesis, engelcke2022genesisv2, locatello2020objectcentric} require synthetic datasets that contain geometric objects in well defined uncluttered backgrounds. 

\section{Methods in Unsupervised Object Discovery}
\label{sec:families}

\begin{table}[]
\centering
\resizebox{0.9\textwidth}{!}{%
\begin{tabular}{@{}ll@{}}
\toprule
\multicolumn{2}{c}{\MakeUppercase{Terminology of the comparative tables of methods}}                                                                        \\ \midrule
\multicolumn{2}{c}{\cellcolor[HTML]{DDDDDD}Shared Characteristics}                                                                                     \\ \midrule
Method             & Acronym or name of the method. If none was proposed by the authors, an abbreviation is proposed. \\
Pub.               & Journal, Conference or Repository where the paper has been published.                                              \\
Year               & Year of publication. (Might differ from date when first available.)                                                \\
Model              & A brief description of the model's formulation or main ideas of the approach.                                      \\
Optimization [Solution]       & A short description of the main techniques utilized to solve the problem's formulation.                            \\
Representation     & Type of representation or features used.                                                                           \\
Learning           & Details about the approach to learning, including losses, learning signal, etc.  

 \\ \midrule
\multicolumn{2}{c}{\cellcolor[HTML]{DDDDDD}Shared properties}                                                                                     \\ \midrule

Intra-image Similarities           & Whether the method explicitly exploits intra-image similarities in the formulation.\\

Inter-image Similarities           & Whether the method explicitly exploits inter-image similarities in the formulation. \\

Probabilistic formulations         & Determines if the method works under a probabilistic framework. \\

Graph-based techniques           & Refers to the usage of techniques for graph processing. \\
Generative modeling           & Whether or not the method leverages generative modeling. \\
Deep learning           & Determines if a method uses deep learning techniques. \\
Code released           & Refers to the availability of official code for the method. 

 \\ \midrule
\multicolumn{2}{c}{\cellcolor{colorclust}{\textcolor{white}{\textbf{Clustering}}}}                                             \\ \midrule

Pipeline           & Specifies the general architectural setting of the method. \\

Clustering Type    & Determines the type of clustering strategy incorporated into the formulation.  \\      

Membership         & Mechanism utilized to implement cluster assignment.                                                                \\ \midrule
\multicolumn{2}{c}{\cellcolor{colorloc}{\textcolor{white}{\textbf{Localization}}}}                                             \\ \midrule
Clustering         & Determines which type of clustering technique is used if any.                                                      \\
Saliency Priors    & Techniques or methods used to extract salient regions from which bounding boxes are derived.                       \\
Box Priors         & Techniques or methods used to generate traditional object proposals. (rectangular windows)                         \\
Box Generation     & Mechanism employed to produce the output bounding box.                                                             \\ \midrule
\multicolumn{2}{c}{\cellcolor{colorseg}{\textcolor{white}{\textbf{Segmentation}}}}                                             \\ \midrule
Clustering         & Determines which type of clustering technique is used if any.                                                      \\
External saliency  & Third-party techniques or methods used to extract saliency masks.                                                  \\
Oversegmentation   & Type of oversegmentation techniques employed if any.                                                               \\
Mask Generation    & Mechanism employed to produce the output segmentation mask.                                                        \\ \midrule
\multicolumn{2}{c}{\cellcolor{colordec}{\textcolor{white}{\textbf{Decomposition}}}}                                            \\ \midrule
Attention Strategy & Defines if objects are localized by rectangular regions (glimpse) or pixel-wise (mask).                            \\
Object Routing     & Approach to the inference of object representations.                                                               \\
Reconstruction     & Determines how the image is composed or reconstructed from the latent representations.                             \\ \bottomrule
\end{tabular}%
}
\captionsetup{width=0.9\textwidth}
\caption{Explanation of different terms used to describe diverse characteristics of methods in this survey.}
\label{tab:terminology}
\end{table}

The wide variety of approaches that have tackled tasks pertinent to objects discovery spans a rich diversity of techniques. Some techniques have even been employed across multiple groups of methods. For instance, VAEs \cite{kingma2013auto, rezende2014stochastic} are frequently used in the design of methods for object-centric decomposition \cite{zhu2021gmair, greff2019multi, engelcke2020genesis} but have also been explored in box-localization \cite{moon2021patchnet}.

The primary objective of this section is to present an organized and comprehensive scheme that facilitates the exploration of the relevant literature on unsupervised object discovery. To this end, we have grouped methods according to criteria that focuses on the aims of the methods as well as the families of techniques employed. Consequently, to better understand this immense variety of approaches, we have structured a simple taxonomy comprising the common families of methods, in addition to the common methodological choices undertaken in their design. In the following subsections, we elaborate on each part of this taxonomy; and present the relevant and recurring ideas introduced by several of these methods.

In each subsection, we start by identifying recurring techniques utilized across the different groups of methods regarding the type of model or architecture, optimization techniques employed, the type of representation and relevant details about how learning takes place (See \hyperref[tab:terminology]{Table~\ref{tab:terminology}} for an outline of the terminology used). Furthermore, an overview of all methods in this survey can be appreciated in \hyperref[tab:overview_all]{Table~\ref{tab:overview_all}}. 

\let\conftitle\tiny

{\tiny\tabcolsep=1.2pt  

\captionsetup[longtable]{width=0.98\linewidth} 
\captionsetup[longtable]{skip=2.5em}
\begin{longtable}{p{0.7em}llccccc}

\toprule
\multicolumn{8}{c}{\cellcolor[HTML]{DDDDDD} {\MakeUppercase{Characteristics among Methods in this survey}}}                                 \\ \midrule
   &
  Method &
  \begin{tabular}[c]{@{}c@{}}Pub.\end{tabular} &
  \begin{tabular}[c]{@{}c@{}}Year\end{tabular} &
  \begin{tabular}[c]{@{}c@{}}Model\end{tabular} &
  \begin{tabular}[c]{@{}c@{}}Optimization\end{tabular} &
  \begin{tabular}[c]{@{}c@{}}Representation\end{tabular} &
  \begin{tabular}[c]{@{}c@{}}Learning\end{tabular} \\ \midrule

\endfirsthead

\multicolumn{8}{c}{\cellcolor[HTML]{DDDDDD} {\MakeUppercase{characteristics among Methods in this survey}}}                                 \\ \midrule
   &
  Method &
  \begin{tabular}[c]{@{}c@{}}Pub.\end{tabular} &
  \begin{tabular}[c]{@{}c@{}}Year\end{tabular} &
  \begin{tabular}[c]{@{}c@{}}Model\end{tabular} &
  \begin{tabular}[c]{@{}c@{}}Optimization\end{tabular} &
  \begin{tabular}[c]{@{}c@{}}Representation\end{tabular} &
  \begin{tabular}[c]{@{}c@{}}Learning\end{tabular} \\ \midrule

\endhead            
\midrule\multicolumn{8}{c}{{Continued on next page}} \\ \midrule
\endfoot    
\endlastfoot


\cellcolor{colorclust}& DEC & \conftitle{ICML} & 2015 & Repr. learning/Probab. clust. & Backprop/Refinement (k-means) & Learned & Target distrib.  \\ [-\arrayrulewidth]

\rowcolor{rowbackground}
 \cellcolor{colorclust}& AAE	& \conftitle{ARXIV} & 2016 & Autoencoder/Adversarial net & Backprop/Adversarial training & Learned & Img. reconstruction  \\ [-\arrayrulewidth]
\cellcolor{colorclust}& JULE & \conftitle{CVPR} & 2016 & Rec. CNN/Agglom. clust. & Backprop/Greedy Search & Learned & Weighted triplet loss  \\ [-\arrayrulewidth]

\rowcolor{rowbackground}
\cellcolor{colorclust} & DAC & \conftitle{ICCV} & 2017 & Binary pairwise-classifier & Backprop/Adaptive training & Pretrained & Pairwise similarities  \\ [-\arrayrulewidth]
\cellcolor{colorclust} & GMVAE & \conftitle{ICLR} & 2017 & Variational autoencoder & Backprop/Variational inference & Learned & ELBO  \\ [-\arrayrulewidth]

\rowcolor{rowbackground}
\cellcolor{colorclust} & IMSAT & \conftitle{ICML} & 2017 & Discrete repr. learning & Backprop/Information Theory & Pretrained & Transformations/MI  \\ [-\arrayrulewidth]
\cellcolor{colorclust} & ADC	& \conftitle{GCPR} & 2018 & Classification network & Backprop/Associative clustering & Learned & Transform. invariant loss  \\ [-\arrayrulewidth]

\rowcolor{rowbackground}
\cellcolor{colorclust} & FineGAN & \conftitle{CVPR} & 2019 & Hierarchical GAN & Backprop/Adversarial training & Learned & Adversarial + MI  \\ [-\arrayrulewidth]
\cellcolor{colorclust} & IIC & \conftitle{ICCV} & 2019 & Classification network & Backprop/Information Theory & Learned & Transformations/MI  \\ [-\arrayrulewidth]

\rowcolor{rowbackground}
\cellcolor{colorclust} & DTI-GMM	& \conftitle{NIPS} & 2020 & Transformation prediction & Backprop/EM & Learned & Transform. invariant loss  \\ [-\arrayrulewidth]
\cellcolor{colorclust} & SCAN & \conftitle{ECCV} & 2020 & Repr. learning+Classif. net. & Backprop/Nearest neighbors & Learned & Semantic clustering loss \\ [-\arrayrulewidth]

\rowcolor{rowbackground}
\multirow[c]{-12}{0.7em}{\cellcolor{colorclust}
\begin{sideways}\textcolor{white}{\textbf{CLUSTERING}}\end{sideways}} & TwoFoldOp & \conftitle{NEUROC} & 2022 & Repr. learning+Classif. net. & Backprop/Alternating loss (EM-like) & Learned & Clust. loss/ KL aug.  \\ [-\arrayrulewidth]

\midrule

\cellcolor{colorloc} & bMCL & \conftitle{TPAMI} & 2014 & MIL framework &  Discriminative EM & Hand-crafted & \EMPTY \\ [-\arrayrulewidth]

\rowcolor{rowbackground}
\cellcolor{colorloc} & CL-FWA &  \conftitle{ECCV} & 2014 & Combinatorial/Quad. progr. &  Frank-Wolfe algorithm & Hand-crafted & \EMPTY \\ [-\arrayrulewidth]
\cellcolor{colorloc} & CL-RWI & \conftitle{CVPR} & 2014 & Joint formulation/Ridge regress. & Quadratic program solver & Hand-crafted & \EMPTY \\ [-\arrayrulewidth]

\rowcolor{rowbackground}
\cellcolor{colorloc} & UODLW & \conftitle{CVPR} & 2015 & Probabilistic Matching & Scoring/Nearest neighbors & Hand-crafted & \EMPTY \\ [-\arrayrulewidth]
\cellcolor{colorloc} & CL-CSD &  \conftitle{ECCV} & 2016 & Prop. scoring/Graph-based & SGD/Graph cuts & Pretrained & \EMPTY  \\ [-\arrayrulewidth]

\rowcolor{rowbackground}
\cellcolor{colorloc} & CL-CCF & \conftitle{ICCV} & 2017 & Graph-based/Geodesic propag. & Thresholding & Pretrained & \EMPTY \\ [-\arrayrulewidth]
\cellcolor{colorloc} & DDT & \conftitle{PATCOG} & 2018 & Feature Correlation Matrix & Thresholding & Pretrained & \EMPTY \\ [-\arrayrulewidth]

\rowcolor{rowbackground}
\cellcolor{colorloc} & OSD & \conftitle{CVPR} & 2019 & Cubic Pseudo-Boolean function &  Greedy block-coordinate ascent & Hand-crafted & \EMPTY \\ [-\arrayrulewidth]
\cellcolor{colorloc} & UOSDC & MTA & 2019 & Feature clustering & Spherical clustering & Pretrained & \EMPTY \\ [-\arrayrulewidth]

\rowcolor{rowbackground}
\cellcolor{colorloc} & OLM &  \conftitle{TIP} & 2020 & Pattern mining/Transact. & Data mining on CNN layers & Pretrained & \EMPTY  \\ [-\arrayrulewidth]
\cellcolor{colorloc} & rOSD & \conftitle{ECCV} & 2020 & OSD+regularization &  Greedy block-coordinate ascent & Pretrained & \EMPTY \\ [-\arrayrulewidth]

\rowcolor{rowbackground}
\cellcolor{colorloc} & LOD &  \conftitle{NIPS} & 2021 & Ranking/Graph-based & Quadratic op./PageRank & Pretrained & \EMPTY \\ [-\arrayrulewidth]
\cellcolor{colorloc} & LOST &  \conftitle{BMVC} & 2021 & Expansion strategy/Graph-based & Corr.-based patch selection & Pretrained & \EMPTY \\ [-\arrayrulewidth] 

\rowcolor{rowbackground}
\cellcolor{colorloc} & PatchNet &  \conftitle{ARXIV} & 2021 & Encoder/Decoder & Variational Bayes/Clustering & Learned & Contrastive   \\ [-\arrayrulewidth]
\cellcolor{colorloc} & TokenCut &  \conftitle{CVPR} & 2022 & Graph-based spectral & Normalized cut & Pretrained & \EMPTY  \\ [-\arrayrulewidth]

\rowcolor{rowbackground}
\multirow[c]{-16}{0.7em}{\cellcolor{colorloc}
\begin{sideways}\textcolor{white}{\textbf{LOCALIZATION}}\end{sideways}} & MOST & \conftitle{ICCV} & 2023 & Token Clust./Fractal-Based  & Box-counting/Clustering & Pretrained & \EMPTY   \\ [-\arrayrulewidth]

\midrule

\cellcolor{colorseg} & DC & \conftitle{ICML} & 2010 & Semidefinite program & Low-rank op. & Hand-crafted & \EMPTY \\ [-\arrayrulewidth] 

\rowcolor{rowbackground}
\cellcolor{colorseg} & CoSand	& \conftitle{ICCV} & 2011 & Temperature maximization & Submodular op.+Ncut & Hand-crafted & \EMPTY  \\	 [-\arrayrulewidth] 

\cellcolor{colorseg} & IFBM & \conftitle{CVPR} & 2012 & MRF-based & Graph cuts & Hand-crafted & \EMPTY \\  [-\arrayrulewidth] 

\rowcolor{rowbackground}
\cellcolor{colorseg} & MCC	& \conftitle{CVPR} & 2012 & Combinatorial pixel labeling & Block-coord. descent & Hand-crafted & \EMPTY \\  [-\arrayrulewidth] 

\cellcolor{colorseg} & COMP & \conftitle{ICCV} & 2013 & Score-based refinement & Heuristic seg.+Grabcut & Hand-crafted & \EMPTY \\  [-\arrayrulewidth] 

\rowcolor{rowbackground}
\cellcolor{colorseg} & CST & \conftitle{ICCV} & 2013 & MRF-based & Graph cuts & Hand-crafted & \EMPTY \\  [-\arrayrulewidth] 

\cellcolor{colorseg} & HS	& \conftitle{CVPR} & 2013 & Tree-structure graphical & Belief propagation & Hand-crafted & \EMPTY \\  [-\arrayrulewidth] 

\rowcolor{rowbackground}
\cellcolor{colorseg} & OD	& \conftitle{CVPR} & 2013 & Pixel correspondence optim. & Coordinate descent & Hand-crafted & \EMPTY \\ [-\arrayrulewidth]

\cellcolor{colorseg} & GMS	& \conftitle{ICIP} & 2014 & Global saliency fusion & Grabcut & Hand-crafted & \EMPTY	 \\ [-\arrayrulewidth] 

\rowcolor{rowbackground}
\cellcolor{colorseg} & CS-ECEM	& \conftitle{TCSVT} & 2014 & Superpixel labelling & \(\alpha\)-Expansion algorithm & Hand-crafted & \EMPTY \\ [-\arrayrulewidth] 

\rowcolor{rowbackground}
\cellcolor{colorseg} & wCTR & \conftitle{CVPR} & 2014 & Graph-based approach & Least-square op. & Hand-crafted & \EMPTY \\ [-\arrayrulewidth] 

\cellcolor{colorseg} & DMFC & \conftitle{CVIU} & 2015 & Graph-based approach & Alternating op./Grabcut & Hand-crafted & \EMPTY \\ [-\arrayrulewidth] 

\rowcolor{rowbackground}
\cellcolor{colorseg} & MRW & \conftitle{CVPR} & 2015 & Graph-based approach & Random walkers & Hand-crafted & \EMPTY \\ [-\arrayrulewidth] 

\cellcolor{colorseg} & WSC & \conftitle{CVPR} & 2015 & Dictionary-based saliency & Sparse coding & Hand-crafted & \EMPTY \\ [-\arrayrulewidth] 

\rowcolor{rowbackground}
\cellcolor{colorseg} & SCF & \conftitle{TMM} & 2016 & Quadratic programming & Grabcut & Hand-crafted & \EMPTY \\
\cellcolor{colorseg} & SMD	& \conftitle{TPAMI} & 2016 & Structured matrix decomp. & Felzenszwalb algorithm & Hand-crafted & \EMPTY \\ [-\arrayrulewidth]

\rowcolor{rowbackground}
\cellcolor{colorseg} & NPG & \conftitle{TIP} & 2017 & Mixed-Integer programs & Grabcut & Pretrained & \EMPTY \\ [-\arrayrulewidth]
\cellcolor{colorseg} & SBF & \conftitle{ICCV} & 2017 & Saliency network & Backprop & Learned & Fused weak saliency \\ [-\arrayrulewidth]

\rowcolor{rowbackground}
\cellcolor{colorseg} & SGC3 & \conftitle{AAAI} & 2017 & Constrained clustering & K-means like optim. & Hand-crafted & \EMPTY \\ [-\arrayrulewidth] 
\cellcolor{colorseg} & CoAttCNN & \conftitle{IJCA} & 2018 & Two-stage optimization & Backprop & Pretrained & Off-the-shelf proposals \\ [-\arrayrulewidth] 

\rowcolor{rowbackground}
\cellcolor{colorseg} & CS-GSI & \conftitle{NEUCOM} & 2018 & Tree structure graph & Grabcut & Hand-crafted & \EMPTY \\ [-\arrayrulewidth] 
\cellcolor{colorseg} & DFF & \conftitle{ECCV} & 2018 & Non-negative factorization & Low-rank approx. & Pretrained & \EMPTY \\ [-\arrayrulewidth] 

\rowcolor{rowbackground}
\cellcolor{colorseg} & USDNL & \conftitle{CVPR} & 2018 & Saliency network & Backprop & Learned & Off-the-shelf saliency \\ [-\arrayrulewidth] 
\cellcolor{colorseg} & CP-GAN & \conftitle{ARXIV} & 2019 & GAN & Backprop & Learned & Adversarial \\ [-\arrayrulewidth] 

\rowcolor{rowbackground}
\cellcolor{colorseg} & DeepCO3 & \conftitle{CVPR} & 2019 & Multi-loss optimization & Backprop & Refined &  Off-the-shelf saliency \\ [-\arrayrulewidth] 
\cellcolor{colorseg} & IIC & \conftitle{ICCV} & 2019 & Clustering neural network & Backprop & Learned & Mutual inform. \\ [-\arrayrulewidth] 

\rowcolor{rowbackground}
\cellcolor{colorseg} & ReDo & \conftitle{NIPS} & 2019 & GAN & Backprop & Learned & Aversarial \\ [-\arrayrulewidth] 
\cellcolor{colorseg} & DeepUSPS & \conftitle{NIPS} & 2019 & Saliency network & Backprop & Learned & Pseudo-labels \\ [-\arrayrulewidth] 

\rowcolor{rowbackground}
\cellcolor{colorseg} & UnsupOD & \conftitle{SYMM.} & 2020 & Segmentation network & Backprop & Learned & Img. reconstruction \\ [-\arrayrulewidth] 
\cellcolor{colorseg} & CS-OHS & \conftitle{JCS} & 2021 & MRF-based & Particle Swarm & Hand-crafted & \EMPTY \\	 [-\arrayrulewidth] 

\rowcolor{rowbackground}
\cellcolor{colorseg} & DRC & \conftitle{NIPS} & 2021 & Generative modeling - MoE & EM-like/MCMC/Bprop & Learned & Img. reconstruction \\ [-\arrayrulewidth] 
\cellcolor{colorseg} & E-BigBigGAN & \conftitle{PMLR} & 2021 & Segmentation network & Backprop & Refined & Synthetic masks \\ [-\arrayrulewidth] 

\rowcolor{rowbackground}
\cellcolor{colorseg} & MaskContrast & \conftitle{ICCV} & 2021 & Pixel-embedding learning & Backprop & Learned & Contrastive \\ [-\arrayrulewidth] 
\cellcolor{colorseg} & PiCIE & \conftitle{CVPR} & 2021 & Pixel-embedding learning & Backprop & Refined & Cluster assignments \\ [-\arrayrulewidth] 

\rowcolor{rowbackground}
\cellcolor{colorseg} & UVISAM & \conftitle{BMVC} & 2022 & Segmentation network & Backprop & Refined & Optical flow \\ [-\arrayrulewidth] 
\cellcolor{colorseg} & DSM & \conftitle{CVPR} & 2022 & Graph-based spectral & Normalized cut & Pretrained & \EMPTY \\ [-\arrayrulewidth] 

\rowcolor{rowbackground}
\cellcolor{colorseg} & MOVE & \conftitle{NIPS} & 2022 & Segmenter/Inpainter net. & Bprop/Adversarial tr. & Pretrained & Composite image \\ [-\arrayrulewidth] 

\multirow[c]{-36}{0.7em}{\cellcolor{colorseg}
\begin{sideways}\textcolor{white}{\textbf{SEGMENTATION}}\end{sideways}}
 & SegDiscover & \conftitle{ARXIV} & 2022 & Segmentation network & Backprop & Refined & Cluster assignments \\ [-\arrayrulewidth] 

\rowcolor{rowbackground}
\cellcolor{colorseg} & SegSwap & \conftitle{CVPR} & 2022 & Graph-based pair matching & Grabcut & Pretrained & Correspondence pairs \\ [-\arrayrulewidth] 
\cellcolor{colorseg} & SelfMask & \conftitle{CVPRw} & 2022 & Saliency network & Backprop & Learned & Ranked saliency mask \\ [-\arrayrulewidth] 

\rowcolor{rowbackground}
\cellcolor{colorseg} & TokenCut & \conftitle{CVPR} & 2022 & Graph-based spectral & Normalized cut & Pretrained & \EMPTY \\ [-\arrayrulewidth] 
\cellcolor{colorseg} & CutLER & \conftitle{CVPR} & 2023 & Segmentation network & Ncut/Backprop & Pretrained & MaskCut pseudo-labels \\ [-\arrayrulewidth] 

\rowcolor{rowbackground}
\cellcolor{colorseg} & SEMPART & \conftitle{ICCV} & 2023 & Graph-based approach & Backprop & Refined & Coarse/fine bi-partition \\ [-\arrayrulewidth] 

\cellcolor{colorseg} & 
U2Seg & \conftitle{CVPR} & 2024 & Segmentation network & EM/Backprop & Refined & Pseudo-labels \\ [-\arrayrulewidth]

\midrule

\cellcolor{colordec} & AIR	& \conftitle{NIPS} & 2016 & Encoder/Decoder	& Variational Bayes & Learned & ELBO	\\ [-\arrayrulewidth] 

\rowcolor{rowbackground}
\cellcolor{colordec} & CST-VAE	& \conftitle{ICLR} 	& 2016	& Encoder/Decoder	& Variational Bayes & Learned & ELBO	\\ [-\arrayrulewidth] 
\cellcolor{colordec} & TAGGER & \conftitle{NIPS} & 2016 & Denoising AE & Backprop & Learned & NLL \\ [-\arrayrulewidth] 

\rowcolor{rowbackground}
\cellcolor{colordec} & N-EM & \conftitle{NIPS} & 2017 & Recurrent AE & Backprop & Learned & Clustering \\ [-\arrayrulewidth] 
\cellcolor{colordec} & R-NEM & \conftitle{ICLR} & 2018 & Recurrent AE & Backprop & Learned & Clustering \\ [-\arrayrulewidth] 

\rowcolor{rowbackground}
\cellcolor{colordec} & SQAIR & \conftitle{NIPS} & 2018 & Recurrent VAE & Variational Bayes & Learned & IWAE \\ [-\arrayrulewidth] 
\cellcolor{colordec} & GMIO & \conftitle{ICML} & 2019 & Encoder/Decoder & Variational Bayes & Learned & ELBO \\ [-\arrayrulewidth] 

\rowcolor{rowbackground}
\cellcolor{colordec} & IODINE & \conftitle{ICML} & 2019 & Encoder/Decoder & Variational Bayes/IAI & Learned & weigthed ELBO \\ [-\arrayrulewidth] 
\cellcolor{colordec} & MONet & \conftitle{ARXIV} & 2019 & Recurrent VAE & Variational Bayes & Learned & ELBO+KL term \\ [-\arrayrulewidth] 

\rowcolor{rowbackground}
\cellcolor{colordec} & SCAE & \conftitle{NIPS} & 2019 & Capsule Autoencoder  & Backprop & Learned & Log-likelihood \\ [-\arrayrulewidth] 
\cellcolor{colordec} & SPAIR & \conftitle{AAAI} & 2019 & Encoder/Decoder & Variational Bayes & Learned & ELBO \\ [-\arrayrulewidth] 

\rowcolor{rowbackground}
\cellcolor{colordec} & SuPAIR & \conftitle{ICML} & 2019 & Sum-product Network & Variational Bayes/MC & Learned & ELBO \\ [-\arrayrulewidth] 
\cellcolor{colordec} & ECON  & \conftitle{ICLR} & 2020 & Encoder/Decoder & Variational Bayes & Learned & ELBO \\ [-\arrayrulewidth] 

\rowcolor{rowbackground}
\cellcolor{colordec} & GENESIS & \conftitle{ICLR} & 2020 & Encoder/Decoder & Variational Bayes & Learned & GECO \\ [-\arrayrulewidth] 
\cellcolor{colordec} & GNM & \conftitle{NIPS} & 2020 & Encoder/Decoder & Variational Bayes & Learned & ELBO \\ [-\arrayrulewidth] 

\rowcolor{rowbackground}
\cellcolor{colordec} & LMIO & \conftitle{CVPR} & 2020 & Encoder/Decoder & Variational Bayes & Learned & IBL \\ [-\arrayrulewidth] 
\cellcolor{colordec} & MulMON & \conftitle{NIPS} & 2020 & Encoder/Decoder & Variational Bayes/IAI & Learned & ELBO \\ [-\arrayrulewidth] 

\rowcolor{rowbackground}
\cellcolor{colordec} & SCALOR & \conftitle{ICLR} & 2020 & Encoder/Decoder & Variational Bayes & Learned & ELBO \\  [-\arrayrulewidth] 
\cellcolor{colordec} & SLOT ATTN. & \conftitle{NIPS} & 2020 & Differentiable interface & \EMPTY & Refined & \EMPTY \\ [-\arrayrulewidth] 

\rowcolor{rowbackground}
\cellcolor{colordec} & SPACE & \conftitle{ICLR} & 2020 & Encoder/Decoder & Variational Bayes & Learned & ELBO\\ [-\arrayrulewidth] 
\cellcolor{colordec} & SRN & \conftitle{NIPS} & 2020 & Differentiable interface & \EMPTY & Refined & \EMPTY \\ [-\arrayrulewidth] 

\rowcolor{rowbackground}
\cellcolor{colordec} & DTISprites & \conftitle{ICCV} & 2021 & Transform Predictor & Backprop & Learned & Reconstruction $L_2$ \\ [-\arrayrulewidth] 
\cellcolor{colordec} & EfficientMORL & \conftitle{ICML} & 2021 & Encoder/Decoder & Variational Bayes/IAI & Learned & weigthed ELBO \\ [-\arrayrulewidth] 

\rowcolor{rowbackground}
\cellcolor{colordec} & GENESISv2 & \conftitle{NIPS} & 2021 & Encoder/Decoder & Variational Bayes & Learned & GECO \\ [-\arrayrulewidth] 
\cellcolor{colordec} & GMAIR & \conftitle{CIN} & 2021 & Encoder/Decoder & Variational Bayes & Learned & ELBO \\ [-\arrayrulewidth] 

\rowcolor{rowbackground}
\cellcolor{colordec} & GSNGs & \conftitle{ICLR} & 2021 & Encoder/Decoder & Variational Bayes & Learned & ELBO \\ [-\arrayrulewidth] 
\cellcolor{colordec} & MarioNette & \conftitle{NIPS} & 2021 & Transform Predictor & Backprop & Learned & Reconstruction $L_2$ \\ [-\arrayrulewidth] 

\rowcolor{rowbackground}
\cellcolor{colordec} & uORF & \conftitle{ICLR} & 2022 & Encoder/Decoder & Backprop & Learned & Reconstruction $L_2$ \\ [-\arrayrulewidth] 

\multirow[c]{-29}{0.7em}{\cellcolor{colordec}
\begin{sideways}\textcolor{white}{\textbf{DECOMPOSITION}}\end{sideways}} & 
DINOSAUR & \conftitle{ICLR} & 2023 & Encoder/Decoder & Backprop/Slot Attn. & Refined & Reconstruction $L_2$ \\ [-\arrayrulewidth]

 \bottomrule

\caption{Overview of shared characteristics of methods across the four different groups in Unsupervised Object Discovery.} 

\label{tab:overview_all}
\end{longtable}
}

As evidenced in the table, a diverse array of techniques ranging from mathematical optimization to deep learning architectures have been exploited in the plethora of available methods. To facilitate comprehension of this richness, the column "model" in the table provides a comprehensive glimpse into the diverse landscape of techniques and approaches utilized in the models' design across the various groups of methodologies. Ranging from fundamental mathematical frameworks like semidefinite programs and quadratic programming to more specialized approaches such as graph-based methodologies and generative adversarial networks, these techniques highlight the breadth of approaches that have been explored in tackling the problem of unsupervised object discovery. Furthermore, the emergence of object-based representation learning as an interesting avenue of research has become evident with the proliferation of methods based on variational autoencoders, emphasizing the relevance of the binding problem \cite{greff2020binding} in object discovery and scene decomposition. Overall, this eclectic collection of techniques and models encapsulates the dynamism driving contemporary research in unsupervised object discovery.

\textcolor{colorclust}{\subsection{Clustering Methods}}
Methods in this category essentially discover object classes in collections of images without localizing the objects. One perspective for understanding this process is that these methods perform unsupervised object discovery by clustering the images in a dataset based on the class of the object contained within each image. Therefore, this task seeks to categorize entire collections of images without supervision, implying the grouping of images into clusters that ideally correspond to semantic classes, all without the need for ground-truth labels. Consequently, image clustering methods aim to separate images into coherent and meaningful groups. Unsurprisingly, different avenues have been explored in order to tackle this problem; however, these methods are generally designed as either end-to-end learning pipelines or multi-stage approaches that combine representation learning with clustering techniques.

End-to-end learning methods seek to learn a mapping from the input images directly to the cluster assignments. In other words, the models learn to perform both feature extraction and clustering in a single pass. Thus, end-to-end methods do not rely on pre-text tasks, hand-designed features or pre-processing steps, but instead are trained to learn the entire process. Conversely, multi-stage approaches rely on learning a high-level representation of the data. These methods seek to capture the underlying structure of an image collection and project it onto a feature space. Then, with the learned representations or embeddings, similar instances can be grouped together using different well-known clustering techniques such as k-means and Gaussian Mixture Model clustering. 
\newline

\subsubsection{Common families of Clustering methods for object discovery}

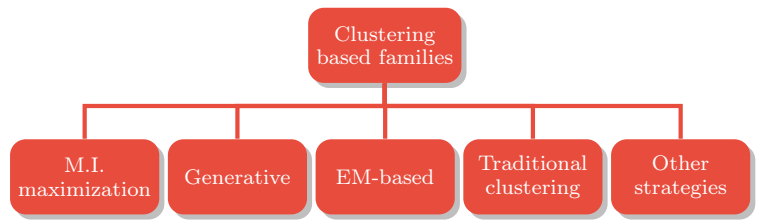
\begin{wrapfigure}{l}{0.6\textwidth}
    \begin{tikzpicture}[
    node/.style={rectangle, draw=none, rounded corners=6pt, align=center, fill=#1, drop shadow, text=white, minimum width=1.85cm, minimum height=1cm, font=\footnotesize},
    line/.style={-, ultra thick, color=colorclust},
    node distance=0.1cm
]
    \node[node=colorclust] (Root) at (0.99,0) {Clustering \\ based families};
    
    \node[node=colorclust] (Leaf1) at (-3,-1.75) {M.I. \\ maximization};
    \node[node=colorclust, right=of Leaf1] (Leaf2)  {Generative};
    \node[node=colorclust, right=of Leaf2] (Leaf3) {EM-based};
    \node[node=colorclust, right=of Leaf3] (Leaf4) {Traditional \\ clustering};
    \node[node=colorclust, right=of Leaf4] (Leaf5) {Other \\ strategies};
    
    \draw[line] (Root.south) -- ++(0,-0.3) -| (Leaf1.north);
    \draw[line] (Root.south) -- ++(0,-0.3) -| (Leaf2.north);
    \draw[line] (Root.south) -- ++(0,-0.3) -| (Leaf3.north);
    \draw[line] (Root.south) -- ++(0,-0.3) -| (Leaf4.north);
    \draw[line] (Root.south) -- ++(0,-0.3) -| (Leaf5.north);
\end{tikzpicture}
    \caption{Families of Clustering methods in UOD}
    \label{fig:chart_fam_clust}
\end{wrapfigure}

Within this group of methods, several avenues of research (see \hyperref[fig:chart_fam_clust]{Fig.~\ref{fig:chart_fam_clust}}) have been explored encompassing both end-to-end and multi-stage approaches. End-to-end approaches typically propose pipelines that combine feature learning with soft clustering; however, other methods can predict cluster centers in a prototype-based fashion to then assign images to clusters based on alignment. 


\vspace{-\abovecaptionskip}

Among end-to-end approaches, a first set of methods learns a clustering function by maximizing mutual information between an image and its augmentations, as exemplified by \ffacronym{IIC} \cite{ji2019invariant}. A second set, leverages an initial feature representation obtained through a CNN to then refine the clusters, as shown in \ffacronym{ADC} \cite{haeusser2019associative} and \ffacronym{DEC} \cite{xie2016unsupervised}. In contrast, multi-stage approaches such as \ffacronym{FineGAN} \cite{singh2019finegan} and \ffacronym{SCAN} \cite{vangansbeke2020scan} rely in a prior for clustering, which can be a pre-text task, that serves to learn good feature representations for clustering. Then, these representations can be clustered simply by using traditional algorithms like k-means. 

\minititle{Maximizing Mutual Information}

A group of methods \cite{ji2019invariant, hu2017learning} proposes to learn a clustering function by maximizing the mutual information \textit{MI} between images and their augmented versions. For instance, Invariant information clustering \ffacronym{IIC} \cite{ji2019invariant} learns cluster assignments by maximizing the \textit{MI} between the embeddings of input images and their transformations. By introducing random perturbations and maximizing information between related pairs of images, the network learns a representation capable of discarding instance-specific details while preserving common features. Then, the network outputs an encoding that can be probabilistically interpreted as the distribution of a discrete random variable over the relevant object classes. Likewise, \ffacronym{IMSAT} \cite{hu2017learning} uses data augmentation to impose invariance on discrete representations learned by a neural network. The method maximizes information-theoretic dependency between data and their predicted discrete representations while regularizing the mapping function through self-augmented training \textit{SAT}. This self-augmented training procedure encourages the predicted representations of augmented data points to be close to those of original data points and is flexible in imposing various types of invariances. Thus combining the \textit{SAT} with the Regularized Information Maximization RIM for clustering, the authors introduced the Information Maximizing Self-Augmented Training method \ffacronym{IMSAT}.

\minititle{Generative}

The application of the \textit{GAN} framework to image clustering has also been explored. The adversarial autoencoder \ffacronym{AAE} \cite{makhzani2016adversarial} is a probabilistic autoencoder that uses generative adversarial networks \textit{GAN} to perform variational inference for discrete and continuous latent variables. The autoencoder is trained with a reconstruction error criterion and an adversarial training criterion that matches the aggregated posterior distribution of the latent representation of the autoencoder to an arbitrary prior distribution. By matching the aggregated posterior to the prior, the decoder learns to map the imposed prior to the data distribution, and this ensures that meaningful samples are generated from any part of the prior space. \ffacronym{GM-VAE} \cite{dilokthanakul2017deep} was introduced as a variant of the \textit{VAE} framework for unsupervised clustering with a Gaussian mixture prior distribution. The authors recognize that over-regularization in \textit{VAEs} leads to cluster degeneracy but can be mitigated with the minimum information constraint heuristic. Thus, they propose \ffacronym{GM-VAE} as an algorithm for unsupervised clustering that optimizes the inference model using a variational Bayes objective and assumes that the observed data is generated from a multimodal prior distribution. 

\minititle{EM-based}

Expectation-Maximization \textit{EM} is an iterative algorithm frequently applied to approximate solutions for the maximum likelihood estimate \textit{MLE} and maximum a posteriori \textit{MAP} estimate. It can be particularly useful in solving mixture modeling problems, where a combination of distributions is fitted to the observed data to maximize the total likelihood. Since maximizing this likelihood is often challenging, EM optimizes an alternative lower bound instead. This process involves iteratively estimating and refining distribution parameters to fit the model. Some recent clustering methods, such as \ffacronym{TwoFoldOp} \cite{tissera2021neural} and \ffacronym{DTI-GMM} \cite{monnier2020deep}, have explored incorporating EM-prodcedures into neural network training, in order to predict the statistical parameters of cluster distributions. For instance, \ffacronym{TwoFoldOp} \cite{tissera2021neural} was proposed as a method for clustering that employs a neural network trained end-to-end with an EM-based mixture model. The approach integrates consistency optimization into the EM optimization (\textit{two-fold optimization}) to prevent overfitting to lower-level information in image clustering. The network is designed to model the cluster distributions, approximating each distribution for a batch, and thus the EM optimization is performed in batch-wise iterations. The forward pass through the network calculates the cluster likelihoods and posterior probabilities, and the backward pass backpropagates the EM-derived loss. 

\minititle{Leveraging traditional clustering techniques}

Certain methods focus on learning rich semantic representations for images to facilitate the clustering process. These image features can either be learned in an independent fashion through a pre=text task or simultaneously alongside clustering parameters as a single task in an end-to-end pipeline. Methods that rely in a pre-text task are generally designed to learn features that are amenable to clustering using conventional techniques such as k-means. For instance, \ffacronym{FineGAN} \cite{singh2019finegan} is a method based on the GAN framework proposed to learn to generate images of fine-grained object categories. Images can be hierarchically generated by disentangling the background, object shape, and object appearance. The key idea is to use information theory to associate each factor to a latent code and condition the relationships between the codes in a specific way to induce the desired hierarchy. \ffacronym{FineGAN} achieves the desired disentanglement and generates realistic and diverse images belonging to fine-grained classes of birds, dogs, and cars. Using \ffacronym{FineGAN}'s automatically learned features, real images can also be grouped with k-means clustering. 

\minititle{Other strategies}

There are other approaches that have been proposed to leverage the inherent properties of CNNs as a prior for clustering. Starting from the extracted representations, these methods refine cluster assignments iteratively by deriving supervisory signals. For example, Associative Deep Clustering \ffacronym{ADC} \cite{haeusser2019associative} embeds an image and its transformed version in a CNN and trains the network end-to-end by minimizing multiple complex losses. Deep Adaptive Clustering \ffacronym{DAC} \cite{chang2017deep} treats image clustering as a binary pairwise classification task and obtains the labels by leveraging the most confident samples. Other methods that follow the multi-stage approach have proposed to train a network to learn a soft clustering function by mining nearest neighbors from a feature space learned via a pre-task, therefore avoiding the need for a traditional clustering technique. One of such methods is \ffacronym{SCAN} \cite{vangansbeke2020scan}, which was introduced as an unsupervised image classification method. It combines self-supervised feature representations and learnable clustering in a two-step approach. In the first step, an adequate pretext task is constrained to produce features that ensure the minimal distance between input images and their augmentations, thus producing a meaningful feature representation space. Then the nearest neighbors of each image are mined based on feature similarity and used as a prior for clustering. The second step consists of a separate neural network that outputs encodings that represent soft clustering assignments. The overall algorithm also incorporates a fine-tuning approach with a self-labeling procedure to account for incorrect assignments due to noisy nearest neighbors. 

Another family of approaches seek to learn invariance to a set of transformations instead of learning representations. The main idea is to align images in pixel-space by predicting affine, colorimetric and morphological transformations to obtain more meaningful distances in pixel-space before clustering pixels. These \textit{transformation-invariant}
clustering frameworks jointly learn prototypes and transformations, and thus implement a \textit{prototype-based} clustering strategy. An example of such methods is \ffacronym{DTI-GMM} \cite{monnier2020deep} which optimized a single clustering objective over clustering parameters and deep image transformation modules.

\subsubsection{Common methodological choices in clustering methods for object discovery}

In this section, we analyze several key design decisions that characterize the methodology employed across various methods in a more specific manner, as shown in \hyperref[tab:overview_clust_spec]{Table~\ref{tab:overview_clust_spec}}. The first characteristic \textit{Pipeline} specifies the overall architectural setting of each method, providing insight into their structural organization in a broad sense. Following this, \textit{Clustering Type} is examined to determine the specific clustering strategy incorporated into each formulation, showcasing the diversity of approaches that have utilized. Additionally, \textit{Membership} refers to the mechanism that outlines how cluster assignments are implemented within each method.

\let\conftitle\footnotesize

\begin{table}[]
\centering
\resizebox{0.63\columnwidth}{!}{%
\begin{tabular}{lccccc}
\toprule
\multicolumn{6}{c}{\cellcolor{colorclust}{\textcolor{white}{\textbf{Clustering Methods}}}}                                 \\ \midrule
  Method &
  \begin{tabular}[c]{@{}c@{}}Pub.\end{tabular} &
  \begin{tabular}[c]{@{}c@{}}Year\end{tabular} &
  \begin{tabular}[c]{@{}c@{}}Pipeline\end{tabular} &
  \begin{tabular}[c]{@{}c@{}}clustering type\end{tabular} &
  \begin{tabular}[c]{@{}c@{}}Membership\end{tabular} \\ \midrule

DEC\cite{xie2016unsupervised} & \conftitle{ICML} & 2015 & End-to-end & Distribution-based & Student’s t-dist. \\

\rowcolor{rowbackground}
AAE\cite{makhzani2016adversarial}	& \conftitle{ARXIV} & 2016 & End-to-end & Soft-clustering & Softmax \\

JULE\cite{yang2016joint} & \conftitle{CVPR} & 2016 & End-to-end & Hierarchical & Cluster labels \\

\rowcolor{rowbackground}
DAC\cite{chang2017deep} & \conftitle{ICCV} & 2017 & End-to-end & Soft-clustering & Softmax \\

GMVAE\cite{dilokthanakul2017deep} & \conftitle{ICLR} & 2017 & End-to-end & Distribution-based & GMM \\

\rowcolor{rowbackground}
IMSAT\cite{hu2017learning} & \conftitle{ICML} & 2017 & End-to-end & Soft-clustering & Softmax \\

ADC\cite{haeusser2019associative}	& \conftitle{GCPR} & 2018 & End-to-end & Centroid-based & Softmax \\

\rowcolor{rowbackground}
FineGAN\cite{singh2019finegan} & \conftitle{CVPR} & 2019 & Multi-stage & Centroid-based & K-means \\

IIC\cite{ji2019invariant} & \conftitle{ICCV} & 2019 & End-to-end & Soft-clustering & Softmax \\

\rowcolor{rowbackground}
DTI-GMM	\cite{monnier2020deep}& \conftitle{NIPS} & 2020 & End-to-end & Centroid-based & Closest prototype \\

SCAN\cite{vangansbeke2020scan} & \conftitle{ECCV} & 2020 & Multi-stage & Soft-clustering & Softmax \\

\rowcolor{rowbackground}
TwoFoldOp\cite{tissera2022neural} & \conftitle{NEUROC} & 2022 & End-to-end & Distribution-based & Softmax \\

 \bottomrule

\end{tabular}%
}
\captionsetup{width=0.63\columnwidth}
\caption{Overview of clustering methods and common characteristics specific to this group of methods.}
\label{tab:overview_clust_spec}
\end{table}

\minititle{Architectural pipeline}

There are fundamentally two distinct approaches to feature learning in clustering methods. The first approach involves designing a single-stage architecture that is trained end-to-end to simultaneously learn feature representations and clustering assignments. Methods such as \ffacronym{DEC} \cite{xie2016unsupervised}, \ffacronym{AAE} \cite{makhzani2016adversarial}, \ffacronym{JULE} \cite{yang2016joint}, 	\ffacronym{DAC} \cite{chang2017deep}, \ffacronym{IIC} \cite{ji2019invariant} and \ffacronym{DTI-GMM} \cite{monnier2020deep} exemplify this approach. The second approach involves first learning a pre-text task that serves as a prior for clustering. Once the representations are learned, clustering can be performed on the feature space employing either traditional clustering techniques such as \ffacronym{FineGAN} \cite{singh2019finegan},
 or by optimizing a clustering objective to learn a soft assignment to clusters as in the case of \ffacronym{SCAN} \cite{vangansbeke2020scan}. 

\minititle{Type of clustering strategy}

A common approach to classifying clustering methods broadly categorizes them into three primary groups: distance-based, density-based, and connectivity-based techniques \cite{chang2017deep}. Distance-based methods, like K-means and agglomerative clustering, rely on distance metrics to establish a relationship between data points. Density-based methods seek to predict clusters by modeling a data distribution. In contrast, a third subgroup of approaches is connectivity-based and seek to assign data points to clusters based on their strong interconnectedness. Another typical classification found in the literature \cite{yang2016joint} separates clustering algorithms into hierarchical and partitional approaches. Hierarchical clustering, exemplified by agglomerative clustering, starts with numerous small clusters and progressively merges them. On the other hand, partitional clustering methods like K-means, aim to minimize a certain distance between data points and their closest cluster centers.

In this study, however, we refer to the clustering type in a slightly different manner. A first group of approaches, Distribution-based methods, such as \ffacronym{DEC} \cite{xie2016unsupervised}, \ffacronym{GMVAE} \cite{dilokthanakul2017deep} and \ffacronym{TwoFoldOp} \cite{tissera2022neural} perform a form of soft assignment by explicitly modeling the cluster distributions through a probability density function. The parameters of these functions are typically learned using neural networks. In contrast, a second group of methods including \ffacronym{AAE} \cite{makhzani2016adversarial}, \ffacronym{IIC} \cite{ji2019invariant}, \ffacronym{DAC} \cite{chang2017deep}, and \ffacronym{SCAN} \cite{vangansbeke2020scan}, perform Soft-Clustering by interpreting the output of a neural network as cluster probabilities, thus avoiding the need to explicitly model the cluster distributions. Subsequently, a third group of methods includes works such as \ffacronym{ADC} \cite{haeusser2019associative}, \ffacronym{FineGAN} \cite{singh2019finegan}, and \ffacronym{DTI-GMM} \cite{monnier2020deep}, which adopt a centroid-based clustering strategy. In this framework, clusters are represented by central points---called centroids or prototypes---and clusters are formed through the assignment to these centroids. Finally, another group of methods involves hierarchical clustering techniques, such as Agglomerative Hierarchical Clustering, which builds a hierarchy of clusters by recursively merging them until a stopping criterion is reached as in \ffacronym{JULE} \cite{yang2016joint}.

\minititle{Membership mechanism for cluster assignment}

As ilustrated in \hyperref[tab:overview_clust_spec]{Table~\ref{tab:overview_clust_spec}}, various subgroups emerge based on the mechanisms employed by different methods to assign output clusters to specific data points. Certain methods, such as \ffacronym{AAE} \cite{makhzani2016adversarial}, \ffacronym{DAC} \cite{chang2017deep}, \ffacronym{IIC} \cite{ji2019invariant}, and \ffacronym{SCAN} \cite{vangansbeke2020scan}), employ the \textit{softmax} function to determine the appropriate cluster assignment for an image. In contrast, other methods utilize more traditional techniques for this assignment, including \textit{Gaussian Mixture Models} as seen in \ffacronym{GMVAE} \cite{dilokthanakul2017deep}, \textit{K-means} clustering used in \ffacronym{FineGAN} \cite{singh2019finegan} , and the \textit{Student t-distribution} kernel applied in \ffacronym{DEC} \cite{xie2016unsupervised} , among others.

\textcolor{colorloc}{\subsection{Box-Localization Methods}}

It is possible to distinguish between two primary groups of bounding-box localization methods: bounding-box selection and bounding-box generation. The first group comprises methods that involve extracting candidate regions, referred to as object proposals, which may contain potential objects. The final output locations are subsequently  selected from these proposals. The latter group is composed of methods that do not discriminate among object proposals to establish the localization of objects but instead infer object locations by leveraging information from feature representations.

Bounding-box selection methods aim to determine which object proposals best depict objects from a pool of candidate regions extracted from the image collection. The selected object proposals are considered as the discovered objects. Thus, the localization quality of these methods is directly tied to the quality of the object proposals utilized. We have found two prominent families of techniques among these methods: techniques based on \textit{convex optimization} and based on \textit{probabilistic formulations}. Bounding-box Generation methods employ diverse techniques to construct a bounding box around discovered objects rather than relying on predefined object proposals obtained by third-party algorithms. They generally use deep features from pre-trained models to represent images and define mechanisms to determine which features correspond to foreground objects or the background. We have identified that, in general, these methods employ \textit{thresholds}, \textit{graph-based formulations}, or \textit{clustering techniques} to perform the separation of foreground and background features.

Next, we discuss these five commonly utilized families of techniques, providing exemplar methods and elaborating on their approaches. Subsequently, we discuss explore the typical methodological choices made in the design of these methods. An overview of these choices is presented in \hyperref[tab:overview_loc_spec]{Table~\ref{tab:overview_loc_spec}}.


\vspace{\baselineskip}
\subsubsection{Common families of Box-Localization methods for object discovery}

\begin{wrapfigure}{l}{0.6\textwidth}
    \begin{tikzpicture}[
    node/.style={rectangle, draw=none, rounded corners=6pt, align=center, fill=#1, drop shadow, text=white, minimum width=1.85cm, minimum height=1cm, font=\footnotesize},
    line/.style={-, ultra thick, color=colorloc},
    node distance=0.1cm
]
    \node[node=colorloc] (Root) at (0.95,0) {Box-Localization \\ common families};
    
    \node[node=colorloc] (Leaf1) at (-3,-1.75) {Convex \\ optimization};
    \node[node=colorloc, right=of Leaf1] (Leaf2)  {Probabilistic \\ formulations};
    \node[node=colorloc, right=of Leaf2] (Leaf3) {Threshold \\ based};
    \node[node=colorloc, right=of Leaf3] (Leaf4) {Clustering \\ based};
    \node[node=colorloc, right=of Leaf4] (Leaf5) {Graph-based \\ formulations};
    
    \draw[line] (Root.south) -- ++(0,-0.3) -| (Leaf1.north);
    \draw[line] (Root.south) -- ++(0,-0.3) -| (Leaf2.north);
    \draw[line] (Root.south) -- ++(0,-0.3) -| (Leaf3.north);
    \draw[line] (Root.south) -- ++(0,-0.3) -| (Leaf4.north);
    \draw[line] (Root.south) -- ++(0,-0.3) -| (Leaf5.north);
\end{tikzpicture}
    \caption{Families of box-localization methods in UOD}
    \label{fig:chart_fam_loc}
\end{wrapfigure}
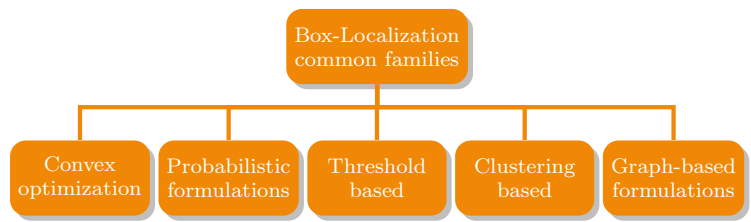

Several families of techniques are frequently employed in methods tackling object discovery through bounding-box localization, as ilustrated in \hyperref[fig:chart_fam_loc]{Fig.~\ref{fig:chart_fam_loc}}. In the subsequent paragraphs, we delineate these methodological families commonly employed for object localization, including convex optimization and probabilistic formulations, typically employed in methods that follow a bounding-box selection paradigm. Furthermore, we also describe families of techniques utilized  by methods that diverge from only selecting the best object proposals extracted via external procedures or off-the-shelf algorithms, which often integrate thesholding strategy, graph-based formulations, or clustering techniques. 

As is the case with other methodological categorizations within the four main groups of methods in unsupervised object discovery, these divisions of families are not inherently mutual exclusive. Methods classified within a particular family may well overlap with techniques from other families.

\minititle{Convex Optimization Methods}

A group of methods tackles this problem by formulating optimization-based approaches. \ffacronym{OSD} \cite{vo2019unsupervised} formulates an optimization problem to discover the implicit graph structure that an image dataset is assumed to exhibit. In such a graph, a link between two images exists when an image contains at least one object from a category depicted in the other image. The optimization problem is solved by first approximating a solution to a continuous relaxation by solving a max-flow problem. Then a solution to the original problem is found by applying block-coordinate ascent iterations. \ffacronym{rOSD} \cite{vo2020toward} extends the application of \cite{vo2019unsupervised} to larger datasets by proposing a two-stage approach: running first on small parts of the dataset to choose the most promising region proposals per image before running on the entire dataset in the second step. In \cite{vo2020toward}, a new approach to region proposal extraction is also introduced, as well as a new constraint to the previous formulation of \cite{vo2019unsupervised}, which acts as a regularizer and enables the method to perform localization of multiple objects per image. \ffacronym{LOD} \cite{vo2021large} re-formulates the combinatorial problem proposed in \cite{vo2019unsupervised} as a simpler graph-theoretical problem where region proposals correspond to nodes in a fully connected graph (instead of images as in \cite{vo2019unsupervised, vo2020toward}, and edge weights encode proposals’ similarity. Furthermore, this simpler formulation allows casting the problem of finding object-proposal nodes as a ranking problem, where the goal is to rank the nodes based on how well they are connected in the graph. This method combines two optimization problems: a quadratic formulation to solve the combinatorial problem and a ranking formulation based on the PageRank algorithm. In addition, the reformulation of \cite{vo2021large} makes it amenable to very large datasets (up to 1.7 million images). On the other hand, \ffacronym{CL-RWI} \cite{tang2014colocalization} introduced a method for co-localizing a common object in real-world images combining an image model and a box model into a joint optimization problem that can be relaxed to a convex quadratic program. Through the image model, the method is able to account for noise in a single-class dataset, which corresponds to images not containing the particular object being localized. In contrast, the box model addresses the object variability by localizing the common object in each image using rich correspondence information. Finally, \ffacronym{CL-FWA} \cite{joulin2014efficient} proposed a quadratic formulation and designs an image model that seeks to select the best bounding box in each image. Each box is represented by features based on the SIFT descriptor. In addition, the formulation incorporates a box similarity term based on a normalized Laplacian matrix and a discriminability clustering term which allows penalizing the selection of boxes whose features are not easily linearly separable from other boxes. 

\minititle{Probabilistic Formulations} 

A second group of methods tackle the bounding box selection problem with a probabilistic approach. \ffacronym{UODLW}\cite{cho2015unsupervised} proposes an iterative algorithm that alternates between retrieving image neighbors and localizing salient regions. The method uses HOG features to describe the object proposals extracted with the Randonmized Prim's algorithm. It employs a coordinate descent-style algorithm that combines a part-based probabilistic matching technique (Probablistic Hough Matching) and a foreground localization technique based on standout scores. The object is discovered by selecting the regions with top saliency in each image. \ffacronym{bMCL} \cite{zhu2014bMCL} casts the problem of object discovery as a multiple instance learning problem, using saliency scores to construct positive and negative bags from input images. In this method, learning is performed by a process that the authors call discriminative EM, in which MIL-Boost is a component for learning instance-level labels. \ffacronym{PatchNet} \cite{moon2021patchnet} discovers frequently appearing objects by training randomly sampled patches with self-supervision. The patch embeddings are learned by a variational autoencoder incorporating a contrastive loss modulated by color-based object saliency and a measure of background dissimilarity. Lastly, a clustering step identifies the frequently appearing objects in the images.

\minititle{Threshold-based}

Certain methods adopt a thresholding approach, where object regions are determined by applying thresholds to feature representations. This enables the separation of foreground from background pixels, facilitating object localization by identifying the bounding box containing the largest connected component of foreground pixels. A particular example of this approach is \ffacronym{DDT} \cite{wei2017unsupervised}, which showed that convolutional activations can play a role as a common object detector by leveraging the correlation between the features learned by a pre-trained CNN network. The method localizes objects in a single-class dataset by separating foreground and background using principal component analysis over an indicator matrix containing positive or negative values, which can reflect the deep descriptors' positive or negative correlations. Lastly, a bounding box that contains the largest connected component of foreground pixels is returned as the prediction. Another method is \ffacronym{OLM} \cite{zhang2020object}, which exploits data mining techniques and feature representations from pre-trained CNN models. The method extracts features from multiple convolutional layers and employs a tunable threshold to select the descriptors that are used to convert to transaction items. Then, frequent patterns from the transaction database are discovered through pattern mining techniques. Finally, possible objects are discovered by merging relevant, meaningful patterns. As well, \ffacronym{CL-CCF} \cite{le2017colocalization} is a two-step method that first finds an object category's category-consistent features (CCFs) and obtains a combined feature map containing aggregated CCF activations over the rough object region. Then, the CCF activations are co-propagated into a stable object using superpixel geodesic distances on the original images, producing an object likelihood map on which a threshold is applied to obtain the object region by placing a tight bounding box around the maximum coverage of the thresholded region. 

\minititle{Clustering}

There exist methods that seek to categorize image features into clusters to then select dominant clusters to represent the object within each image; or in the context of co-localization, across image sets. An exemplar of this approach is \ffacronym{UODSC} \cite{murasaki2019paper}, which proposed a method for multi-class co-localization based on spherical clustering. The method extracts deep features from a pre-trained CNN and classifies the features into several clusters. The main cluster indicating the foreground object is decided based on the intensities of the deep features belonging to each cluster. 

\minititle{Graph-based Formulations}

Another common approach involves leveraging graph structures to capture similarities among image features. Among these methods, \ffacronym{LOST} \cite{simeoni2021localizing} proposed to leverage features obtained from a visual transformer pre-trained with DINO self-supervision. Particularly they proposed to use the key component of the last attention layer to compute a patch similarity graph, in which two nodes (patches) are connected if their features are positively correlated. The authors empirically observe that patches of foreground objects are less correlated than patches corresponding to the background and thus propose to select the patch with the least number of similar patches (\textit{seed}) to localize a part of an object. Then a seed expansion strategy is performed to incorporate other patches highly correlated to the initial seed and thus likely to be part of the same object. Then an object mask is constructed by computing similarities of each patch to the seed patches, and a bounding box is produced from the largest connected component in the mask that contains the seed.  Similar to \cite{simeoni2021localizing}, \ffacronym{TokenCut} \cite{wang2022self} proposes constructing an undirected graph where nodes are patches and edges represent similarities between patches, whose features are extracted from the last self-attention layer of a pre-trained transformer. Salient objects are localized using a normalized graph cut to group self-similar regions and delimit the foreground objects. The authors propose to solve the graph-cut problem using spectral clustering with generalized eigendecomposition and show that the second smallest eigenvector provides a cutting solution indicating the likelihood that a token belongs to a foreground object. On the other hand, \ffacronym{CL-CSD} \cite{liCLCSD2016image} employs deep features from convolutional neural networks trained for image classification and explicitly exploits the single-class assumption to learn to co-localize objects by learning sparse confidence distributions, mimicking behaviors of supervised detectors. The common object is localized by performing segmentation on detection heat maps generated by computing the confidence scores of region proposals. Then the algorithm extracts superpixels and attempts to label them as foreground or background using a standard graph cut segmentation. \newline

\vspace{1.25cm} 
\subsubsection{Common methodological choices in box-localization methods for object discovery}

In this section, we examine various important methodological choices that define the approaches used across different methods more specifically, as illustrated in \hyperref[tab:overview_loc_spec]{Table~\ref{tab:overview_loc_spec}}. The first characteristic \textit{Clustering} specifies the type of clustering technique employed at any stage of the method's pipeline. The next characteristic, \textit{Saliency Priors}, refers to the techniques or third-party methods leveraged to extract salient regions from which bounding boxes can be derived. Then, the next characteristic is \textit{Box Priors}, which specifies techniques or methods used in the generation of traditional bounding-boxes (\textit{object proposals}). Furthermore, \textit{Box-Box Output} denotes the mechanism employed to generate the resulting bounding box used for evaluation.

\let\conftitle\footnotesize

\begin{table}[]
\centering
\resizebox{0.85\columnwidth}{!}{%

\begin{tabular}{@{}lcccccc@{}}
\toprule
\multicolumn{7}{c}{\cellcolor{colorloc}{\textcolor{white}{\textbf{\large Box-Localization Methods}}}}                                 \\ \midrule
  Method &
  \begin{tabular}[c]{@{}c@{}}Pub.\end{tabular} &
  \begin{tabular}[c]{@{}c@{}}Year\end{tabular} &
  \begin{tabular}[c]{@{}c@{}}Clustering\end{tabular} &
  \begin{tabular}[c]{@{}c@{}}Saliency prior\end{tabular} &
  \begin{tabular}[c]{@{}c@{}}Bounding-box prior\end{tabular} &
  \begin{tabular}[c]{@{}c@{}}Bounding-box output\end{tabular} \\ \midrule

bMCL \cite{zhu2014bMCL} & \conftitle{TPAMI} & 2014 & K-means init &  Bottom-up saliency & Segment-based composition & Labeled proposal \\
\rowcolor{rowbackground}
CL-FWA \cite{joulin2014efficient} & \conftitle{ECCV} & 2014 & Discriminative & Contrast-based method & Objectness proposals & Best candidate \\
CL-RWI \cite{tang2014colocalization} & \conftitle{CVPR} & 2014 & \EMPTY & Contrast-based method & Objectness proposals & Best candidate \\
\rowcolor{rowbackground}
UODLW \cite{cho2015unsupervised} & \conftitle{CVPR} & 2015 & \EMPTY & Hough space score & Randomized prim & Most stading-out region \\
CL-CSD \cite{li2016unsupervised} & \conftitle{ECCV} & 2016 & \EMPTY & Objectness score & Edgebox & Heatmap derived \\
\rowcolor{rowbackground}
CL-CCF \cite{le2017colocalization} & \conftitle{ICCV} & 2017 & K-means & CNN feat. map. & \EMPTY & Object-likelihood map \\
DDT \cite{wei2017unsupervised} & \conftitle{PATCOG} & 2018 & \EMPTY & First principal component & \EMPTY & CNN feat. map derived \\
\rowcolor{rowbackground}
OSD \cite{vo2019unsupervised} & \conftitle{CVPR} & 2019 & \EMPTY & Stand-out score  & Randomized prim & Ranked proposal \\
UOSDC \cite{murasaki2019paper} & \conftitle{MTA} & 2019 & Spherical & CNN feat. map. & \EMPTY & Highest intensity cluster \\
\rowcolor{rowbackground}
OLM \cite{zhang2020object} & \conftitle{TIP} & 2020 & \EMPTY & Pattern-based support map & \EMPTY & Support map derived \\
rOSD \cite{vo2020toward} & \conftitle{ECCV} & 2020 & \EMPTY & Stand-out score  & CNN-based & Ranked proposal \\
\rowcolor{rowbackground}
LOD \cite{vo2021large} & \conftitle{NIPS} & 2021 & \EMPTY & Prob. Hough matching & rOSD proposals & Ranked proposal \\
LOST \cite{simeoni2021localizing} & \conftitle{BMVC} & 2021 & \EMPTY & Tranformer feat. map & \EMPTY & Largest connected component \\
\rowcolor{rowbackground}
PatchNet \cite{moon2021patchnet} & \conftitle{ARXIV} & 2021 & K-means & Obj. score/Bg dissim. & Sampled patches & NMS scored patches \\
TokenCut \cite{wang2022self} & \conftitle{CVPR} & 2022 & Spectral & Tranformer feat. map & \EMPTY & Mask-based \\
\rowcolor{rowbackground}
MOST \cite{rambhatla2023MOST} & \conftitle{ICCV} & 2023 & DBSCAN & Tranformer feat. map & Entropy-based & Token pooling \\

 \bottomrule

\end{tabular}%
}
\captionsetup{width=0.85\columnwidth}
\caption{Overview of Box-Localization methods with common techniques and characteristics specific to this type of methods.}
\label{tab:overview_loc_spec}
\end{table}

\minititle{Clustering} 

Several box-localization methods integrate clustering techniques as part of their approach. Some methods, like \ffacronym{bMCL} \cite{zhu2014bMCL} and \ffacronym{CL-CCF} \cite{le2017colocalization}, use K-means clustering, while others like \ffacronym{UODSC} \cite{murasaki2019paper} employ spherical clustering. More recent approaches, such as \ffacronym{TokenCut} \cite{wang2022self}, utilize spectral clustering, and \ffacronym{MOST} \cite{rambhatla2023MOST} implements DBSCAN. Notably, as evidenced in \hyperref[tab:overview_loc_spec]{Table~\ref{tab:overview_loc_spec}}, this characteristic is not inherent to all localization methods.

\minititle{Saliency prior}

Saliency priors refer to the techniques these methods use in order to identify initial image regions that may correspond to objects; however, these location priors do not correspond exactly to the actual location of the objects, but rather indicate a belief that the contents of certain regions may correspond to objects. The approaches can vary from traditional computer vision techniques to more recent deep learning-based methods. Earlier methods such as \ffacronym{bMCL} \cite{zhu2014bMCL} and \ffacronym{CL-FWA} \cite{joulin2014efficient} use bottom-up saliency and contrast-based methods, respectively. As the field progressed, CNN feature maps became popular for saliency estimation, as employed by \ffacronym{CL-CCF} \cite{le2017colocalization} and \ffacronym{UODSC} \cite{murasaki2019paper}. More recent methods such as \ffacronym{LOST} \cite{simeoni2021localizing} and \ffacronym{TokenCut} \cite{wang2022self} leverage transformer feature maps to determine salient regions.

\minititle{Bounding-box prior}

This characteristic of localization methods refers to how they define the actual initial location of objects. These techniques correspond to the generation of object proposals, from which a final set is selected as localized objects. Early approaches like \ffacronym{CL-FWA} \cite{joulin2014efficient} and \ffacronym{CL-RWI} \cite{tang2014colocalization} used objectness proposals. Other methods employ techniques such as segment-based composition (\ffacronym{bMCL} \cite{zhu2014bMCL}) or randomized prim algorithm (\ffacronym{UODLW} \cite{cho2015unsupervised} and \ffacronym{OSD} \cite{vo2019unsupervised}). As evidenced, not all localization methods rely on explicit bounding-box priors or initial object proposals.

\minititle{Bounding-box output}

The final characteristic describes how each method generates the bounding-box outputs for evaluation. Approaches range from selecting the best candidate among proposals (\ffacronym{CL-FWA} \cite{joulin2014efficient}, \ffacronym{CL-RWI} \cite{tang2014colocalization}) to deriving boxes from heatmaps or feature maps (\ffacronym{CL-CSD} \cite{li2016unsupervised}, \ffacronym{DDT} \cite{wei2017unsupervised}). Other methods use techniques like object-likelihood maps (\ffacronym{CL-CCF} \cite{le2017colocalization}) or identify the largest connected component (\ffacronym{LOST} \cite{simeoni2021localizing}). Moreover, a group of methods (\ffacronym{OSD} \cite{vo2019unsupervised}, \ffacronym{rOSD} \cite{vo2020toward}, \ffacronym{LOD} \cite{vo2021large}) select the best bounding-boxes---initially extracted with a separate object proposal generation procedure---as a consequence of solving combinatorial or ranking problems. Recent methods such as \ffacronym{TokenCut} \cite{wang2022self} and \ffacronym{MOST} \cite{rambhatla2023MOST} use mask-based approaches and token pooling, respectively, to determine the final bounding box.

\vspace{-0.5cm} 
\textcolor{colorseg}{\subsection{Segmentation Methods}}

In the domain of segmentation methods for object discovery, a myriad of methodologies has been investigated, spanning from graph-based modeling to network-driven techniques and optimization-based strategies, among others. One group of methods leverage graph-based approaches \cite{zhu2014saliency, chang2015optimizing, lee2015multiple} which utilize graph structures to model relationships between image elements. Other methods employ MRF-based techniques \cite{rubio2012unsupervised, dai2013cosegmentation}, tree-structures \cite{yan2013hierarchical, li2018unsupervised}, spectral techniques \cite{melas-kyriazi2022deep, wang2022self}, pair matching \cite{shen2022learning}, constrained clustering \cite{tao2017image}, structured matrix decomposition \cite{rubio2012unsupervised} and dictionary-based saliency \cite{li2015weighted}. Similarly, other methods have involved traditional optimization techniques including semidefinite programs \cite{joulin2010discriminative}, score-based refinement \cite{faktor2013cosegmentation}, pixel correspondence optimization \cite{rubinstein2013unsupervised}, mixed-integer programs \cite{wang2017multiple}, non-negative factorization \cite{collins2018deep}, among others as described in \hyperref[tab:overview_all]{Table~\ref{tab:overview_all}}. Conversely, a different group of methods involve the use of neural networks to perform segmentation tasks and are typically formulated as saliency networks \cite{zhang2017supervision, zhang2018deep, nguyen2021deepusps, shin2022unsupervised}, segmentation networks \cite{zhao2020unsupervised, voynov2021object, choudhury2022guess, huang2022segdiscover, shen2022learning}, generative adversarial networks \cite{arandjelovic2019object, chen2019unsupervised} and pixel-embedding learning \cite{vangansbeke2021unsupervised, cho2021picie}. On the other hand, optimization schemes that have been employed include low-rank optimization  \cite{joulin2010discriminative}, submodular optimization \cite{gunheekim2011distributed}, block-coordinate descent \cite{rubinstein2013unsupervised, joulin2012multiclass}, least-squares optimization \cite{zhu2014saliency}, K-means-like optimization \cite{tao2017image}, particle Swarm optimization \cite{shoitan2021unsupervised} and EM-like procedures \cite{yu2021DRC}. In contrast, graph-based methods generally employ graph cuts \cite{rubio2012unsupervised, dai2013cosegmentation} and normalized cuts \cite{melas-kyriazi2022deep, wang2022self}, but have also utilized belief propagation  \cite{yan2013hierarchical}, random walkers \cite{lee2015multiple}), and the Felzenszwalb algorithm \cite{peng2017salient}. As well, several methods have employed Grabcut \cite{faktor2013cosegmentation, jerripothula2016image, shen2022learning}, while others rely on deep learning techniques optimized via backpropagation \cite{zhang2017supervision, arandjelovic2019object, huang2022segdiscover, ji2019invariant, hsu2019deepco3}.
\newline

\subsubsection{Common families of Segmentation methods for object discovery}

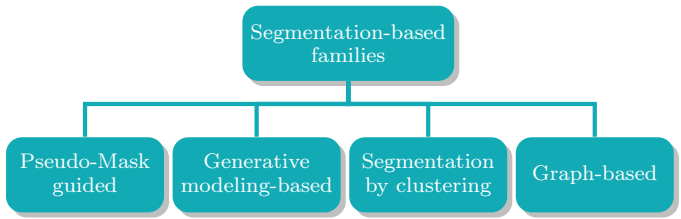
\begin{wrapfigure}{l}{0.55\textwidth}
    \begin{tikzpicture}[
    node/.style={rectangle, draw=none, rounded corners=6pt, align=center, fill=#1, drop shadow, text=white, minimum width=2.1cm, minimum height=1cm, font=\footnotesize},
    line/.style={-, ultra thick, color=colorseg},
    node distance=0.1cm
]
    \node[node=colorseg] (Root) at (0.5,0) {Segmentation-based \\ families};
    
    \node[node=colorseg] (Leaf1) at (-3,-1.75) {Pseudo-Mask \\ guided};
    \node[node=colorseg, right=of Leaf1] (Leaf2)  {Generative \\ modeling-based};
    \node[node=colorseg, right=of Leaf2] (Leaf3) {Segmentation \\ by clustering};
    \node[node=colorseg, right=of Leaf3] (Leaf4) {Graph-based};
    
    \draw[line] (Root.south) -- ++(0,-0.3) -| (Leaf1.north);
    \draw[line] (Root.south) -- ++(0,-0.3) -| (Leaf2.north);
    \draw[line] (Root.south) -- ++(0,-0.3) -| (Leaf3.north);
    \draw[line] (Root.south) -- ++(0,-0.3) -| (Leaf4.north);
\end{tikzpicture}
    \caption{Families of segmentation methods in UOD}
    \label{fig:chart_fam_seg}
\end{wrapfigure}

Following the trend from clustering and box-localization methods, several families of techniques have been explored by segmentation methods, as illustrated in \hyperref[fig:chart_fam_seg]{Fig.~\ref{fig:chart_fam_seg}}. A line of work implements a neural network to predict segmentation masks for objects in an image. These methods are usually trained in a self-supervised fashion by leveraging pseudo-masks as the learning signal. Therefore, diverse mechanisms have been explored in order to generate the necessary pseudo-masks. For example, \cite{zhang2018deep} and \cite{nguyen2021deepusps} are ensemble methods that obtain the pseudo masks from multiple traditional hand-crafted methods. \cite{shin2022unsupervised} clusters dense features from a pre-trained transformer-based encoder to generate the pseudo-masks. \cite{huang2022segdiscover} uses a self-supervised convolutional encoder on superpixels to generate representations that are then clustered to create pseudo-masks. Other methods leverage different sources to generate the learning signal. For example, \ffacronym{UVISAM} \cite{choudhury2022guess} and \ffacronym{UnsupOD} \cite{zhao2020unsupervised} use temporal cues such as optical flow or inter-frame structure, while \cite{voynov2021object} leverages the latent spaces learned by a GAN to generate synthetic images and masks for training. Another line of work leverages generative image modeling. \cite{arandjelovic2019object} produces an object mask with a GAN-based generator, which is then used to composite a realistic image by merging the generated sample with a suitable background obtained from a different image. \cite{yu2021DRC} uses an energy-based loss along with an image generator, and \cite{voynov2021object} leverages a direction in the latent space of a pre-trained GAN to segment foreground and background regions. In the following, some of these representative ideas are extended. 


\minititle{Pseudo-Mask guided segmentation}

In this family of methods, initial estimations of object masks (pseudo-labels) are first generated using auxiliary information such as coarse segmentation masks, temporal information or simple object detectors. These pseudo-labels serve as surrogate ground truth annotations and are then used to train a segmentation network in a \textit{supervised} manner. By leveraging pseudo-labels, these methods can effectively harness large amounts of unlabeled data for training segmentation networks. Additionally, by incorporating pseudo-labels during training, these methods can enhance the robustness and generalization capabilities of segmentation networks. One of such methods is \ffacronym{SelfMask} \cite{shin2022unsupervised}. The authors of \ffacronym{SelfMask} observed that self-supervised features display a greater potential for object segmentation when spectral clustering is used instead of k-means. Therefore, the method consists of extracting foreground regions among multiple masks, generated by applying spectral clustering on various types of image features from pre-trained networks and using a winner-takes-all voting strategy for selecting the salient masks. Then an object segmentation model is trained using the selected masks as pseudo-masks. Another method in this family is \ffacronym{SegDiscover} \cite{huang2022segdiscover} which proposes to discover concepts in images via semantic segmentation. The method starts by decomposing images into small patches using traditional superpixel algorithms. The patches represent regions containing potential visual concepts (\textit{concept primitives}) and are merged with neighboring patches using a primitive-merging approach. In the second step, the concept primitives are fed into a self-supervised encoder whose feature representations are clustered through an overclustering with k-means and a cluster reassignment algorithm based on spectral clustering to produce discovered concepts. Then, in the third step, the clusters are used as pseudo-segmentation labels to train a segmentation network. Likewise, \ffacronym{E-BigBiGAN} \cite{voynov2021object} utilizes the synthetic images produced by an off-the-shelf GAN whose latent space can be manipulated to allow distinguishing pixels that belong to the object or background in the generated images. These masks are used as saliency pseudo-masks to train a discriminative U-Net model as an object segmentation network. Another method, \ffacronym{UVISAM} \cite{choudhury2022guess}, proposed an approach that exploits motion and objectness for segmenting visual objects in images. The authors propose to train an image segmentation network to predict regions that contain simple optical flow patterns, using motion anticipation as a learning signal. The method uses MaskFormer as the segmentation network with Swin-tiny transformer, pre-trained on ImageNet with self-supervision, as the backbone. The optical flow is estimated following the practice of MotionGrouping. Lastly, \ffacronym{CoAttnCNN} \cite{hsu2018coattention} proposes a model composed of two CNN modules that work collaboratively to generate common object segmentation. The first module, called the generator, is a co-attention saliency map generator that uses a pool of object proposals obtained with an object proposal generation algorithm and trained with a mask loss to regularize the generator removing false negatives on objects and false positives on the background. The second module is a pre-trained CNN used as a feature extractor to compute a co-attention loss that guides the training of the generator, enabling it to produce maps that correctly localize the common object. 

\minititle{Generative modeling-based methods}

This family of methods may leverage certain aspects of generative methods as part of their formulation for the segmentation problem. Typically, methods based on generative modeling leverage probabilistic models to partition the image into meaningful regions. In this family, \ffacronym{DRC} \cite{yu2021DRC} proposes to use an energy-based prior and generative image modeling in the form of a Mixture of Experts (MoE) to perform foreground extraction. The method generates foreground and background regions using sampled latent variables with a generator for each region. DRC iteratively samples the latent variables with a two-step workflow in an EM-like algorithm. The E-step employs gradient-based MCMC sampling to infer the latent variables. In the M-step, the sampled latent variables are fed into the model for image generation, minimizing a reconstruction loss to update the generators. As well, \ffacronym{CP-GAN} \cite{arandjelovic2019object} leverages the GAN framework discriminatively to segment out an existing object in an image. In this method, the generator learns to discover an object in one image by compositing it into another such that the discriminator is not able to determine if the composed image is fake. The generator network is a U-Net which produces a pixel-wise segmentation mask and a seediness score, which allows picking a pixel as a seed. The discriminator network is also U-Net-based, where the middle encoding is used to determine the real or fakeness decision.



\minititle{Segmentation by clustering}

A group of methods primarily employ clustering techniques as the core of the problem formulation. These methods treat segmentation as a clustering problem, where the goal is to group pixels---or pixel embeddings---into meaningful clusters that represent objects. One particular method is \ffacronym{MaskContrast} \cite{vangansbeke2021unsupervised} which proposes a two-step framework to learn discriminative pixel embeddings by incorporating object mask proposals, obtained with a saliency estimator, as a mid-level prior in a contrastive optimization objective. Pixels can then be directly clustered with k-means to perform unsupervised semantic segmentation. Another method in this group is \ffacronym{IIC} \cite{ji2019invariant} which is a general-purpose clustering algorithm trained end-to-end contrastively by maximizing mutual information between pairs of related data samples. In its setting for semantic segmentation, the algorithm clusters patches instead of entire images by extending the image augmentations to account for local spatial invariance. This spatial invariance is achieved by pairing each patch with transformed neighboring patches, which are defined by small spatial displacements. The segmentation objective maximizes information between patches, in expectation over the batch of images, patches within each image, and the set of transformations. 

\minititle{Graph-based segmentation}

This family includes methods such as \ffacronym{CS-ECEM} \cite{hongliangli2014unsupervised}, \ffacronym{CS-GSI} \cite{li2018unsupervised}, \ffacronym{CS-OHS} \cite{shoitan2021unsupervised}, \ffacronym{DMFC} \cite{chang2015optimizing}, \ffacronym{DSM} \cite{melas-kyriazi2022deep} and \ffacronym{TokenCut} \cite{wang2022self}. These methods typically represent an image with a graph formulation, where pixels are nodes and edges represent relationships. One recurrent technique utilized in methods like these to solve graph formulations is normalized cuts. \ffacronym{Ncut} aims to find a partition of the graph that minimizes the dissimilarities between clusters while maximizing the similarities within each cluster. As well, methods like \ffacronym{MCC} \cite{joulin2012multiclass}, \ffacronym{SelfMask} \cite{shin2022unsupervised} and \ffacronym{TokenCut} \cite{wang2022self} employ spectral clustering as part of their formulation, which involves clustering the graph nodes based on the eigenvectors of the Laplacian matrix derived from the graph.  This type of clustering is effective for identifying non-convex and irregularly shaped regions in images. Other methods such as \ffacronym{GMS} \cite{jerripothula2014automatic}, \ffacronym{SCF} \cite{jerripothula2016image}, \ffacronym{NPG} \cite{wang2017multiple}, \ffacronym{CS-GSI} \cite{li2018unsupervised}, and \ffacronym{SegSwap} \cite{shen2022learning} are based on GrabCut, which combines graph cuts and Gaussian mixture models to iteratively refine an initial segmentation of an image typically obtained by extracting superpixels.

\let\conftitle\footnotesize

\begin{table}[]
\centering
\resizebox{0.9\columnwidth}{!}{%

\begin{tabular}{lcccccc}
\toprule
\multicolumn{7}{c}{\cellcolor{colorseg}{\textcolor{white}{\textbf{Segmentation Methods}}}}                                 \\ \midrule
  Method &
  \begin{tabular}[c]{@{}c@{}}Pub.\end{tabular} &
  \begin{tabular}[c]{@{}c@{}}Year\end{tabular} &
  \begin{tabular}[c]{@{}c@{}}External saliency\end{tabular} &
  \begin{tabular}[c]{@{}c@{}}Oversegmentation\end{tabular} &
  \begin{tabular}[c]{@{}c@{}}Clustering\end{tabular} &
  \begin{tabular}[c]{@{}c@{}}Mask\\Generation\end{tabular} \\ \midrule

DC \cite{joulin2010discriminative} & \conftitle{ICML} & 2010 & \EMPTY & Watershed & Discrim./Spectral & Threshold \\

\rowcolor{rowbackground}
CoSand \cite{gunheekim2011distributed}	& \conftitle{ICCV} & 2011 & \EMPTY & Superpixels & Spectral & Superpixel label assign. \\	

IFBM \cite{rubio2012unsupervised} & \conftitle{CVPR} & 2012 & \EMPTY & Mean-shift & GMM/Spectral & Pixel labeling \\

\rowcolor{rowbackground}
MCC	\cite{joulin2012multiclass} & \conftitle{CVPR} & 2012 & \EMPTY & Superpixels & Discrim./Spectral & Superpixel label assign. \\	

COMP \cite{faktor2013cosegmentation} & \conftitle{ICCV} & 2013 & Likelihood maps & \EMPTY & \EMPTY & Superpixel label assign. \\

\rowcolor{rowbackground}
CST \cite{dai2013cosegmentation} & \conftitle{ICCV} & 2013 & \EMPTY & \EMPTY & \EMPTY & Pixel labeling \\

HS \cite{yan2013hierarchical} & \conftitle{CVPR} & 2013 & \EMPTY & Watershed & \EMPTY & Saliency fusion \\

\rowcolor{rowbackground}
OD	\cite{rubinstein2013unsupervised} & \conftitle{CVPR} & 2013 & Contrast-based saliency & \EMPTY & \EMPTY & Refinement \\

GMS	\cite{jerripothula2014automatic} & \conftitle{ICIP} & 2014 & Contrast-based saliency & Superpixels & K-means & Superpixel label assign. \\

\rowcolor{rowbackground}
CS-ECEM	\cite{hongliangli2014unsupervised} & \conftitle{TCSVT} & 2014 & Objectness estimation & Superpixels & Affinity propagation & Superpixel label assign. \\

SMD	\cite{peng2017salient} & \conftitle{TPAMI} & 2017 & Sparse/Dense reconstr. & Superpixels & \EMPTY & Context propag. \\

\rowcolor{rowbackground}
wCTR \cite{zhu2014saliency} & \conftitle{CVPR} & 2014 & Contrast-based saliency & Superpixels & \EMPTY & Superpixel label assign. \\

DMFC \cite{chang2015optimizing} & \conftitle{CVIU} & 2015 & Probability-based saliency & Superpixels & GMM & Threshold \\

\rowcolor{rowbackground}
MRW \cite{lee2015multiple} & \conftitle{CVPR} & 2015 & \EMPTY & Superpixels & MRW-based & Bilateral filter \\

WSC \cite{li2015weighted} & \conftitle{CVPR} & 2015 & \EMPTY & Superpixels & \EMPTY & Threshold \\

\rowcolor{rowbackground}
SCF \cite{jerripothula2016image} & \conftitle{TMM} & 2016 & Multiple methods & Superpixels & K-means & Threshold \\

SBF \cite{zhang2017supervision} & \conftitle{ICCV} & 2017 & Multiple methods & Superpixels & \EMPTY & CNN \\	

\rowcolor{rowbackground}
NPG \cite{wang2017multiple} & \conftitle{TIP} & 2017 & Segmentation map & Hierarchical mrg. & \EMPTY & Label assign. \\

SGC3 \cite{tao2017image} & \conftitle{AAAI} & 2017 & Adaptive threshold & Superpixels & Cosine sim. based & Cluster assign. \\

\rowcolor{rowbackground}
CoAttCNN \cite{hsu2018coattention} & \conftitle{IJCA} & 2018 & CNN feature map & \EMPTY & \EMPTY & CRF \\

CS-GSI \cite{li2018unsupervised} & \conftitle{NEUCOM} & 2018 & Saliency tree & Gpb-owt-ucm & K-means & Label assign. \\

\rowcolor{rowbackground}
DFF \cite{collins2018deep} & \conftitle{ECCV} & 2018 & CNN feature map & \EMPTY & NMF & Threshold \\

USDNL \cite{zhang2018deep} & \conftitle{CVPR} & 2018 & Multiple methods & \EMPTY & \EMPTY & CNN \\

\rowcolor{rowbackground}
CP-GAN \cite{arandjelovic2019object} & \conftitle{ARXIV} & 2019 & \EMPTY & \EMPTY & \EMPTY & Composition \\

DeepCO3 \cite{hsu2019deepco3} & \conftitle{CVPR} & 2019 & SBF & \EMPTY & \EMPTY & Proposal ranking \\

\rowcolor{rowbackground}
IIC \cite{ji2019invariant} & \conftitle{ICCV} & 2019 & \EMPTY & Overclustering & Soft clustering & Cluster assign. \\

ReDo \cite{chen2019unsupervised} & \conftitle{NIPS} & 2019 & \EMPTY & \EMPTY & \EMPTY & Composition \\

\rowcolor{rowbackground}
DeepUSPS \cite{nguyen2021deepusps} & \conftitle{NIPS} & 2019 & Handcrafted saliency & \EMPTY & \EMPTY & CRF \\

UnsupOD \cite{zhao2020unsupervised} & \conftitle{SYMM.} & 2020 & \EMPTY & \EMPTY & \EMPTY & CNN \\

\rowcolor{rowbackground}
CS-OHS \cite{shoitan2021unsupervised} & \conftitle{JCS} & 2021 & DL saliency method & Superpixels & \EMPTY & Superpixel label assign. \\	

DRC \cite{yu2021DRC} & \conftitle{NIPS} & 2021 & \EMPTY & \EMPTY & \EMPTY & Pixel labeling \\

\rowcolor{rowbackground}
E-BigBigGAN \cite{voynov2021object} & \conftitle{PMLR} & 2021 & Latent space guided & \EMPTY & \EMPTY & CNN \\

MaskContrast \cite{vangansbeke2021unsupervised} & \conftitle{ICCV} & 2021 & DL saliency method & \EMPTY & K-means & Cluster assign. \\

\rowcolor{rowbackground}
PiCIE \cite{cho2021picie} & \conftitle{CVPR} & 2021 & \EMPTY & \EMPTY & K-means & Cluster assign. \\

UVISAM \cite{choudhury2022guess} & \conftitle{BMVC} & 2022 & \EMPTY & \EMPTY & \EMPTY & CNN \\

\rowcolor{rowbackground}
DSM \cite{melas-kyriazi2022deep} & \conftitle{CVPR} & 2022 & Tranformer Feat. map & Overclustering & Spectral/k-means & CRF \\

MOVE \cite{bielski2022MOVE} & \conftitle{NIPS} & 2022 & Tranformer Feat. map & \EMPTY & \EMPTY & CNN \\

\rowcolor{rowbackground}
SegDiscover \cite{huang2022segdiscover} & \conftitle{ARXIV} & 2022 & \EMPTY & Superpixels & K-means/Spectral & CNN \\

SegSwap \cite{shen2022learning} & \conftitle{CVPR} & 2022 & LOST/bilateral solver & \EMPTY & K-means & Label assign. \\

\rowcolor{rowbackground}
SelfMask \cite{shin2022unsupervised} & \conftitle{CVPRw} & 2022 & CNN/Transf. feat. maps & \EMPTY & Spectral & Threshold \\

TokenCut \cite{wang2022self} & \conftitle{CVPR} & 2022 & Tranformer Feat. map & \EMPTY & Spectral & Threshold \\

\rowcolor{rowbackground}
CutLER \cite{wang2023CUTLER} & \conftitle{CVPR} & 2023 & Tranformer Feat. map & \EMPTY & \EMPTY & CNN \\

SEMPART \cite{ravindran2023SEMPART} & \conftitle{ICCV} & 2023 & Tranformer Feat. map & \EMPTY & \EMPTY & Sal. refinement \\

\rowcolor{rowbackground}
U2Seg \cite{niu2023u2seg} & \conftitle{CVPR} & 2024 & CutLer/Stego masks & \EMPTY & K-means & CNN  \\

 \bottomrule

\end{tabular}%
}
\captionsetup{width=0.9\columnwidth}
\caption{Overview of Segmentation methods and common characteristics shared among these methods.}
\label{tab:overview_seg_spec}
\end{table}


\subsubsection{Common methodological choices in segmentation methods for object discovery}

Segmentation methods typically rely in certain sub-steps as part of their formulation. Such common methodological decisions are comprised in \hyperref[tab:overview_seg_spec]{Table~\ref{tab:overview_seg_spec}}. A frequently encountered choice refers to third-party techniques or methods used to extract initial saliency masks. Another choice determines the type of clustering technique used, while the type of over-segmentation technique is also a common one. Additionally, all segmentation methods need to produce a segmentation mask to isolate the discovered object; thus, the mask generation mechanism employed to produce the output segmentation mask has to be determined as well.

\minititle{External saliency techniques}

In general, a common assumption contemplated by many segmentation methods is that foreground pixels representing an object of interest should be salient. This assumption implies that pixels pertaining to an object must be dissimilar from other pixels within the image, and also that they should appear sparsely distributed, resembling their nearest neighbors albeit with potential variations in color, size, and position \cite{rubinstein2013unsupervised}. Many methods rely in techniques external to the method's formulation to facilitate an initial saliency estimation. Some methods like \cite{faktor2013cosegmentation} utilize likelihood maps, which assess the likelihood of pixels belonging to salient regions. Another prominent group \cite{rubinstein2013unsupervised, jerripothula2014automatic, zhu2014saliency} leverage contrast-based saliency, where salient regions are identified based on differences in color, intensity, or texture compared to their surroundings. Other methods \cite{hongliangli2014unsupervised} may use objectness estimation which prioritize regions likely to contain objects of interest has been used, or probability-based saliency \cite{chang2015optimizing} which assign saliency scores based on statistical probabilities. Furthermore, there are methods \cite{zhang2018deep, zhang2017supervision} that combine multiple techniques for saliency, for instance, leveraging both handcrafted and deep learning (DL) saliency mechanisms. DL saliency methods operate in the latent space guided by neural networks, often utilizing CNN \cite{hsu2018coattention, collins2018deep} or Transformer feature maps \cite{melas-kyriazi2022deep, wang2022self} to capture salient information. Transformer feature maps, in particular, have gained attention for their effectiveness in saliency detection tasks. Additionally, there are methods such as \cite{peng2017salient} that incorporate sparse/dense reconstruction techniques. Some methods \cite{nguyen2021deepusps} employ explicit handcrafted features for saliency, while others leverage learned features from DL models. Finally, another group of methods \cite{hsu2019deepco3, shen2022learning} employ previous methods to obtain initial saliency estimation. In a way, these methods focus on applying saliency enhancement to the tasks of object discovery.

\minititle{Oversegmentation techniques}

Another differentiating factor between methods can be found in the use of oversegmentation techniques. One category encompasses methods \cite{joulin2010discriminative, yan2013hierarchical} that utilize watershed segmentation, where an image is partitioned into distinct regions by simulating a flooding process from predefined markers. Another numerous group \cite{joulin2012multiclass, jerripothula2014automatic, hongliangli2014unsupervised, zhu2014saliency, chang2015optimizing, lee2015multiple, li2015weighted, jerripothula2016image, peng2017salient, zhang2017supervision, tao2017image, shoitan2021unsupervised} relies on superpixels generation algorithms, which divide the image into perceptually uniform regions, facilitating subsequent processing steps. Additionally, there are methods \cite{ji2019invariant, melas-kyriazi2022deep} that focus on overclustering, intentionally generating a larger number of clusters than necessary to achieve finer granularity in segmentation results. Techniques such as mean-shift clustering \cite{rubio2012unsupervised} and hierarchical merging \cite{wang2017multiple} have also been explored to refine the oversegmentation process. Lastly, particular techniques such as The gpb-owt-ucm method have also been employed \cite{li2018unsupervised}, combining gradient-based segmentation, optimal thresholding, and ultrametric contour maps to produce comprehensive oversegmentation results.

\minititle{Clustering techniques}

Certain methods utilize  specific clustering techniques as a component of the segmentation pipeline. One prevalent category includes methods \cite{joulin2010discriminative, joulin2012multiclass} that utilize discriminative or spectral clustering approaches, which aim to partition the data into clusters based on similarity or affinity matrices derived from pairwise comparisons. Other methods such as \cite{hongliangli2014unsupervised} have leveraged affinity propagation which identifies exemplars within the data and assigns each data point to one of these exemplars based on similarity measures. Similarly, cosine similarity-based clustering \cite{hongliangli2014unsupervised} utilizes cosine similarity measures to quantify the similarity between data points and cluster them accordingly. Another group encompasses methods \cite{rubio2012unsupervised, chang2015optimizing} that utilize Gaussian Mixture Models (GMM) alone or in conjunction with spectral clustering techniques. GMM-based methods model the data distribution as a mixture of Gaussian distributions while spectral clustering partitions the data based on eigenvectors of a similarity matrix. The latter strategy has been greatly explored \cite{gunheekim2011distributed, melas-kyriazi2022deep, huang2022segdiscover, shin2022unsupervised, wang2022self}. Furthermore, there are methods \cite{jerripothula2014automatic, jerripothula2016image, li2018unsupervised, vangansbeke2021unsupervised, cho2021picie, niu2023unsupervised} that rely on traditional k-means clustering, which partitions the data into a predetermined number of clusters by iteratively assigning data points to the nearest cluster centroid. Soft clustering methods assign data points to multiple clusters with varying degrees of membership and are also typically utlized \cite{ji2019invariant}. Non-negative matrix factorization (NMF) is another technique employed for clustering \cite{collins2018deep}, which factorizes the data matrix into non-negative components to discover underlying patterns. Additionally, spectral clustering combined with k-means \cite{melas-kyriazi2022deep, huang2022segdiscover} has also been explored.

\minititle{Mask Generation techniques}

Every segmentation method aims to produce output masks to isolate the discovered objects. The mask generation mechanism referenced in \hyperref[tab:overview_seg_spec]{Table~\ref{tab:overview_seg_spec}} specifically denotes the final stage in a method's pipeline, immediately preceding the production of the output for evaluation. These mechanisms vary from simple Thresholding \cite{chang2015optimizing, li2015weighted, collins2018deep, shin2022unsupervised, wang2022self}---\textit{i.e.} applying a threshold to determine object membership---to decoder networks \cite{bielski2022MOVE, wang2023CUTLER, niu2023u2seg} that leverage learned representations to produce detailed segmentation masks. Furthermore, some methods \cite{faktor2013cosegmentation, jerripothula2014automatic, zhu2014saliency, jerripothula2016image, tao2017image, shen2022learning} can employ clustering algorithms while other methods \cite{ravindran2023SEMPART, yan2013hierarchical} refine a saliency map obtained throughout the pipeline. Other methods \cite{hsu2018coattention, nguyen2021deepusps, melas-kyriazi2022deep} have used CRF and bilateral filters \cite{lee2015multiple}.

\textcolor{colordec}{\subsection{Scene Decomposition Methods}}
\subsubsection{Common families of Decomposition methods for object discovery}

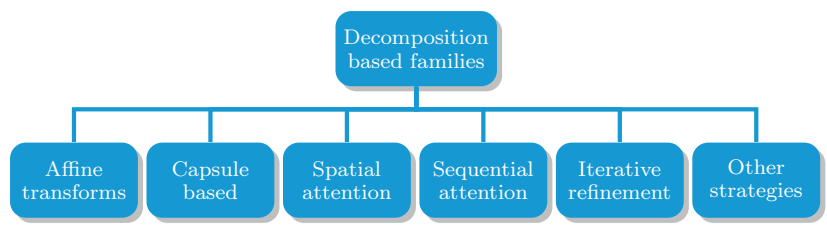
\begin{wrapfigure}{l}{0.65\textwidth}
    \begin{tikzpicture}[
    node/.style={rectangle, draw=none, rounded corners=6pt, align=center, fill=#1, drop shadow, text=white, minimum width=1.7cm, minimum height=1cm, font=\footnotesize},
    line/.style={-, ultra thick, color=colordec},
    node distance=0.1cm
]
    \node[node=colordec] (Root) at (1.55,0) {Decomposition \\ based families};
    
    \node[node=colordec] (Leaf1) at (-3,-1.75) {Affine \\ transforms};
    \node[node=colordec, right=of Leaf1] (Leaf2)  {Capsule\\based};
    \node[node=colordec, right=of Leaf2] (Leaf3) {Spatial \\ attention};
    \node[node=colordec, right=of Leaf3] (Leaf4) {Sequential \\ attention};
    \node[node=colordec, right=of Leaf4] (Leaf5) {Iterative \\ refinement};
    \node[node=colordec, right=of Leaf5] (Leaf6) {Other \\ strategies};
    
    \draw[line] (Root.south) -- ++(0,-0.3) -| (Leaf1.north);
    \draw[line] (Root.south) -- ++(0,-0.3) -| (Leaf2.north);
    \draw[line] (Root.south) -- ++(0,-0.3) -| (Leaf3.north);
    \draw[line] (Root.south) -- ++(0,-0.3) -| (Leaf4.north);
    \draw[line] (Root.south) -- ++(0,-0.3) -| (Leaf5.north);
    \draw[line] (Root.south) -- ++(0,-0.3) -| (Leaf6.north);
\end{tikzpicture}
    \caption{Families of Decomposition methods in UOD}
    \label{fig:chart_fam_dec}
\end{wrapfigure}

Several methods have leveraged structured latent variable models to discover objects in images in an unsupervised manner. As observed in \hyperref[fig:chart_fam_dec]{Fig.~\ref{fig:chart_fam_dec}}, we have identified certain families of techniques that are frequently employed by these methods. These families include two types of sprite-learning methods one that learns sets of transformations and another that employs capsule networks to represent objects. Additionally, a line of methods has leveraged the VAE framework the Variational Autoencoder (VAE) framework to address the routing of objects into latent representations. Within these approaches, three primary families of techniques have arisen: spatial attention based on Spatial Transformer Networks (\textit{STNs}), sequential attention based on Recurrent Neural Networks (\textit{RNNs}), and iterative refinement methods that leverage iterative amortized inference (\textit{IAI}).

\minititle{Transformations-based}

There exists a class of methods aimed at learning object prototypes, templates, or parts, commonly known as \textit{sprites}, which serve as a basis for inferring image components. Typically, these methods rely on learning a set of transformations enabling the representation of objects as transformed versions of the prototypes or a combination of them. A particular method in this class is \ffacronym{DTISprites} \cite{monnier2021unsupervised}, which proposed a model to learn a layered decomposition of images, where each layer is a transformed instance of a prototypical object. The model jointly learns sprites, transformations, and occlusions. The sprites are modeled as transformations of a set of prototypical and are mapped through geometric and colorimetric transformations to generate object layers, which are used to assemble the reconstructed image. A further instance of this class is \ffacronym{MarioNette} \cite{smirnov2021marionette}, which jointly learns a grid-based anchor system and a textured-patch dictionary to extract an image representation based on transformations of sprites. The 2D sprite dictionary is meant to capture recurring visual elements in a collection of images (extracted from a sprite-based video game). An encoder explains frames (images) as compositions of potentially transparent sprites locally transformed and anchored on a regular grid. The representations learned by the model can be used in downstream tasks like editing or analysis. 

\minititle{Capsule-based Methods}

Capsule-based methods seek to represent objects as capsules. The notion of a capsule represents groups of neurons that encode various properties of an object, such as pose, scale, and deformation. Methods based on capsules aim to overcome certain limitations of CNNs by explicitly modeling hierarchical relationships between object parts. The Spatial Capsule Autoencoder \ffacronym{SCAE} \cite{kosiorek2019stacked} extends the capsule network architecture to unsupervised learning tasks. It utilizes a convolutional encoder to extract features and a capsule layer to encode the spatial relationships between the features. SCAE learns to reconstruct an image by iteratively updating pose parameters of capsules. 



\minititle{Methods based on Spatial Attention}

The primary concept underlying this group of methods is to predict a rectangular window, often referred to as a \textit{glimpse}, which likely contains an object. Typically, this prediction is implemented by leveraging Spatial Transformer Networks (\textit{STNs}). Each glimpse is then transformed into an object representation by learning inference and generative models based on latent variables. Finally, the reconstruction of an image is done by positioning the decoded glimpses onto a canvas. A particular method within this group is \ffacronym{AIR} \cite{eslami2016attend}. The authors proposed to learn a set of latent variables in an iterative procedure. The model infers latent variables per object for location, appearance, and presence (existence) by implementing amortized inference through a recurrent neural network that discovers the objects sequentially. Then a decoder reconstructs the image by generating each object, decoding them into attention glimpses that a spatial transformer tries to place correctly onto a blank canvas. 

\minititle{Methods based on Sequential Attention}

The principle guiding this category of methods involves utilizing sequential attention mechanisms rather than spatial attention or \textit{glimpses}, to identify objects within an image. In these methods, a recurrent neural network is commonly employed to systematically scan the image, focusing attention on different regions at each time. Unlike spatial attention methods, which predict a rectangular window, sequential attention methods dynamically locate object regions as the network iterates through the image. This iterative process allows the network to capture the spatial layout of objects in a more flexible and adaptive manner. Among these methods is \ffacronym{MONet} \cite{burgess2019monet}, which proposed a method to model scenes compositionally. The model is a variational autoencoder implemented with a recurrent segmentation network that attends to individual objects sequentially. At each step, the masking network segments a scene component from the current scope mask---the yet unexplained portion of the image. Then, a convolutional encoder receives both the image and the current attention mask as input to infer independent latent representations following a pixel-wise gaussian posterior distribution. \ffacronym{GENESIS} \cite{engelcke2020genesis} is another latent-variable model, which differs from similar methods such as \cite{burgess2019monet, greff2020multiobject}, in that it seeks to capture relationships between components. In this approach, the scene is also modeled as a Gaussian mixture with corresponding mixing probabilities, and latent representations are learned with an autoregressive prior. In addition, the model is trained by minimizing a Generalized ELBO with Constrained Optimization (\textit{GECO}) instead of the traditional \textit{ELBO} maximization. Finally, although the latent variables can be decoded in parallel, the spatial dependencies are still captured by an autoregressive prior implemented by a recurrent neural network, which means that inference is inherently sequential. 

\minititle{Iterative Refinement}

A particular instance among these methods is \ffacronym{EfficientMORL} \cite{emami2021efficient}, which seeks to show that iterative amortized inference (\textit{IAI}) can be used to learn object-centric representations while being as efficient as other methods without sacrificing representation quality---which is based on incorporating fundamental biases to induce representations to be symmetric, unordered, and disentangled. The approach uses a two-stage inference algorithm to decompose an image. In the first stage, a multi-layer hierarchical variational autoencoder (\textit{HVAE}) extracts symmetric and disentangled representations. In the second stage, a lightweight refinement network refines the HVAE posterior through iterative inference. Finally, the generative network employs the usual spatial broadcast decoder to obtain component reconstructions and assignment masks. Another important method within this group is \ffacronym{IODINE} \cite{greff2020multiobject}, a multi-object representation learning approach that leverages variational inference to learn the generative and inference models jointly. It defines a multi-slot representation where each slot shares an underlying format for representing objects. The object representations are inferred as a set of latent variables with the help of a refinement network that implements iterative amortized inference. The generative network models an image as a spatial mixture where each component corresponds to a single object. The decoder structure includes a broadcast decoder where the objects' latent variables are decoded separately into pixel-wise means and mask-logits. 

\minititle{Other techniques}

Certain other methods present interesting approaches that may not necessarily belong to a specific family of methods, or there may only be a few instances of similar methods. That is exemplified by \ffacronym{Slot Attention} \cite{locatello2020objectcentric}, which is a differentiable module that can interface between low-level representations and higher-level task-dependent abstractions, such as a set-structured latent space, and that can be used to decompose an image into object layers. The module leverages an iterative attention mechanism with a procedure similar to soft k-means clustering. In the application for unsupervised object discovery, representations from a CNN encoder are passed to the slot attention module, which maps it into multiple slots in an iterative refinement procedure. A cross-attention mechanism between input and slots at each iteration seeks to allow each slot to attend to a specific region. The slots are then decoded individually by a spatial broadcast decoder into image components for the final image reconstruction. Another interesting idea was explored in \ffacronym{uORF} \cite{yu2021uORF}. The authors proposed a method based on object radiance fields to learn an object-centric representation for 3D scene decomposition. The underlying scene is modeled from a single image as a composition of object radiance fields. An encoder learns a set of latent representations through a Slot Attention module \cite{locatello2020objectcentric}, and a decoder implemented by conditional NeRF networks \cite{mildenhall2020nerf} learns to represent the 3D objects. During training, multiple views of the scene are generated to leverage supervision in pixel space through an optimization objective that consists of reconstruction, perceptual and adversarial losses.

\subsubsection{Common methodological choices in decomposition methods for object discovery}

The design of decomposition methods needs to consider several key considerations that are certainly intrinsic to this type. A commonly encountered consideration is the \textit{Attention Strategy}, which determines the basic localization mechanism employed for scanning objects: either through rectangular regions (glimpse) or pixel-wise (mask). Another crucial aspect is the \textit{Object Routing} mechanism, which dictates the approach to be followed regarding nference of latent object representations. Lastly, \textit{Reconstruction} refers to the mechanism by which the image is composed or reconstructed from the learned latent representations. An overview of this methodological considerations can be found in \hyperref[tab:overview_dec_spec]{Table~\ref{tab:overview_dec_spec}}.

\minititle{Attention Strategy}

A distinguishing characteristic among decomposition methods is the attention strategy they employ to find the location of objects. These methods typically employ two primary strategies: glimpse-based and mask-based. However, not all methods explicitly define one of these strategies. Glimpse-based approaches \cite{eslami2016attend, huang2016efficient, kosiorek2018sequential,  yuangenerative, crawford2019spatially, stelznerfaster, jiang2021generative, jiang2020scalor, lin2020space, deng2021generative} learn to extract rectangular regions corresponding to objects by leveraging the learnable transformations with spatial transformer networks. Conversely, Mask-based strategies \cite{greff2016tagger, greff2017neural, vansteenkiste2018relational, greff2020multiobject, burgess2019monet, vonkugelgen2020causal, engelcke2020genesis, yang2020learning, nanbo2021learning, locatello2020objectcentric, huang2020better, emami2021efficient, engelcke2022genesisv2}, learn masks that isolate objects---background, layers, etc.--- in recurrent or iterative processes. Other methods \cite{kosiorek2019stacked, monnier2021unsupervised, smirnov2021marionette, yu2022unsupervised} do not use explicit attention strategies to define image regions for learning latent representations. Instead, they first identify common patterns or prototypes in the entire dataset and then reconstruct images by combining transformed prototypes.

\minititle{Object Routing}

Another important property refers to the strategy used to systematically traverse the set of potential objects in an image. We refer to this property as object routing, which commonly encompasses three strategies: sequential, iterative and template-based. Sequential approaches \cite{eslami2016attend, huang2016efficient, kosiorek2018sequential, yuangenerative, crawford2019spatially, stelznerfaster, jiang2021generative, jiang2020scalor, lin2020space, deng2021generative}, process objects one at a time; that is, the model focuses on explaining one object in the scene at each step, based on its understanding of preceding objects. In contrast, methods that employ Iterative routing \cite{greff2017neural, vansteenkiste2018relational, greff2020multiobject, nanbo2021learning, locatello2020objectcentric, huang2020better, emami2021efficient, engelcke2022genesisv2, seitzer2022bridging}
refine latent representations of objects---\textit{i.e.} explain them---over multiple steps of simultaneous refinement. Finally, other methods \cite{kosiorek2019stacked, monnier2021unsupervised, smirnov2021marionette} do not explicitly route regions into latent object representations. Instead, they employ a template-based routing strategy by learning common templates that can be transformed and combined to reconstruct images.

\minititle{Reconstruction}

The reconstruction techniques refer to the approaches used to recover an image for evaluation purposes from the learned object representations. Two techniques have gained prominence over the years: Spatial transformer networks STNs \cite{jaderberg2016spatial}---as utilized in studies \cite{eslami2016attend, huang2016efficient, yuangenerative, crawford2019spatially, jiang2020scalor, lin2020space, monnier2021unsupervised, zhu2021gmair, deng2021generative, smirnov2021marionette}---and Spatial broadcast decoder networks \cite{watters2019spatial}, which have been employed in works \cite{greff2020multiobject, burgess2019monet, vonkugelgen2020causal, engelcke2020genesis, yang2020learning, nanbo2021learning, locatello2020objectcentric, emami2021efficient, engelcke2022genesisv2}. Nonetheless, other techniques have also been employed such as: CNN decoders \cite{greff2017neural, vansteenkiste2018relational}, Sub-pixel CNNs \cite{kosiorek2018sequential}, Transpose convolutions \cite{huang2020better}, Radiance fields \cite{yu2022unsupervised}, and other similar approaches to STNs for Affine transforms of templates \cite{kosiorek2019stacked}.

\let\conftitle\footnotesize

\begin{table}[htpb]
\centering
\resizebox{0.65\columnwidth}{!}{%
\begin{tabular}{lccccc}


\toprule
\multicolumn{6}{c}{{\cellcolor{colordec}\textcolor{white}{{\textbf{Decomposition Methods}}}}} \\

\midrule
  Method &
  \begin{tabular}[c]{@{}c@{}}Pub.\end{tabular} &
  \begin{tabular}[c]{@{}c@{}}Year\end{tabular} &
  \begin{tabular}[c]{@{}c@{}}Attention\\Strategy\end{tabular} &
  \begin{tabular}[c]{@{}c@{}}Object\\Routing\end{tabular} &
  \begin{tabular}[c]{@{}c@{}}Reconstruction\end{tabular} \\ \midrule

AIR \cite{eslami2016attend}	& \conftitle{NIPS} & 2016 & Glimpse-based	& Sequential & STN-based \\
\rowcolor{rowbackground}
CST-VAE \cite{huang2016efficient}	& \conftitle{ICLR} 	& 2016	& Glimpse-based	& Sequential & STN-based \\
TAGGER \cite{greff2016tagger} & \conftitle{NIPS} & 2016 & Mask-based & Iterative & Fully Conntected \\
\rowcolor{rowbackground}
N-EM \cite{greff2017neural} & \conftitle{NIPS} & 2017 & Mask-based & Iterative & CNN decoder \\
R-NEM \cite{vansteenkiste2018relational} & \conftitle{ICLR} & 2018 & Mask-based & Iterative & CNN decoder \\
\rowcolor{rowbackground}
SQAIR \cite{kosiorek2018sequential} & \conftitle{NIPS} & 2018 & Glimpse-based & Sequential & Subpixel-CNN \\
GMIO \cite{yuangenerative} & \conftitle{ICML} & 2019 & Glimpse-based & Sequential & STN-based \\
\rowcolor{rowbackground}
IODINE \cite{greff2020multiobject} & \conftitle{ICML} & 2019 & Mask-based & Iterative & Spatial broadcast \\
MONet \cite{burgess2019monet} & \conftitle{ARXIV} & 2019 & Mask-based & Sequential & Spatial broadcast \\
\rowcolor{rowbackground}
SCAE \cite{kosiorek2019stacked} & \conftitle{NIPS} & 2019 & \EMPTY & Template-based & Affine-transforms \\
SPAIR \cite{crawford2019spatially} & \conftitle{AAAI} & 2019 & Glimpse-based & Sequential & STN-based \\
\rowcolor{rowbackground}
SuPAIR \cite{stelznerfaster} & \conftitle{ICML} & 2019 & Glimpse-based & Sequential & STN-based \\
ECON \cite{vonkugelgen2020causal} & \conftitle{ICLR} & 2020 & Mask-based & Sequential & Spatial broadcast \\
\rowcolor{rowbackground}
GENESIS \cite{engelcke2020genesis} & \conftitle{ICLR} & 2020 & Mask-based & Sequential & Spatial broadcast \\
GNM \cite{jiang2021generative} & \conftitle{NIPS} & 2020 & Glimpse-based & Sequential & STN-based \\
\rowcolor{rowbackground}
LMIO \cite{yang2020learning} & \conftitle{CVPR} & 2020 & Mask-based & Sequential & Spatial broadcast \\
MulMON \cite{nanbo2021learning} & \conftitle{NIPS} & 2020 & Mask-based & Iterative & Spatial broadcast \\
\rowcolor{rowbackground}
SCALOR \cite{jiang2020scalor} & \conftitle{ICLR} & 2020 & Glimpse-based & Sequential & STN-based \\ 
SLOT ATTN. \cite{locatello2020objectcentric} & \conftitle{NIPS} & 2020 & Mask-based & Iterative & Spatial broadcast \\
\rowcolor{rowbackground}
SPACE \cite{lin2020space} & \conftitle{ICLR} & 2020 & Glimpse-based & Sequential & STN-based \\
SRN \cite{huang2020better} & \conftitle{NIPS} & 2020 & Mask-based & Iterative & TransposeConvs \\
\rowcolor{rowbackground}
DTISprites \cite{monnier2021unsupervised} & \conftitle{ICCV} & 2021 & \EMPTY & Template-based & STN-based \\
EfficientMORL \cite{emami2021efficient} & \conftitle{ICML} & 2021 & Mask-based & Iterative & Spatial broadcast \\
\rowcolor{rowbackground}
GENESISv2 \cite{engelcke2022genesisv2} & \conftitle{NIPS} & 2021 & Mask-based & Iterative & Spatial broadcast \\
GMAIR \cite{zhu2021gmair} & \conftitle{CIN} & 2021 & Glimpse-based & Sequential & STN-based \\
\rowcolor{rowbackground}
GSNGs \cite{deng2021generative} & \conftitle{ICLR} & 2021 & Glimpse-based & Sequential & STN-based \\
MarioNette \cite{smirnov2021marionette} & \conftitle{NIPS} & 2021 & \EMPTY & Template-based & STN-based \\
\rowcolor{rowbackground}
uORF \cite{yu2022unsupervised} & \conftitle{ICLR} & 2022 & \EMPTY & Iterative & Radiance fields \\
DINOSAUR \cite{seitzer2022bridging} & \conftitle{ICLR} & 2023 & Mask-based & Iterative & Spatial broadcast \\

 \bottomrule

\end{tabular}%
}
\captionsetup{width=0.65\columnwidth}
\caption{Overview of methods showing differences in techniques and characteristics specific to Decomposition.}
\label{tab:overview_dec_spec}
\end{table}

\begin{table}[]
\centering

\resizebox{0.9\linewidth}{!}{%
\begin{tabular}{llllllllllllll} 
\toprule
\multicolumn{14}{c}{\cellcolor[HTML]{DDDDDD}{\MakeUppercase{Acronyms of the Methods in this Survey}}} \\ 
\midrule

\begin{tabular}[c]{@{}l@{}}
1\\2\\3\\4\\5\\6\\7\\8\\9\\10\\11\\12\\13\\14\\15\\16\\17\\18\\19\\20\\21\\22
\end{tabular} & 

\begin{tabular}[c]{@{}l@{}}
AAE \cite{makhzani2016adversarial}\\
ADC \cite{haeusser2019associative}\\
AIR \cite{eslami2016attend}\\
bMCL \cite{zhu2014bMCL}\\
CL-CCF \cite{le2017colocalization} \tsb\\
CL-CSD \cite{li2016unsupervised} \tsb\\
CL-FWA \cite{joulin2014efficient} \tsb\\
CL-RWI \cite{tang2014colocalization} \tsb\\
CoAttnCNN \cite{hsu2018coattention} \tsb\\
COMP \cite{faktor2013cosegmentation}\\
CoSand \cite{gunheekim2011distributed}\\
CP-GAN \cite{arandjelovic2019object} \tsb\\
CS-OHS \cite{shoitan2021unsupervised} \tsb\\
CS-ECEM \cite{hongliangli2014unsupervised} \tsb\\
CS-GSI \cite{li2018unsupervised} \tsb\\
CST \cite{dai2013cosegmentation}\\
CST-VAE \cite{huang2016efficient}\\
CutLER \cite{wang2023CUTLER} \\
DAC \cite{chang2017deep}\\
DC \cite{joulin2010discriminative}\\
DDT \cite{wei2017unsupervised}\\
DEC \cite{xie2016unsupervised}
\end{tabular} &  & 

\begin{tabular}[c]{@{}l@{}}
23\\24\\25\\26\\27\\28\\29\\30\\31\\32\\33\\34\\35\\36\\37\\38\\39\\40\\41\\42\\43\\44
\end{tabular} & 

\begin{tabular}[c]{@{}l@{}}

DeepCO3 \cite{hsu2019deepco3}\\
DeepUSPS \cite{nguyen2021deepusps}\\
DFF \cite{collins2018deep}\\
DINOSAUR \cite{seitzer2022bridging} \\
DMFC \cite{chang2015optimizing}\\
DPR-CAE \cite{xiang2021dprcae}\\
DRC \cite{yu2021DRC}\\
DSM \cite{melas-kyriazi2022deep}\\
DTI \cite{monnier2020deep}\\
DTISprites \cite{monnier2021unsupervised}\\
E-BigBiGAN \cite{voynov2021object}\\
ECON \cite{vonkugelgen2020causal}\\
EfficientMORL \cite{emami2021efficient}\\
FineGAN \cite{singh2019finegan}\\
GENESIS \cite{engelcke2020genesis}\\
GenesisV2 \cite{engelcke2022genesisv2}\\
GMAIR \cite{zhu2021gmair}\\
GMIO \cite{yuangenerative} \tsb\\
GMS \cite{jerripothula2014automatic}\\
GMVAE \cite{dilokthanakul2017deep}\\
GNM \cite{jiang2021generative}\\
GS \cite{weigeodesic}
\end{tabular} &  & 

\begin{tabular}[c]{@{}l@{}}
45\\46\\47\\48\\49\\50\\51\\52\\53\\54\\55\\56\\57\\58\\59\\60\\61\\62\\63\\64\\65\\66
\end{tabular} & 

\begin{tabular}[c]{@{}l@{}}
GSGNs \cite{deng2021generative}\\
HS \cite{yan2013hierarchical}\\
IFBM \cite{rubio2012unsupervised}\\
IIC \cite{ji2019invariant}\\
IMSAT \cite{hu2017learning}\\
INCT \cite{li2016unsupervised}\\
IODINE \cite{greff2020multiobject}\\
JULE \cite{yang2016joint}\\
LMIO \cite{yang2020learning} \tsb\\
LOD \cite{vo2021large}\\
LOST \cite{simeoni2021localizing}\\
MarioNette \cite{smirnov2021marionette}\\
MaskContrast \cite{vangansbeke2021unsupervised}\\
MCC \cite{joulin2012multiclass}\\
MONet \cite{burgess2019monet}\\
MOST \cite{rambhatla2023MOST}\\
MOVE \cite{bielski2022MOVE} \\
MRW \cite{lee2015multiple}\\
MulMON \cite{nanbo2021learning}\\
N-EM \cite{greff2017neural}\\
NPG \cite{wang2017multiple}\\
OD \cite{rubinstein2013unsupervised}
\end{tabular} &  & 

\begin{tabular}[c]{@{}l@{}}
67\\68\\69\\70\\71\\72\\73\\74\\75\\76\\77\\78\\79\\80\\81\\82\\83\\84\\85\\86\\87\\88
\end{tabular} & 

\begin{tabular}[c]{@{}l@{}}

Odin \cite{henaff2022object}\\
OLM \cite{zhang2020object}\\
OSD \cite{vo2019unsupervised}\\
PatchNet \cite{moon2021patchnet}\\
PiCIE \cite{cho2021picie}\\
PSGNet \cite{bear2020learning}\\
RC \cite{chengglobal}\\
ReDO \cite{chen2019unsupervised}\\
R-NEM \cite{vansteenkiste2018relational}\\
rOSD \cite{vo2020toward}\\
SBF \cite{zhang2017supervision}\\
SCAE \cite{kosiorek2019stacked}\\
SCALOR \cite{jiang2020scalor}\\
SCAN \cite{vangansbeke2020scan}\\
SCF \cite{jerripothula2016image}\\
SegDiscover \cite{huang2022segdiscover}\\
SegSwap \cite{shen2022learning}\\
SelfMask \cite{shin2022unsupervised}\\
SEMPART \cite{ravindran2023SEMPART}\\
SGC3 \cite{tao2017image}\\
Slot Attn \cite{locatello2020objectcentric}\\
SMD \cite{peng2017salient}
\end{tabular} &  & 

\begin{tabular}[c]{@{}l@{}}
89\\90\\91\\92\\93\\94\\95\\96\\97\\98\\99\\100\\101\\102\\103\\104\\105\\106\\ \\ \\ \\ \\  
\end{tabular} & 

\begin{tabular}[c]{@{}l@{}}

SPACE \cite{lin2020space}\\
SPAIR \cite{crawford2019spatially}\\
SQAIR \cite{kosiorek2018sequential}\\
SRN \cite{huang2020better}\\
StampNet \cite{visser2019stampnet}\\
SuPAIR \cite{stelznerfaster}\\
Tagger \cite{greff2016tagger}\\
TokenCut \cite{wang2022self}\\
TwofoldOp \cite{tissera2022neural} \tsb\\
UnsupOD \cite{zhao2020unsupervised}\\
uORF \cite{yu2022unsupervised}\\
wCTR \cite{zhu2014saliency}\\
WSC \cite{li2015weighted}\\
U2Seg \cite{niu2023u2seg}\\
UODLW \cite{cho2015unsupervised} \tsb\\
UODSC \cite{murasaki2019paper} \tsb\\
USDNL \cite{zhang2018deep} \tsb\\
UVISAM \cite{choudhury2022guess} \tsb
\\
\\
\\
\\
\\
\end{tabular} \\

\bottomrule
\end{tabular}
}
\captionsetup{width=0.9\columnwidth}
\caption{List of methods in alphabetical order. The majority of these names and acronyms have been coined by the original authors. In some cases, the acronyms were suggested in subsequent works or survey papers. For the methods that lacked names (\tsb), we applied straightforward naming criteria to suggest appropriate acronyms. The prefixes CL and CS refer to co-localization and co-segmentation, respectively.}
\label{tab:method_index}
\end{table}

\section{Experimental Settings}
\label{sec:experimental}
\newcommand{\titledataset}{\footnotesize}

In this section, we begin by presenting a comprehensive compendium of the datasets frequently used in the experiments and evaluations conducted by the methods covered in this survey. Then, we describe the commonly employed metrics, including their formulations. Finally, we present several comparisons among these methods in order to provide an overview of the progression of research in unsupervised object discovery.

It is important to note that comparing this extensive collection of methods is challenging because many among them do not adhere to the same evaluation protocols. Instead, methods often establish their particular setups and experimental settings. This can lead to variations in how subsets of datasets are selected and how metrics are computed. We will elaborate on this topic across sections \hyperref[subsec:datasets]{Sec.~\ref{subsec:datasets}} and \hyperref[subsec:metrics]{Sec.~\ref{subsec:metrics}}.

In \hyperref[tab:method_index]{Table~\ref{tab:method_index}}, we present an exhaustive catalog of all the methods examined in this survey, featuring their acronyms or dubbed names, as well as the corresponding citations for easy reference and exploration.

\subsection{Common Datasets}
\label{subsec:datasets}

In the following subsections, we describe the datasets commonly used for evaluations within each of the four method categories discussed in this survey: clustering, localization, segmentation, and decomposition.

\subsubsection{Datasets in Clustering Methods} 

In \hyperref[tab:datasets_clust]{Table~\ref{tab:datasets_clust}}, we highlight several widely used datasets in clustering methods. Among these datasets we include: MNIST\cite{lecun1998gradient}, CIFAR-10\cite{krizhevsky2009learning}, CIFAR-100\cite{krizhevsky2009learning}, CUB-200-2011\cite{wah2011caltech}, STL-10\cite{coates2011analysis}, and SVHN\cite{netzer2011reading} . A further description of each of these datasets is subsequently provided.

\begin{table}[ht!]
\centering
\resizebox{0.85\columnwidth}{!}{%
\begin{tabular}{@{}llccccc@{}}
    \toprule
    \multicolumn{7}{c}{\cellcolor{colorclust}{\textcolor{white}{\textbf{Clustering Methods}}}}                                 \\ \midrule
    
    \textbf{Dataset} & 
    \textbf{Description} & 
    \begin{tabular}[c]{@{}c@{}}\textbf{Image} \\ \textbf{Dimensions}\end{tabular} &
    \begin{tabular}[c]{@{}c@{}}\textbf{Training} \\ \textbf{Images}\end{tabular} &
    \begin{tabular}[c]{@{}c@{}}\textbf{Testing} \\ \textbf{Images}\end{tabular} &
    \begin{tabular}[c]{@{}c@{}}\textbf{Unlabeled} \\ \textbf{Images}\end{tabular} &
    \textbf{\# of Classes} \\
    \midrule
    MNIST & Handwritten digits & 28x28 & 60,000 & 10,000 & \EMPTY & 10 \\
    CIFAR-10 & 10 classes of objects & 32x32 & 50,000 & 10,000 & \EMPTY & 10 \\
    CIFAR-100 & 100 classes grouped into 20 superclasses & 32x32 & 50,000 & 10,000 & \EMPTY & 100 \\
    CUB-200-2011 & 200 bird species & variable resolution & 6,033 & 5,755 & \EMPTY & 200 \\
    STL-10 & 10 object classes from ImageNet & 96x96 & 5,000 & 8,000 & 100,000 & 10 \\
    SVHN & Street View House Numbers & 32x32 & 73,257 & 26,032 & 531,131 & 10 \\
    \bottomrule
\end{tabular}
}
\captionsetup{width=0.65\columnwidth}
\caption{Overview of datasets commonly used in clustering methods.}
\label{tab:datasets_clust}
\end{table}

{\titledataset \textbf{MNIST.}} 
MNIST\cite{lecun1998gradient} is a well-known dataset of handwritten digits containing 70,000 grayscale images with a resolution of 28x28 pixels. The training set comprises 60,000 images, while the test set contains 10,000 images. \newline

{\titledataset \textbf{CIFAR-10 \& CIFAR-100.}}
The CIFAR-10\cite{krizhevsky2009learning} dataset contains 60,000 color images of dimension 32x32 divided into ten classes, with 6,000 images per class. It contains image-level labels, and the classes included in the dataset are, in alphabetical order: airplane, automobile, bird, cat, deer, dog, frog, horse, ship, and truck. The dataset is divided into a training set with 50,000 images and a testing set with 10,000 images. Each image in the dataset has a corresponding label, indicating the class to which it belongs. \newline

The dataset CIFAR-100\cite{krizhevsky2009learning} also contains 60,000 images of 32x32 but allocates 600 images to each of the 100 classes included. Furthermore, the classes in this dataset are grouped into 20 superclasses, each containing five classes. The superclasses are aquatic mammals, fish, flowers, food containers, fruit, household electrical devices, household furniture, insects, large carnivores, large herbivores, medium-sized mammals, musical instruments, non-insect invertebrates, people, reptiles, small mammals, trees, vehicles 1 and vehicles 2. Finally, this dataset is also split into training and testing sets like CIFAR-10, with 50,000 training and 10,000 testing images. \newline

{\titledataset \textbf{CUB.}}
The CUB-200-2011\cite{wah2011caltech} dataset contains 11,788 images of 200 bird species. It includes 6,033 in the training set and 5,755 images for testing. The images are in color and have a variable resolution. The CUB-200-2011 is an extended version of CUB-200 with roughly double the number of images per category and added annotations for parts localization. \newline

{\titledataset \textbf{STL-10.}}
The STL-10\cite{coates2011analysis} dataset is inspired by CIFAR-10 but with fewer labeled images and a higher resolution. The dataset provides 13,000 labeled images divided into a training set with 5,000 images and a testing set with 8,000 images. The images in this dataset are extracted from ImageNet, are downsampled to 96x96px, and correspond to the ten object classes: airplane, bird, car, cat, deer, dog, horse, monkey, ship, and truck. In addition, the STL-10 dataset includes 100,000 unlabeled images extracted from a broader distribution than the training set. \newline

{\titledataset \textbf{SVHN.}} 
The SVHN\cite{netzer2011reading} dataset (Street View House Numbers) contains around 600,000 images of house numbers taken from Google Street View. The images are in color and are provided in two formats: original images with variable resolution and character level bounding boxes and cropped images centered around single digits with a resolution of 32x32 pixels, in a similar style to MNIST. The dataset contains 73,257 labeled images for training and 26,032 labeled images for testing. An additional 531,131 unlabeled images are provided to use as extra training data.

\subsubsection{Datasets in Localization Methods}
\label{sec:datasets_loc}
In \hyperref[tab:datasets_loc]{Table~\ref{tab:datasets_loc}}, we highlight several widely used datasets in localization methods, including the datasets: Object Discovery, PASCAL VOC 2007, PASCAL VOC 2012, and MS COCO 20k. A detailed description of each dataset is provided below.

\begin{table}[h]
\centering
\resizebox{0.85\columnwidth}{!}{%
\begin{tabular}{@{}llccc@{}}
    \toprule
    \multicolumn{5}{c}{\cellcolor{colorloc}{\textcolor{white}{\textbf{Datasets in Localization Methods}}}}                                 \\ \midrule
    
    \textbf{Dataset} & 
    \textbf{Description} & 
    \begin{tabular}[c]{@{}c@{}}\textbf{Total} \\ \textbf{Images}\end{tabular} &
    \begin{tabular}[c]{@{}c@{}}\textbf{Used} \\ \textbf{Images}\end{tabular} &
    \textbf{\# of objects} \\
    \midrule
    Object Discovery & Internet dataset with classes: car, horse, airplane & 300 & 300 & - \\
    VOC 2007 & 9,963 images of 20 object classes & 9,963 & 3,550 & 6,661 \\
    VOC 2012 & 11,530 images in train/val data & 11,530 & 7,838 & 13,957 \\
    COCO 20k & Subset of COCO 2014 trainval, filtered for no crowd instances & 20,000 & 19,817 & 143,951 \\

    \bottomrule
\end{tabular}
}
\captionsetup{width=0.65\columnwidth}
\caption{Overview of datasets commonly used in localization methods.}
\label{tab:datasets_loc}
\end{table}
{\titledataset \textbf{OBJECT DISCOVERY.}}
The Object Discovery\cite{Rubinstein13Unsupervised} dataset, also called the \textit{Internet dataset}, was collected using the Bing API to automatically download images for three queries, each of which represents one class: car (4,347 images), horse (6,381 images), and airplane (4,542 images). Since the images were downloaded from the internet with a search engine, this collection mechanism introduced noise in the search results; therefore, each class contains images without an object of the respective class. Consequently, only a subset of this dataset has been used to enforce fair comparisons among methods. This subset contains 100 images per class, with 18\% outliers in the airplane category, 11\% in the car category, and 7\% in the horse category. \newline

{\titledataset \textbf{PASCAL VOC 2007.}}
The PASCAL VOC 2007\cite{pascal-voc-2007} dataset contains 9,963 realistic images of 20 object classes. From these, 5,011 images contain objects not annotated as \textit{difficult}. Two subsets emerged to enable comparisons among methods. The first subset, VOC07\_6x2, includes all images in the trainval set corresponding to 6 classes (aeroplane, bicycle, boat, bus, horse, and motorbike) and two views (left, right) with a total of 12 class combinations and 463 images. The second subset  includes all images from the 20 classes in the trainval set except for images containing only object instances annotated as \textit{difficult} or \textit{truncated}. Some authors have called this dataset PASCAL07\_all \cite{cho2015unsupervised} and referred to it as containing 4,548 images, while others have used the name VOC07\_all \cite{vo2019unsupervised, vo2020toward} and refer that the dataset contains 3,550 images. Both groups of authors report the same procedure for obtaining this subset. \newline

{\titledataset \textbf{PASCAL VOC 2012.}}
The PASCAL VOC 2012\cite{pascal-voc-2012} dataset has 11,530 images in the train/val data, spanning 20 object classes and containing 27,450 ROI annotated objects. VOC12 is a subset of the PASCAL VOC 2012 dataset and is obtained by eliminating all images containing only difficult or truncated objects as well as difficult or truncated objects in retained images. It contains 7,838 images and figures 13,957 objects. \newline

{\titledataset \textbf{MS COCO (20k).}}
The COCO 20k dataset is a subset derived from the COCO 2014 trainval dataset \cite{lin2015microsoftcococommonobjects}, comprising 19,817 randomly selected images. This subset specifically filters out instances annotated as crowd. The dataset consists of a total of 143,951 annotated objects across various categories.

\subsubsection{Datasets in Segmentation Methods}
In \hyperref[tab:segdatasets]{Table~\ref{tab:segdatasets}}, we comprise several widely used datasets among segmentation methods. These datasets include Cityscapes \cite{cordts2016cityscapesdatasetsemanticurban}, CUB \cite{Wah2011TheCB}, DUT-OMRON \cite{Yang2013SaliencyDV}, DUTS \cite{Wang2017LearningTD}, ECSSD \cite{shi2015hierarchicalsaliencydetectionextended}, Flowers \cite{Nilsback2010DelvingDI}, iCoseg, MS COCO, MSRA \cite{Cheng2015GlobalCB, Liu2007LearningTD}, MSRC, Object Discovery, PASCAL VOC, SED2, SOD, among others. For the most used datasets among segmentation methods, we include further comparisons in specific tables. A detailed description of each dataset is provided below.
\newcommand*\CROSS{---}

\begin{table}[t]
\centering

\resizebox{0.8\columnwidth}{!}{%
\begin{tabular}{lclcl
        cc
        >{\columncolor{colorseg1}}c
        >{\columncolor{colorseg1}}c
        >{\columncolor{colorseg1}}c
        c
        >{\columncolor{colorseg2}}c
        >{\columncolor{colorseg3}}c
        c
        >{\columncolor{colorseg4}}c
        >{\columncolor{colorseg5}}c
        >{\columncolor{colorseg6}}c
        ccc} 
\toprule
\multicolumn{20}{c}{\cellcolor{colorseg}\textcolor{white}{Datasets used in Segmentation Methods}} \\
\midrule
Method & Year &  & \shortstack{Custom \\ datasets} &  & \begin{sideways}Cityscapes\end{sideways} & \begin{sideways}CUB\end{sideways} & \begin{sideways}DUT-OMRON\end{sideways} & \begin{sideways}DUTS\end{sideways} & \begin{sideways}ECSSD\end{sideways} & \begin{sideways}Flowers\end{sideways} & \begin{sideways}iCoseg\end{sideways} & \begin{sideways}MS
  COCO\end{sideways} & \begin{sideways}MSRA\end{sideways} & \begin{sideways}MSRC\end{sideways} & \begin{sideways}Object
  Discovery\end{sideways} & \begin{sideways}PASCAL
  VOC\end{sideways} & \begin{sideways}SED2\end{sideways} & \begin{sideways}SOD\end{sideways} & \begin{sideways}Others\end{sideways} \\ 
\midrule
DC & 2010 &  & \CROSS &  & ~ & ~ & ~ & ~ & ~ & ~ & ~ & ~ & ~ & \MARK & ~ & ~ & ~ & ~ & \MARK \\
\arrayrulecolor{graydivider}\hline
CoSand & 2011 &  & \CROSS &  & ~ & ~ & ~ & ~ & ~ & ~ & ~ & ~ & ~ & \MARK & ~ & ~ & ~ & ~ & \MARK \\
\arrayrulecolor{graydivider}\hline
GS & 2012 &  & \CHECK &  & ~ & ~ & ~ & ~ & ~ & ~ & ~ & ~ & ~ & ~ & ~ & ~ & ~ & ~ & \MARK \\
IFBM & 2012 &  & \CROSS &  & ~ & ~ & ~ & ~ & ~ & ~ & \MARK & ~ & ~ & \MARK & ~ & ~ & ~ & ~ & ~ \\
MCC & 2012 &  & \CROSS &  & ~ & ~ & ~ & ~ & ~ & ~ & \MARK & ~ & ~ & \MARK & ~ & ~ & ~ & ~ & ~ \\
\arrayrulecolor{graydivider}\hline
COMP & 2013 &  & \CROSS &  & ~ & ~ & ~ & ~ & ~ & ~ & \MARK & ~ & ~ & \MARK & ~ & \MARK & ~ & ~ & ~ \\
CST & 2013 &  & \CROSS &  & ~ & ~ & ~ & ~ & ~ & ~ & \MARK & ~ & ~ & \MARK & ~ & ~ & ~ & ~ & ~ \\
HS & 2013 &  & \CHECK &  & ~ & ~ & ~ & ~ & ~ & ~ & ~ & ~ & \MARK & ~ & ~ & ~ & ~ & ~ & \MARK \\
OD & 2013 &  & \CHECK &  & ~ & ~ & ~ & ~ & ~ & ~ & \MARK & ~ & ~ & \MARK & \MARK & ~ & ~ & ~ & ~ \\
\arrayrulecolor{graydivider}\hline
CS-ECEM & 2014 &  & \CROSS &  & ~ & ~ & ~ & ~ & ~ & ~ & \MARK & ~ & ~ & ~ & ~ & ~ & ~ & ~ & \MARK \\
GMS & 2014 &  & \CROSS &  & ~ & ~ & ~ & ~ & ~ & ~ & \MARK & ~ & ~ & \MARK & ~ & ~ & ~ & ~ & \MARK \\
RC & 2014 &  & \CHECK &  & ~ & ~ & ~ & ~ & ~ & ~ & ~ & ~ & ~ & ~ & ~ & ~ & ~ & ~ & \MARK \\
SMD & 2014 &  & \CROSS &  & ~ & ~ & \MARK & ~ & \MARK & ~ & \MARK & ~ & \MARK & ~ & ~ & ~ & ~ & \MARK & ~ \\
wCTR & 2014 &  & \CROSS &  & ~ & ~ & ~ & ~ & ~ & ~ & ~ & ~ & \MARK & ~ & ~ & ~ & \MARK & ~ & \MARK \\
\arrayrulecolor{graydivider}\hline
DMFC & 2015 &  & \CROSS &  & ~ & ~ & ~ & ~ & ~ & ~ & \MARK & ~ & ~ & ~ & ~ & ~ & ~ & ~ & ~ \\
MRW & 2015 &  & \CROSS &  & ~ & ~ & ~ & ~ & ~ & ~ & \MARK & ~ & ~ & ~ & ~ & ~ & ~ & ~ & ~ \\
WSC & 2015 &  & \CROSS &  & ~ & ~ & ~ & ~ & ~ & ~ & ~ & ~ & \MARK & ~ & ~ & ~ & ~ & \MARK & \MARK \\
\arrayrulecolor{graydivider}\hline
INCT & 2016 &  & \CHECK &  & ~ & ~ & ~ & ~ & ~ & ~ & \MARK & ~ & ~ & \MARK & ~ & ~ & ~ & ~ & \MARK \\
SCF & 2016 &  & \CROSS &  & ~ & ~ & ~ & ~ & ~ & ~ & \MARK & ~ & ~ & \MARK & \MARK & ~ & ~ & ~ & \MARK \\
\arrayrulecolor{graydivider}\hline
CS-GSI & 2017 &  & \CROSS &  & ~ & ~ & ~ & ~ & ~ & ~ & \MARK & ~ & ~ & ~ & ~ & ~ & ~ & ~ & ~ \\
NPG & 2017 &  & \CROSS &  & ~ & ~ & ~ & ~ & ~ & ~ & \MARK & ~ & ~ & \MARK & ~ & \MARK & ~ & ~ & \MARK \\
SGC3 & 2017 &  & \CROSS &  & ~ & ~ & ~ & ~ & ~ & ~ & \MARK & ~ & ~ & ~ & \MARK & ~ & ~ & ~ & ~ \\
SBF & 2017 &  & \CROSS &  & ~ & ~ & \MARK & ~ & \MARK & ~ & ~ & ~ & \MARK & ~ & ~ & \MARK & ~ & \MARK & ~ \\
\arrayrulecolor{graydivider}\hline
CoAttnCNN & 2018 &  & \CROSS &  & ~ & ~ & ~ & ~ & ~ & ~ & \MARK & ~ & ~ & ~ & \MARK & \MARK & ~ & ~ & ~ \\
DFF & 2018 &  & \CROSS &  & ~ & ~ & ~ & ~ & ~ & ~ & \MARK & ~ & ~ & ~ & ~ & \MARK & ~ & ~ & ~ \\
USDNL & 2018 &  & \CROSS &  & ~ & ~ & \MARK & ~ & \MARK & ~ & ~ & ~ & \MARK & ~ & ~ & \MARK & \MARK & \MARK & \MARK \\
\arrayrulecolor{graydivider}\hline
CP-GAN & 2019 &  & \CROSS &  & ~ & ~ & ~ & ~ & ~ & ~ & ~ & ~ & ~ & ~ & ~ & ~ & ~ & ~ & \MARK \\
DeepCO3 & 2019 &  & \CROSS &  & ~ & ~ & ~ & ~ & ~ & ~ & ~ & \MARK & ~ & ~ & ~ & \MARK & ~ & \MARK & ~ \\
IIC-seg & 2019 &  & \CROSS &  & ~ & ~ & ~ & ~ & ~ & ~ & ~ & \MARK & ~ & ~ & ~ & ~ & ~ & ~ & \MARK \\
ReDo & 2019 &  & \CROSS &  & ~ & \MARK & ~ & ~ & ~ & \MARK & ~ & ~ & ~ & ~ & ~ & ~ & ~ & ~ & \MARK \\
UnsupOD & 2020 &  & \CROSS &  & ~ & ~ & ~ & ~ & ~ & ~ & \MARK & ~ & ~ & \MARK & \MARK & ~ & ~ & ~ & \MARK \\
\arrayrulecolor{graydivider}\hline
CS-OHS & 2021 &  & \CROSS &  & ~ & ~ & ~ & ~ & ~ & ~ & \MARK & ~ & ~ & \MARK & ~ & ~ & ~ & ~ & ~ \\
DeepUSPS & 2021 &  & \CROSS &  & ~ & ~ & \MARK & ~ & \MARK & ~ & ~ & ~ & \MARK & ~ & ~ & ~ & \MARK & ~ & ~ \\
DRC & 2021 &  & \CROSS &  & ~ & \MARK & ~ & ~ & ~ & ~ & ~ & ~ & ~ & ~ & ~ & ~ & ~ & ~ & \MARK \\
MaskContrast & 2021 &  & \CROSS &  & ~ & ~ & ~ & ~ & ~ & ~ & ~ & ~ & ~ & ~ & ~ & \MARK & ~ & ~ & ~ \\
PiCIE & 2021 &  & \CROSS &  & \MARK & ~ & ~ & ~ & ~ & ~ & ~ & \MARK & ~ & ~ & ~ & ~ & ~ & ~ & ~ \\
E-BigBiGAN & 2021 &  & \CROSS &  & ~ & \MARK & \MARK & \MARK & \MARK & \MARK & ~ & ~ & ~ & ~ & ~ & ~ & ~ & ~ & ~ \\
UVISAM & 2021 &  & \CROSS &  & ~ & \MARK & \MARK & \MARK & \MARK & ~ & ~ & ~ & ~ & ~ & ~ & ~ & ~ & ~ & ~ \\
\arrayrulecolor{graydivider}\hline
DSM & 2022 &  & \CROSS &  & ~ & \MARK & \MARK & \MARK & \MARK & ~ & ~ & ~ & ~ & ~ & ~ & \MARK & ~ & ~ & ~ \\
SegDiscover & 2022 &  & \CROSS &  & \MARK & ~ & ~ & ~ & ~ & ~ & ~ & \MARK & ~ & ~ & ~ & ~ & ~ & ~ & \MARK \\
SegSwap & 2022 &  & \CROSS &  & ~ & ~ & \MARK & \MARK & \MARK & ~ & ~ & ~ & ~ & ~ & \MARK & ~ & ~ & ~ & \MARK \\
SelfMask & 2022 &  & \CROSS &  & ~ & ~ & \MARK & \MARK & \MARK & ~ & ~ & ~ & ~ & ~ & ~ & ~ & ~ & \MARK & \MARK \\
TokenCut & 2022 &  & \CROSS &  & ~ & ~ & \MARK & \MARK & \MARK & ~ & ~ & \MARK & ~ & ~ & ~ & ~ & ~ & ~ & ~ \\
\bottomrule
\end{tabular}
}
\captionsetup{width=0.8\columnwidth}
\caption{Benchmark datasets employed in the experimental setups of Segmentation methods. Datasets corresponding to colored columns are further compared in separate tables.}
\label{tab:segdatasets}
\end{table}

{\titledataset \textbf{\MakeUppercase{Cityscapes.}}}
The Cityscapes \cite{cordts2016cityscapesdatasetsemanticurban} is a dataset that feature street scenes from 50 cities worldwide. It comprises 30 classes of instances that can be categorized into 8 group (flat surfaces, humans, vehicles, constructions, objects, nature, sky, and void). The dataset consists of approximately 5,000 finely annotated images and 20,000 coarsely annotated ones, \newline

{\titledataset \textbf{\MakeUppercase{CUB.}}}
The CUB-200-2011 \cite{Wah2011TheCB} is a dataset composed of bird photographs that are accompanied by segmentation masks and other important annotations. It is also known as Caltech-UCSD Birds-200-2011, and comprises 11,788 images categorized into 200 bird subcategories. It includes 5,994 images for training and 5,794 for testing, each annotated with a subcategory label, 15 part locations, 312 binary attributes, and 1 bounding box.  \newline

{\titledataset \textbf{\MakeUppercase{DUT-OMRON.}}}
The DUT-OMRON \cite{Yang2013SaliencyDV} dataset is generally utilized to evaluate the Salient Object Detection task. It comprises 5,168 high-quality images. These images feature one or more salient objects set against a relatively cluttered background.  \newline

{\titledataset \textbf{\MakeUppercase{DUTS.}}}
The DUTS dataset \cite{Wang2017LearningTD} is commony used in saliency detection tasks. It contains 10,553 training images and 5,019 test images. The training images are selected from the training and validations sets of the ImageNet dataset. In contrast, the test images are collected from both the SUN dataset \cite{Xiao2010SUNDL} and ImageNet test set. Both the training and test set can incorporate challenging scenes for tasks such as saliency detection and object segmentation. The dataset comes with pixel-level ground truths are manually annotated. \newline

{\titledataset \textbf{\MakeUppercase{ECSSD.}}}
The ECSSD dataset \cite{Shi2016HierarchicalIS} is an image collection containing 1,000 images with their corresponding ground-truth saliency maps. The acronym ECSSD stands for Extended Complex Scene Saliency Dataset. This dataset comprises images that contain complex and real-world scenes. \newline

{\titledataset \textbf{\MakeUppercase{OXFORD FLOWERS.}}}
The Flowers dataset, introduced in \cite{Nilsback2010DelvingDI}, comprises 8,189  images of flowers, each accompanied by automatically generated saliency masks tailored for flower analysis. This dataset consists of 102 categories of flowers commonly found in the United Kingdom, with each category containing between 40 and 258 images. The dataset exhibits variations in scale, pose, and lighting conditions, as well as considerable intra-category and inter-category similarity. It is partitioned into a training set and a validation set, each containing 10 images per category (totaling 1,020 images per set), and a test set comprising the remaining 6,149 images, with a minimum of 20 images per category in the test set. \newline

{\titledataset \textbf{iCOSEG}}
The iCoseg dataset \cite{Batra2010iCosegIC} contains 643 images divided into 38 classes. This is dataset widely used for evaluating co-segmentation methods. Here, each image has been annotated with a pixel-wise mask that encompasses the object shared across all images. The dataset is particularly challenging due to its high variability in viewpoint, lighting conditions, and object deformation. \newline

{\titledataset \textbf{MS COCO (COCO-Stuff).}}
COCO-Stuff \cite{Caesar2016COCOStuffTA} is scene-centric dataset that comprises images with 91 stuff categories and 80 things categories. Classes are usually merged to form 27 categories, 15 from stuff and 12 from things. Evaluations in some methods, specially more recent ones, consider both categories for things and stuff; other methods evaluated in stuff only. \newline

{\titledataset \textbf{MSRA-10K \& MSRA-B. }}
The MSRA10K dataset \cite{Cheng2015GlobalCB}, designed for salient object detection, comprises 10,000 images annotated with pixel-level saliency labels derived from the MSRA salient object detection dataset. In the original MSRA database, salient objects are annotated using bounding boxes provided by 3 to 9 users. The images in this dataset are randomly selected 10,000 with consistent bounding box labeling from the MSRA database. 

The MSRA-B dataset, was introduced in \cite{Liu2007LearningTD}, and consists of 5,000 images spanning training, validation, and testing sets. Specifically, 2,500 images are allocated for training, with an additional 500 images reserved for validation. The remaining 2,000 images form the test set. Notably, most images in the MSRA-B dataset feature only one salient object. \newline

{\titledataset \textbf{MSRC.}}
The MSRC dataset \cite{Shotton2006TextonBoostJA} consists of 591 manually labeled images corresponding to 21 object classes.The assigned colors act as indices into the list of object classes. Different lighting conditions, camera viewpoint, poses, and scale were considered. It is important to note that this dataset has been defined and interpreted differently across various methods. In particular, the provided details correspond to the original definition from \cite{Shotton2006TextonBoostJA} and it is followed by UnsupOD \cite{zhang2020object}. In constrast, MSRC is defined in CS-OHS \cite{shoitan2021unsupervised} as comprising 233 images of 8 object classes; it has also been defined in NPG \cite{wang2017multiple} as containing 14 classes with 420 images in total. \newline

{\titledataset \textbf{\MakeUppercase{Object Discovery.}}}
As described in \hyperref[sec:datasets_loc]{Sec.~\ref{sec:datasets_loc}}, the Object Discovery dataset \cite{Rubinstein13Unsupervised} was collected by automatically downloading images through the Bing API. As well, it has been annotated with high-detail segmentation masks. It contains approximately 15k images. \newline

{\titledataset \textbf{\MakeUppercase{PASCAL VOC.}}}
PASCAL VOC \cite{pascal-voc-2012} is a dataset containing images of complex scenes, with variable numbers of object instances from 21 semantic classes. Certain methods can define subsets from PASCAL VOC to evaluate performance. For instance, MaskContrast \cite{vangansbeke2021unsupervised} uses a validation set from PASCAL VOC, that contains 1,449 images with pixel-wise annotations. In contrast, DFF \cite{collins2018deep} utilizes an extension of the PASCAL dataset called PASCAL-Parts \cite{Chen2014DetectWY}, which decomposes object classes into fine grained parts. \newline

{\titledataset \textbf{\MakeUppercase{SED2.}}}
The SED2 \cite{Alpert2012ImageSB} dataset contains 100 images. Each image contains two salient objects of largely different sizes and locations. \newline

{\titledataset \textbf{\MakeUppercase{SOD.}}}
SOD \cite{Movahedi2010DesignAP} is dataset of salient objects and it is based on the Berkeley Segmentation Dataset. It contains 300 images of 7 subjects in total, and images may have multiple salient objects.


\subsubsection{Datasets in Decomposition Methods}
In \hyperref[tab:decdatasets]{Table~\ref{tab:decdatasets}}, we highlight several widely used datasets in decomposition methods. These datasets include Atari Games, CIFAR-10, CLEVR, Multi-dSprites Multi-MNIST, Objects Rooms, Shape Stacks, SVHN, Tetrominoes, among others. For the most used datasets among segmentation methods, we include further decomposition in specific tables. As well, a detailed description of each dataset is provided below.
\begin{table}[H]
\centering


\resizebox{0.9\columnwidth}{!}
{%
\begin{tabular}{lclcl
        ccl
        >{\columncolor{colordec1}}c
        >{\columncolor{colordec1}}cl
        >{\columncolor{colordec2}}c
        >{\columncolor{colordec2}}cl
        cclcccclc} 
\toprule
\multicolumn{23}{c}{\cellcolor{colordec}\textcolor{white}{Datasets used in Decomposition Methods}} \\
\midrule
Method & Year &  & \shortstack{Custom \\ datasets} &  & \begin{sideways}Atari Games\end{sideways} & \begin{sideways}CIFAR-10\end{sideways} &  & \begin{sideways}CLEVR\end{sideways} & \begin{sideways}\shortstack{CLEVR\\ variants}\end{sideways} &  & \begin{sideways}Multi-dSprites\end{sideways} & \begin{sideways}\shortstack{Other dSprites\\ variants}\end{sideways} &  & \begin{sideways}Multi-MNIST\end{sideways} & \begin{sideways}\shortstack{Other MNIST\\ variants}\end{sideways} &  & \begin{sideways}Objects Room\end{sideways} & \begin{sideways}Shape

Stacks\end{sideways} & \begin{sideways}SVHN\end{sideways} & \begin{sideways}Tetrominoes\end{sideways} &  & \begin{sideways}Others\end{sideways} \\ 
\midrule
AIR & 2016 &  & \CHECK &  & ~ & ~ &  & ~ & ~ &  & ~ & ~ &  & \MARK & ~ &  & ~ & ~ & ~ & ~ &  & \MARK \\ 
CST-VAE & 2016 &  & \CHECK &  & ~ & ~ &  & ~ & ~ &  & ~ & ~ &  & ~ & \MARK &  & ~ & ~ & ~ & ~ &  & ~ \\ 
TAGGER & 2016 &  & \CHECK &  & ~ & ~ &  & ~ & ~ &  & ~ & ~ &  & ~ & \MARK &  & ~ & ~ & ~ & ~ &  & \MARK \\
\arrayrulecolor{graydivider}\hline
N-EM & 2017 &  & \CHECK &  & ~ & ~ &  & ~ & ~ &  & ~ & ~ &  & ~ & \MARK &  & ~ & ~ & ~ & ~ &  & ~ \\
\arrayrulecolor{graydivider}\hline
R-NEM & 2018 &  & \CHECK &  & \MARK & ~ &  & ~ & ~ &  & ~ & ~ &  & ~ & ~ &  & ~ & ~ & ~ & ~ &  & \MARK \\ 
SQAIR & 2018 &  & \CROSS &  & ~ & ~ &  & ~ & ~ &  & ~ & ~ &  & \MARK & ~ &  & ~ & ~ & ~ & ~ &  & \MARK \\
\arrayrulecolor{graydivider}\hline
GMIOO & 2019 &  & \CHECK &  & ~ & ~ &  & ~ & ~ &  & ~ & ~ &  & ~ & \MARK &  & ~ & ~ & ~ & ~ &  & ~ \\
MONet & 2019 &  & \CHECK &  & ~ & ~ &  & \MARK & ~ &  & \MARK & ~ &  & ~ & ~ &  & \MARK & ~ & ~ & ~ &  & ~ \\
SCAE & 2019 &  & \CROSS &  & ~ & \MARK &  & ~ & ~ &  & ~ & ~ &  & ~ & \MARK &  & ~ & ~ & \MARK & ~ &  & ~ \\
SPAIR & 2019 &  & \CHECK &  & \MARK & ~ &  & ~ & ~ &  & ~ & ~ &  & ~ & \MARK &  & ~ & ~ & ~ & ~ &  & \MARK \\
SuPAIR & 2019 &  & \CROSS &  & ~ & ~ &  & ~ & ~ &  & ~ & ~ &  & \MARK & ~ &  & ~ & ~ & ~ & ~ &  & \MARK \\
\arrayrulecolor{graydivider}\hline
GENESIS & 2020 &  & \CROSS &  & ~ & ~ &  & ~ & ~ &  & \MARK & ~ &  & ~ & ~ &  & ~ & \MARK & ~ & ~ &  & ~ \\
IODINE & 2020 &  & \CHECK &  & ~ & ~ &  & \MARK & \MARK &  & \MARK & ~ &  & ~ & ~ &  & \MARK & ~ & ~ & \MARK &  & \MARK \\
LMIO & 2020 &  & \CHECK &  & ~ & ~ &  & ~ & ~ &  & \MARK & ~ &  & ~ & ~ &  & \MARK & ~ & ~ & ~ &  & ~ \\
PSGNet & 2020 &  & \CHECK &  & ~ & ~ &  & ~ & ~ &  & ~ & ~ &  & ~ & ~ &  & ~ & ~ & ~ & ~ &  & \MARK \\
SCALOR & 2020 &  & \CROSS &  & ~ & ~ &  & ~ & ~ &  & ~ & \MARK &  & ~ & \MARK &  & ~ & ~ & ~ & ~ &  & \MARK \\
SLOT ATTN & 2020 &  & \CROSS &  & ~ & ~ &  & \MARK & \MARK &  & \MARK & ~ &  & ~ & ~ &  & ~ & ~ & ~ & \MARK &  & ~ \\
SPACE & 2020 &  & \CHECK &  & \MARK & ~ &  & ~ & ~ &  & ~ & ~ &  & ~ & ~ &  & ~ & ~ & ~ & ~ &  & \MARK \\
SRN & 2020 &  & \CHECK &  & ~ & ~ &  & \MARK & ~ &  & ~ & ~ &  & ~ & ~ &  & ~ & ~ & ~ & ~ &  & ~ \\
\arrayrulecolor{graydivider}\hline
DPR-CAE & 2021 &  & \CROSS &  & ~ & \MARK &  & ~ & ~ &  & ~ & ~ &  & ~ & \MARK &  & ~ & ~ & \MARK & ~ &  & ~ \\
DTI-Sprites & 2021 &  & \CHECK &  & ~ & ~ &  & ~ & \MARK &  & \MARK & ~ &  & ~ & ~ &  & ~ & ~ & \MARK & \MARK &  & \MARK \\
ECON & 2021 &  & \CHECK &  & ~ & ~ &  & ~ & ~ &  & ~ & ~ &  & ~ & ~ &  & ~ & ~ & ~ & ~ &  & ~ \\
EfficientMORL & 2021 &  & \CROSS &  & ~ & ~ &  & \MARK & \MARK &  & \MARK & ~ &  & ~ & ~ &  & ~ & ~ & ~ & \MARK &  & ~ \\
GMAIR & 2021 &  & \CHECK &  & ~ & ~ &  & ~ & ~ &  & ~ & ~ &  & \MARK & ~ &  & ~ & ~ & ~ & ~ &  & ~ \\
GNM & 2021 &  & \CHECK &  & ~ & ~ &  & ~ & ~ &  & ~ & ~ &  & ~ & \MARK &  & ~ & ~ & ~ & ~ &  & \MARK \\
GSGNs & 2021 &  & \CHECK &  & ~ & ~ &  & ~ & \MARK &  & \MARK & ~ &  & ~ & ~ &  & ~ & ~ & ~ & ~ &  & ~ \\
MulMON & 2021 &  & \CHECK &  & ~ & ~ &  & ~ & \MARK &  & ~ & ~ &  & ~ & ~ &  & ~ & ~ & ~ & ~ &  & \MARK \\
uORF & 2021 &  & \CHECK &  & ~ & ~ &  & ~ & \MARK &  & ~ & ~ &  & ~ & ~ &  & ~ & ~ & ~ & ~ &  & ~ \\
\arrayrulecolor{graydivider}\hline
GenesisV2 & 2022 &  & \CROSS &  & ~ & ~ &  & ~ & ~ &  & ~ & ~ &  & ~ & ~ &  & \MARK & \MARK & ~ & ~ &  & \MARK \\
MarioNette & 2022 &  & \CROSS &  & \MARK & ~ &  & ~ & ~ &  & ~ & ~ &  & ~ & ~ &  & ~ & ~ & ~ & ~ &  & \MARK \\
\bottomrule
\end{tabular}
}
\captionsetup{width=0.9\columnwidth}
\caption{Benchmark datasets employed in experiments undertaken by decomposition methods. Datasets in colored columns are further compared in separate tables.}
\label{tab:decdatasets}

\end{table}

{\titledataset \textbf{ATARI.}}
An ATARI dataset \cite{Bellemare2012TheAL} consists of random images obtained from a pre-trained agent playing the games, In the case of SPACE\cite{lin2020space}, the dataset is obtained by sampling 60,000 random images for each game \textit{(Space Invaders, Air Raid, River Raid, Montezuma’s Revenge.)}. From these, 50,000 are selected for the training set, 5,000 for the validation set, and 5,000 for the testing set. In the case of MarioNette\cite{smirnov2021marionette}, ATARI Space Invaders (5,000 frames) is used, while in the case of SPAIR, the authors describe that images from Space Invaders are obtained using the Arcade Learning Environment, collected with a random policy. \newline

{\titledataset \textbf{CIFAR-10.}}
As described in \hyperref[sec:datasets_loc]{Sec.~\ref{sec:datasets_loc}}, CIFAR-10 \cite{krizhevsky2009learning} consists of 32×32 colored images from 10 classes containing 60K images. \newline

{\titledataset \textbf{\MakeUppercase{Multi-dSprites.}}}
This dataset was introduced in \cite{burgess2019monet}. The dataset consists of 64x64 RGB images and it is created from the dSprites dataset \cite{dsprites17} by adapting 1-4 randomly chosen sprites that are coloured and then composited (with occlusion) onto a single image with a uniform randomly coloured background. \newline

{\titledataset \textbf{\MakeUppercase{CLEVR.}}}
This dataset consists of 128x128 images resized from 192x192 crops that are obtained from the  synthetic 3D environment CLEVR \cite{johnson2016clevrdiagnosticdatasetcompositional} which consists of simple 3D rendered objects. The resulting images contain 2-10 visible objects per image.

The {\titledataset \textbf{CLEVR6}} dataset is a  variant of the CLEVR dataset and uses only images with 3-6 objects. \newline

{\titledataset \textbf{\MakeUppercase{Objects Room.}}}
The Objects Room dataset was created using the MuJoCo environment used by the Generative Query Networks \cite{johnson2016clevrdiagnosticdatasetcompositional}. It includes 1M 64x64 RGB images of a cubic room with colored walls, floors, and up to three visible objects. The dataset features random object sizes, shapes, and colors, sampled in HSV color-space. The wall, the floor and the objects are all coloured
randomly as well. Test variants include empty rooms, scenes with exactly six objects, and rooms with 4-6 objects of identical color. Each image contains masks for the sky, floor, walls, and objects. \newline

{\titledataset \textbf{\MakeUppercase{Multi-MNIST.}}}
The Multi-MNIST dataset contains 50×50 images constructe with MNIST digits. Each image contains zero, one or two non-overlapping random MNIST digits with equal probability. 

Other variants include \textit{Flying MNIST}, a sequential extension of
MNIST (gray-scale 24 × 24 images containing two down-sampled MNIST digits), \textit{40×40 MNIST} with digits translated up to 6 pixels in each direction, \textit{MNIST-4} with images partitioned into four quadrants and one digit placed in each quadrant,
\textit{MNIST-10} with images split into four quadrants where four mutually exclusive sets of digit classes are assigned. \newline

{\titledataset \textbf{\MakeUppercase{ShapeStacks.}}}
This is a simulation-based dataset \cite{Groth2018ShapeStacksLV} that feature 20K configurations. Each configuration can be composed of geometric primitives. As defined in GENESIS\cite{engelcke2020genesis}, Images show simulated block towers of different heights and Individual blocks can have different shapes, sizes, and colours. \newline

{\titledataset \textbf{\MakeUppercase{Tetrominoes.}}}
Introduced as the Tetris dataset in \cite{greff2019multi}, the Tetrominoes dataset consists of 60K images of size 35x35, generated by placing three non-overlapped Tetrominoes, and each Tetromino is composed of four blocks that each of 5×5 pixels. In IODINE\cite{greff2019multi}, there are a total of 17 different Tetrominoes (counting rotations). In DTI-Sprites\cite{monnier2021unsupervised}, there is a total of 19 different Tetrominoes (counting discrete rotations). \newline

{\titledataset \textbf{\MakeUppercase{SVHN.}}}
The Street View House Numbers (SVHN) dataset \cite{Netzer2011ReadingDI} is a standard dataset generally used to evaluate clustering tasks and as a digit classification benchmark. It consists of 32×32 RGB images containing digits extracted from house numbers cropped from Google Street View images. \newline

\subsection{Evaluation Metrics}
\label{subsec:metrics}
In this section, we present an overview of the commonly used metrics for evaluating the various methodologies employed in unsupervised object discovery approaches. In line with our previous observations, based on the consideration of the four different classes of methods that address the task of object discovery, we further observe that certain metrics have been consistently employed by methods within a same group (see \hyperref[tab:metrics_usage]{Table~\ref{tab:metrics_usage}})---\textit{e.g. CorLoc} in Localization methods, while other metrics have been employed across different groups---\textit{e.g. Acc} in Clustering and Segmentation methods. This behaviour becomes evident in the upcoming descriptions of metrics.

It is also worth acknowledging that evaluating unsupervised methods is difficult because there is no access to ground-truth information. Therefore, to enable an evaluation, predictions should be matched to ground-truth labels in a linear assignment problem. A common practice that emerged in the literature to solve this problem, is based on the algorithm Hungarian Matching \cite{Kuhn1955TheHM}, which seeks to find the best one-to-one permutation mapping between predictions and ground-truth labels. In a certain way, formulating this evaluation as a linear assignment problem helps to name the predictions with corresponding labels in order to enable an evaluation that would not be possible otherwise. Moreover, even if this evaluation protocol requires access to ground-truth labels, this procedure does not constitute supervised learning. \newline

\begin{table}[h!]
\centering
\resizebox{0.65\columnwidth}{!}{%

\begin{tabular}{lcccccccc}
\toprule
{Metric} & \cellcolor{colorclust}\textcolor{white}{Clustering} && \cellcolor{colorloc}\textcolor{white}{Localization} && \cellcolor{colorseg}\textcolor{white}{Segmentation} && \cellcolor{colordec}\textcolor{white}{Decomposition} \\
\midrule
Accuracy       & \MARK  & & \EMPTY & & \MARK  & & \EMPTY  \\
NMI            & \MARK  & & \EMPTY & & \EMPTY & & \EMPTY \\
CorLoc         & \EMPTY  & & \MARK  & & \EMPTY  & & \EMPTY \\
F-Measure      & \EMPTY  & & \EMPTY & & \MARK  & & \EMPTY  \\
IoU            & \EMPTY  & & \MARK  & & \MARK  & & \MARK \\
pACC           & \EMPTY  & & \EMPTY & & \MARK  & & \EMPTY \\
MAE            & \EMPTY  & & \EMPTY & & \MARK  & & \EMPTY \\
FID            & \EMPTY  & & \EMPTY & & \EMPTY & & \MARK  \\
ARI            & \EMPTY   & & \EMPTY & & \EMPTY & & \MARK \\
\bottomrule
\end{tabular}
}
\captionsetup{width=0.65\columnwidth}
\caption{Metrics used across different classes of methods.}
\label{tab:metrics_usage}
\end{table}

We now present brief descriptions of these metrics including their formulations. \newline

{\titledataset \textbf{\MakeUppercase{ACC.}}}
This metric measures the accuracy of cluster assignments and is defined as the number of true positives divided by sample size. It can be computed as:

\[
ACC = \max_m \frac{ \sum_{n=1}^N \mathbf{1}{\{l_n = m(c_n)\}}}{N}, 
\]

where $m$ ranges over all possible one-to-one mappings between ground-truth labels and assigned clusters, $c_n$ represents the cluster assignment and $l_n$ the ground-truth label. The mapping is generally obtained through linear assignment \cite{Kuhn1955TheHM}.  \newline

{\titledataset \textbf{\MakeUppercase{NMI.}}}
Normalized Mutual Information is a normalization of the Mutual Information score that scales the score to the range [0,1]. In this case, the value 0 represents no mutual information while the value 1 is a perfect correlation. The NMI is calculated as:

\[
NMI(G,C) = \frac{2 \cdot I(G;C)}{H(G)+H(C)}, 
\]

where $I(G;C)$ is the Mutual Information between ground-truth labels $G$ and cluster assignments $C$, $H(C)$ is the entropy of clusters $C$, and $H(G)$ is the entropy of labels $G$. \newline

{\titledataset \textbf{\MakeUppercase{CorLoc.}}}
The Correct Localization (CorLoc) metric measures the percentage of accurately predicted bounding boxes. A predicted box is considered to be correct if its intersection over union (IoU) score measured against any of the ground-truth bounding boxes is greater than 0.5. 

\[
{CorLoc} = \frac{1}{N} \sum_{i=1}^{N} \mathbf{1} \left( \exists j \; \mathrm{s.t.} \; {IoU}(\hat{B}_i, B_j) > 0.5 \right)
\]

where \( {IoU}(\hat{B}_i, B_j) \) is the intersection over union between the predicted bounding box \( \hat{B}_i \) and the ground'truth bounding box \( B_j \). \newline

{\titledataset \textbf{\MakeUppercase{F-measure.}}}
The F-measure is a standard measure in the saliency detection literature. It is computed as

\[
F_\beta = \frac{(1 + \beta^2) \cdot Precision \cdot Recall}{\beta^2 \cdot Precision + Recall}
\]

where $Precision = \frac{tp}{tp + fp}$ and $Recall = \frac{tp}{tp + fn}$ are calculated based on binarised predicted masks and ground truth. Here, $tp, tn, fp,$ and $fn$ denote true positives, true negatives, false positives, and false negatives, respectively. In general, the F-measure is computed for 255 uniformly distributed thresholds and only its maximum value $maxF_\beta$ is reported. In addition, a common choice for computing this metric is $\beta = 0.3$. \newline

{\titledataset \textbf{{IoU.}}}
The Intersection over Union score IoU measures the size of the overlap between the ground-truth and the foreground or object regions determined by a predicted mask. In the binary case, \textit{e.g.} saliency methods, the $IoU$ is computed on the binary predicted mask $M$ \footnote{A binarisation threshold of 0.5 is commonly used in saliency methods.} and the ground-truth $G$ as:

\[
{IoU}(G, M) = \frac{\mu(G \cap M)}{\mu(G \cup M)},
\]

where $\mu$ denotes the overlap area. \newline

{\titledataset \textbf{{pACC.}}}
The Pixel Accuracy metric evaluates the proportion of pixels that have been correctly assigned to the corresponding classes. That is, it measures pixel-wise accuracy based on a ground-truth mask G and a mask prediction M. In the case of segmentation methods, the Hungarian algorithm is generally used for label assignment before computing the accuracy. In the case of saliency methods, the classes reduce to the object and background regions and the binarisation threshold for distinguishing foreground from background is set to 0.5. This metric can be computed as:
\[
{pAcc} = \frac{1}{H \times W} \sum_{i=1}^{H} \sum_{j=1}^{W} \delta_{G_{ij}, M_{ij}}
\]
where $\delta$ denotes the Kronecker delta function. \newline

{\titledataset \textbf{\MakeUppercase{mae.}}}
The Mean Absolute Error \textit{MAE} provides an estimate of the dissimilarity between the predicted and the ground-truth saliency map. It is defined as the average per-pixel difference between the predicted mask $S$ and the ground-truth map $GT$, normalized to $[0, 1]$. It can be computed \cite{zhang2018deep} as:

\[
{MAE} = \frac{1}{W \times H} \sum_{x=1}^{W} \sum_{y=1}^{H} \left| S(x, y) - {GT}(x, y) \right|,
\]

where \(W\) and \(H\) are the width and height of the respective maps. \newline

{\titledataset \textbf{\MakeUppercase{precision.}}}
The Precision metric $P$ indicates the percentage of correctly labeled pixels. It was used in earlier methods and is equivalent to the pixel Accuracy metric. 

\[
P = \frac{{Number \; of \; Correctly \; Labeled \; Pixels}}{{Total \; Number \; of \; Labeled \; Pixels}}
\]

In other words, if \( TP \) represents the number of true positive pixels and \( FP \) represents the number of false positive pixels, Precision can be expressed as:

\[
P = \frac{TP}{TP + FP}
\]

Additionally, it is worth observing that this metric can also be subject of different computations depending on the experimental settings established by a particular method. Notably, the average precision may be computed across all images in the validation set---in a global pixel-wise fashion, or averaging the individual scores obtained per image. \newline

{\titledataset \textbf{\MakeUppercase{ari.}}}
The Adjusted Rand Index $ARI$ \cite{Rand1971, Hubert1985ComparingP} is a metric for evaluating clustering similarity. This score ranges from 0, which indicates random chance, to 1, which indicates a perfect clustering. It can be used as a measure of instance segmentation quality by considering foreground pixels as one point and its segmentation as cluster assignment. It can be calculated as:

\[
{ARI} = \frac{ {RI} - {Expected \; RI} }{ {Max \; RI} - {Expected \; RI} }
\]

where \({RI}\) is the Rand Index, which measures the proportion of pairs of points that are either clustered together or separated together in both the ground truth and the clustering. \({Expected RI}\) is the expected value of the Rand Index for a random clustering. \({Max RI}\) is the maximum value of the Rand Index, which occurs when the clustering perfectly matches the ground truth. The Rand Index $RI$ is computed as:

\[
{RI} = \frac{ TP + TN }{ TP + TN + FP + FN }
\]

where $TP$ is the number of true positives, $TN$ is the number of true negatives, $FP$ represents the number of false positives, and $FN$ accounts for the number of false negatives.

In addition, the background label is frequently excluded from the computation of this metric. This variant is usually known as \textit{FG-ARI}. \newline

{\titledataset \textbf{\MakeUppercase{fid.}}}
The Fréchet Inception Distance \textit{FID} \cite{heusel2018ganstrainedtimescaleupdate} is a metric used to evaluate the quality of generated images by comparing them to real images. The FID measures how similar are the distributions of features extracted ---by using a pre-trained Inception v3 network--- from generated and real images. It measures the distance between the multivariate Gaussian distributions of these features for real and generated images. A lower FID score indicates higher similarity. 

\[
{FID} = \|\mu_r - \mu_g\|^2 + {Tr}(\Sigma_r + \Sigma_g - 2(\Sigma_r \Sigma_g)^{1/2})
\]

where \(\mu_r\) and \(\mu_g\) are the mean feature vectors for the real and generated images, respectively. \(\Sigma_r\) and \(\Sigma_g\) are the covariance matrices for the real and generated images, respectively; and, \({Tr}(\cdot)\) denotes the trace of a matrix. \newline \newline \newline

Having described these commonly used metrics, it is important to recognise that not all methods that evaluate performance on the same task and/or dataset, employ the same metrics. Moreover, the computation of these metrics can vary depending on the specific implementation choices and experimental setup employed by each method. 

For instance, while various methods may report the mean Intersection over Union (mIoU) as an evaluation metric, they might employ slightly different approaches to compute it. Specifically, the mIoU can be computed across all images in the evaluation set, \textit{i.e.} in a pixel-wise approach, or by averaging the individual IoU scores obtained for each image and/or for each class. To further illustrate these differences, consider the case of computing the mIoU metric globally across the dataset. Since the mIoU is generally reported as a single value for comparisons against baselines, the total mIoU score may be different if it is taken as the average of class mIoUs computed across the dataset as: ${mIoU} = \frac{1}{C} \sum_{c=1}^{C} {IoU}_c$, having ${IoU}_c = \frac{TP_c}{TP_c + FP_c + FN_c}$ for each class, rather than computing the mIoU as the average across all predicted instances: ${mIoU} = \frac{1}{N} \sum_{i=1}^{N} {IoU}_i$, where an instance can be represented by an individual image, and therefore, requiring to compute the IoU for each image individually, before averaging globally. Furthermore, the scores can be influenced by pre-processing procedures, the selection of specific subsets from the dataset, class imbalance within the evaluation set, and whether or not only foreground classes are included.---\textit{e.g.} including the background class can have an important impact in the final score. 

Recent methods \cite{melas-kyriazi2022deep, seitzer2022bridging, vangansbeke2021unsupervised}, are increasingly converging on the evaluation practice of computing mIoU at the pixel level across the entire dataset for each class separately, and then averaging these class-specific mIoUs. Additionally, there is a growing consensus to apply the Hungarian algorithm for matching predicted clusters/segments to ground-truth classes, excluding the background class from the matching process.

These specific practices and variations in evaluation protocols, underscore the existing difficulties that have to be addressed in order to formulate fair evaluations and comparisons with relevant methods. Consequently, these challenges highlight the important need for standardized evaluation benchmarks. \newline

\subsection{Comparisons}
In the following sections, we present results reported from experiments undertaken by potentially comparable methods. Each table groups the corresponding methods based on one of the four main classes of methods analyzed in this survey: clustering, localization, segmentation and decomposition. Only the most frequently used datasets are considered in these tables, in order to facilitate the comparisons of methods across a comparable task.

We emphasize that these are \textit{potential comparisons}, as these results may not always reflect directly comparable evaluations. This particularity arises from the fact that various methods can be evaluated using certain subsets of the employed datasets, and certain metrics may be calculated within specific settings defined by each method. In fact, constructing these comparisons has revealed that, across the range of methodologies analyzed in this survey, there exists an important diversity of evaluation settings and practices that introduce significant challenges in comparing and assessing the actual performance of seemingly comparable approaches. Thus, the findings presented herein are primarily intended to provide a general overview of potential comparability among these methods.

We hope this work serves as a catalyst for researchers to recognize the importance of standardizing benchmarks for object discovery and motivates the use of standardized evaluation practices.

\subsubsection{Comparison of Clustering Methods}
In this section, we present a comparison of the results obtained by different clustering methods based on two important metrics for this class of methods. The comparison is presented across the most common datasets generally utilized in such experiments.
\begin{table}[H]
\centering
\resizebox{\columnwidth}{!}{%
\begin{tabular}{llccccccccccclccccc}
\toprule
 &
   &
  \multicolumn{2}{c}{\cellcolor{colorclust}\textcolor{white}{MNIST}} &
  &
  \multicolumn{2}{c}{\cellcolor{colorclust}\textcolor{white}{CIFAR10}} &
  \multicolumn{1}{l}{} &
  \multicolumn{2}{c}{\cellcolor{colorclust}\textcolor{white}{CIFAR100}} &
  \multicolumn{1}{l}{} &
  \multicolumn{2}{c}{\cellcolor{colorclust}\textcolor{white}{STL10}} &
   &
  \multicolumn{2}{c}{\cellcolor{colorclust}\textcolor{white}{SVHN}} &
  \multicolumn{1}{l}{} &
  \multicolumn{2}{c}{\cellcolor{colorclust}\textcolor{white}{CUB}} \\ \cmidrule(lr){3-4} \cmidrule(lr){6-7} \cmidrule(lr){9-10} \cmidrule(lr){12-13} \cmidrule(lr){15-16} \cmidrule(l){18-19} 
{Method} &
   &
  $Acc$ &
  $NMI$ &
  \multicolumn{1}{l}{} &
  $Acc$ &
  $NMI$ &
  \multicolumn{1}{l}{} &
  $Acc$ &
  $NMI$ &
  \multicolumn{1}{l}{} &
  $Acc$ &
  $NMI$ &
   &
  $Acc$ &
  $NMI$ &
  \multicolumn{1}{l}{} &
  $Acc$ &
  $NMI$ \\ \midrule
  AAE       &  & 95.90\tpm1.13 & \nuh  &  & \nuh  & \nuh  &  & \nuh  & \nuh  &  & \nuh  & \nuh  &  & \nuh  & \nuh &  & \nuh  & \nuh  \\
  \rowcolor{rowbackground}
  ADC       &  & 99.20         & \nuh  &  & 32.50 & \nuh  &  & \nuh  & \nuh  &  & 53.00 & \nuh  &  & 45.30 & \nuh &  & \nuh  & \nuh  \\
  DAC       &  & 97.75         & 93.51 &  & 52.18 & 39.59 &  & 23.75 & 18.52 &  & 46.99 & 36.56 &  & \nuh  & \nuh &  & \nuh  & \nuh  \\
  \rowcolor{rowbackground}
  DEC       &  & 84.30         & \nuh  &  & \nuh  & \nuh  &  & \nuh  & \nuh  &  & 35.90 & \nuh  &  & \nuh  & \nuh &  & \nuh  & \nuh  \\
  DTI-GMM   &  & 97.10         & 93.70 &  & \nuh  & \nuh  &  & \nuh  & \nuh  &  & \nuh  & \nuh  &  & 63.30 & \nuh &  & \nuh  & \nuh  \\
  \rowcolor{rowbackground}
  FineGAN   &  & \nuh          & \nuh  &  & \nuh  & \nuh  &  & \nuh  & \nuh  &  & \nuh  & \nuh  &  & \nuh  & \nuh &  & 12.60 & 40.30 \\
  GMVAE     &  & 96.92         & \nuh  &  & \nuh  & \nuh  &  & \nuh  & \nuh  &  & \nuh  & \nuh  &  & \nuh  & \nuh &  & \nuh  & \nuh  \\
  \rowcolor{rowbackground}
  IIC       &  & 99.20         & \nuh  &  & 61.70 & \nuh  &  & 25.70 & \nuh  &  & 59.60 & \nuh  &  & \nuh  & \nuh &  & \nuh  & \nuh  \\
  IMSAT     &  & 98.40\tpm0.40  & \nuh &  & 
                 45.60\tpm0.80\textsuperscript{\tsa} & \nuh &  & 
                 27.50\tpm0.40\textsuperscript{\tsa} & \nuh &  & 
                 94.10\tpm0.40\textsuperscript{\tsa} & \nuh &  & 
                 57.30\tpm3.90  & \nuh &  &
                 \nuh           & \nuh \\
  \rowcolor{rowbackground}
  JULE      &  & 96.40         & 91.30 &  & \nuh  & \nuh  &  & \nuh  & \nuh  &  & \nuh  & \nuh  &  & \nuh  & \nuh &  & 
                 4.50\textsuperscript{\tsb} & 20.40\textsuperscript{\tsb} \\
  SCAN      &  & \nuh          & \nuh  &  & 88.30 & 79.70 &  & 50.70 & 48.60 &  & 80.90 & 69.80 &  & \nuh  & \nuh &  & \nuh   & \nuh    \\
  \rowcolor{rowbackground}
  TwoFoldOp &  & 98.88\tpm0.07 & 96.74\tpm0.16 &  & 
                 57.97\tpm3.03 & 47.03\tpm2.04 &  & 
                 25.94\tpm0.80 & 19.72\tpm0.41 &  & 
                 63.84\tpm2.60 & 50.30\tpm2.13 &  &
                 \nuh          & \nuh          &  &
                 \nuh          & \nuh \\ \bottomrule
\end{tabular}%
}
\caption{Comparison of methods on image clustering according to evaluations reported by the authors. \tsa\, On this dataset, the method uses features pre-training on Imagenet. \tsb\, Denotes experiments by FineGAN\cite{singh2019finegan}.}
\label{tab:compclust}
\end{table}

As seen in \hyperref[tab:compclust]{Table~\ref{tab:compclust}}, the performance of clustering methods across different datasets can significantly vary depending on the dataset's complexity and characteristics. For instance, in MNIST, a relatively simple and well-structured dataset, ADC and IIC both achieve an accuracy of 99.20\%, but shifting to more complex datasets like CIFAR10 and CIFAR100 introduces substantial performance variability. In CIFAR10, SCAN obtains a good performance with an accuracy of 88.30\% and an NMI of 79.70\%, demonstrating an ability to handle more challenging images. As the complexity increases with CIFAR100, is expected that methods show a decline in performance. Here for example, TwoFoldOp and DAC struggle with an accuracy around 25\%, which underscores the difficulty in maintaining high clustering performance when increasing class diversity. In STL10, which has fewer classes but higher image resolution, IMSAT obtains an accuracy of 94.10\%, the highest among the methods compared, likely benefiting from its use of features extracted by pre-training on ImageNet. A lower number of clustering methods have been evaluatred in SVHN and CUB. In the SVHN dataset, which involves street view house numbers, DTI-GMM achieves the highest accuracy of 63.30\%, which is still far from the accuracy of 97.1\% obtained in MNIST. This variation between MNIST and SVHN highlights the impact of image variability and noise on clustering performance. Finally, the CUB dataset, focused on fine-grained bird species classification, poses a more challenging scenario. Only two methods have reported results on this dataset. FineGAN, with a low accuracy of 12.60\% but a relatively higher NMI of 40.30\%, suggests that while its overall clustering accuracy is low, it still captures significant information about the clusters. As seen in this comparison, even when accuracy is low, some methods can still effectively identify underlying data structures.

Overall, this table underscores the relevance of experimental settings as no single method excels across datasets. That is, methods that perform well on simpler datasets like MNIST often degrade on more complex datasets such as CIFAR100 or CUB. 

\subsubsection{Comparison of Localization Methods}
In \hyperref[tab:comploc]{Table~\ref{tab:comploc}}, we present a comparison of the results obtained by different localization methods based on the CorLoc metric---which is one of the most common metrics utilized in this class of methods. The comparison is presented across the most common datasets generally utilized in such experiments. 
\begin{table}[H]
\centering


\resizebox{\columnwidth}{!}{%

\begin{tabular}{lclccclcccclclc} 
\toprule
~ & \multicolumn{1}{l}{~} &  & \multicolumn{3}{c}{\cellcolor{colorloc}\textcolor{white}{VOC07 ~ ~}} & \multicolumn{1}{c}{} & \multicolumn{2}{c}{\cellcolor{colorloc}\textcolor{white}{VOC12 ~}} & \multicolumn{1}{l}{} & {\cellcolor{colorloc}\textcolor{white}{COCO}} & \multicolumn{1}{c}{} & \multirow{2}{*}{\cellcolor{colorloc}\textcolor{white}{\shortstack{OD \\ OD }}} & \multicolumn{1}{c}{} & \multicolumn{1}{l}{\multirow{2}{*}{\cellcolor{colorloc}\textcolor{white}{\shortstack{ImageNet Subsets \\ Subsets}}}} \\ 
\cline{4-6}\cline{8-9}\cmidrule{11-11}
~ ~ Method & Year &  & 6x2 & All & trainval &  & All & trainval &  & 20k &  &  &  & \multicolumn{1}{l}{} \\ 
\midrule
\multicolumn{1}{c}{} & \multicolumn{1}{l}{} &  & \multicolumn{12}{c}{\textit{Single-Class Setting}} \\ 
\cmidrule{4-15}
CL-FWA & 2014 &  & - & 24.59 & - &  & - & - &  & - &  & - &  & - \\
\rowcolor{rowbackground}
CL-RWI & 2014 &  & 39.31 & - & - &  & - & - &  & - &  & 76.58 &  & - \\
UODLW & 2015 &  & 67.68 & 36.60 & - &  & - & - &  & - &  & 84.19 &  & - \\
\rowcolor{rowbackground}
CL-CSD & 2016 &  & - & 40.0 & - &  & 43.80 & - &  & - &  & - &  & 48.30 \\
DDT & 2017 &  & - & 46.90 & - &  & 49.40 & - &  & - &  & 88.13 &  & 69.10 \\
\rowcolor{rowbackground}
DFF & 2018 &  & - & 43.51 & - &  & - & - &  & - &  & - &  & - \\
CL-CCF & 2019 &  & - & 41.20 & - &  & 47.45 & - &  & - &  & 85.67 &  & 63.50 \\
\rowcolor{rowbackground}
OSD & 2019 &  & 69.4 ± 0.30 & 39.2 ± 0.20 & - &  & - & - &  & - &  & 85.8 ± 0.60 &  & - \\
UODSC & 2019 &  & - & 60.20 & - &  & - & - &  & - &  & 89.70 &  & 79.40 \\
\rowcolor{rowbackground}
rOSD & 2020 &  & 76.1 ± 0.70 & 46.7 ± 0.20 & - &  & 49.2 ± 0.10 & - &  & - &  & 90.2 ± 0.30 &  & - \\
UnsupOD & 2020 &  & - & - & - &  & - & - &  & - &  & 80.64 &  & - \\ 
\midrule
\multicolumn{1}{c}{} & \multicolumn{1}{l}{} &  & \multicolumn{12}{c}{\textit{Multi-Class Setting}} \\ 
\cmidrule{4-15}
UODLW & 2015 &  & 53.73 & 31.30 & - &  & - & - &  & - &  & 82.35 &  & - \\
\rowcolor{rowbackground}
DDT & 2017 &  & - & - & 50.20\# &  & - & 53.10\textsuperscript{$\dagger$} &  & 38.20\textsuperscript{$\dagger$} &  & - &  & - \\
UODSC & 2019 &  & - & 49.50 & - &  & - & - &  & - &  & 89.40 &  & 77.60 \\
\rowcolor{rowbackground}
OSD & 2019 &  & 60.2 ± 0.40 & 39.8 ± 0.20 & - &  & - & - &  & - &  & 83.0 ± 0.40 &  & - \\
OLM & 2020 &  & 58.29 & 37.90 & - &  & - & - &  & - &  & 85.80 &  & 60.10 \\
\rowcolor{rowbackground}
rOSD & 2020 &  & 72.5 ± 0.50 & 49.3 ± 0.20 & 54.50\textsuperscript{$\dagger$} &  & 51.2 ± 0.20 & 55.30\textsuperscript{$\dagger$} &  & 48.5 ± 0.10 &  & 89.2 ± 0.40 &  & - \\
LOD & 2021 &  & - & 48.00 & 53.60\textsuperscript{$\dagger$} &  & 50.50 & 55.10\textsuperscript{$\dagger$} &  & 48.50 &  & - &  & - \\
\rowcolor{rowbackground}
LOST & 2021 &  & - & 54.90 & 61.90 &  & 57.50 & 64.00 &  & 50.70 &  & - &  & - \\
DSM & 2022 &  & - & - & 62.70 &  & - & 66.40 &  & 52.20 &  & - &  & ~ \\
\rowcolor{rowbackground}
TokenCut & 2022 &  & - & - & 68.8 &  & - & 72.1 &  & 58.8 &  & - &  & - \\
\bottomrule
\end{tabular}

}

\caption{CorLoc performance on popular datasets in classical tasks of \textit{co-localization} and \textit{true discovery}. In the single-class setting (co-localization), methods are run on each class separately and CorLoc is averaged for the entire dataset. In the multi-class setting (true discovery), methods are run in the entire dataset. {$\dagger$}\ Denotes results reported by LOST\cite{simeoni2021localizing}.}

\label{tab:comploc}
\end{table}
As seen in the table, there are two main settings in which localization methods are generally evaluated: single-class and multi-class settings. In the single-class setting, rOSD achieves the highest CorLoc of 76.1 on the \textit{VOC 07 6x2} subset, as an improvement over OSD that obtained 69.4. For the \textit{VOC 07 All} subset, UODSphClust shows the highest reported CorLoc at 60.20, greatly improving over DDT that obtained 46.90. In VOC12 experiments, DDT achieved the highest CorLoc of 49.40 on the \textit{VOC12 ALL} subset. In the Object Discovery \textit{OD} dataset, rOSD obtained the highest CorLoc of 90.2, improving over UODSphClust with 89.70. A lower number of methods have approached \textit{ImageNet} subsets, having UODSphClust obtain a higher performance with a CorLoc of 79.40. It is worth recalling that the Single-Class Setting evaluation corresponds to Co-Localization methods. This comparison reveals VOC07 and OD as the datasets of choice for this task. It is also noticeable that none of these methods evaluate this task on MS COCO. 

In the multi-class setting, rOSD achieves the highest CorLoc of 72.5 on \textit{VOC 07 6x2}. More recent methods such as LOST and TokenCut demonstrated improvements in the other subsets with CorLocs of 54.90 and 68.8 respectively in \textit{VOC 07 All}. The latter method achieved the highest CorLoc in the \textit{VOC07 trainval} subset. On the other hand, in \textit{VOC12 trainval}, TokenCut has the highest CorLoc with 72.1, followed by DSM 66.4 and LOST 64.0. In the case of COCO 20k, TokenCut achieves the highest CorLoc of 58.8, with LOST and DSM following with CorLocs of 50.70 and 52.20 respectively.

In addition, it can be observed that more recent methods focus more in the datasets \textit{VOC12} and \textit{MS COCO}, while only a few methods evaluated in ImageNet subsets. Also, it is notable that no single method obtains superior performance across the diverse settings.

\subsubsection{Comparison of Segmentation Methods}
The following tables present different comparisons of the results reported by various segmentation methods across the most commonly used metrics and datasets. 

\newcommand*\dutasymbol{$\star$}
\newcommand*\dutbsymbol{$\ddagger$}
\newcommand*\duta{\textsuperscript{\dutasymbol}}
\newcommand*\dutb{\textsuperscript{\dutbsymbol}}

\begin{table}[ht]
\centering


\resizebox{\columnwidth}{!}{%

\begin{tabular}{lccccccccccccccccc} 
\toprule
~ & \multicolumn{1}{l}{~} & \multicolumn{1}{l}{} & \multicolumn{3}{c}{\cellcolor{colorseg1}\textcolor{white}{\shortstack{DUT \\ OMRON}}} & \multicolumn{1}{l}{} & \multicolumn{5}{c}{\cellcolor{colorseg1}\textcolor{white}{DUTS}} & \multicolumn{1}{l}{} & \multicolumn{5}{c}{\cellcolor{colorseg1}\textcolor{white}{ECSSD}} \\ 
\cmidrule{4-6}\cmidrule{8-12}\cmidrule{14-18}
Method & Year &  & Acc. & IoU & $maxF_\beta$ &  & Acc. & IoU & $maxF_\beta$ & $F_\beta$ & MAE &  & Acc. & IoU & $maxF_\beta$ & $F_\beta$ & MAE \\ 
\midrule
HS & 2013 &  & 84.30\duta & 43.30\duta & 56.10\duta &  & 82.60\duta & 36.90\duta & 50.40\duta & 52.05\dutb & 0.2274\dutb &  & 84.70\duta & 50.80\duta & 67.30\duta & 62.34\dutb & 0.2283\dutb \\
\rowcolor{rowbackground}
SMD & 2014 &  & - & 44.10 & - &  & - & - & - & ~ & ~ &  & - & 52.30 & - & - & - \\
wCTR & 2014 &  & 83.80\duta & 41.60\duta & 54.10\duta &  & 83.50\duta & 39.20\duta & 52.20\duta & 51.00\dutb & 0.2011\dutb &  & 86.20\duta & 51.70\duta & 68.40\duta & 65.18\dutb & 0.1832\dutb \\
\rowcolor{rowbackground}
WSC & 2015 &  & 86.50\duta & 38.70\duta & 52.30\duta &  & 38.40\duta & 86.20\duta & 52.80\duta & - & - &  & 85.20\duta & 49.80\duta & 68.30\duta & - & - \\
SBF & 2017 &  & - & - & - &  & - & - & - & 58.30 & 0.1350 &  & - & - & - & 78.70 & 0.0850 \\
\rowcolor{rowbackground}
USD & 2018 &  & - & - & - &  & - & - & - & 71.56 & 0.0860 &  & - & - & - & 87.83 & 0.0704 \\
DeepUSPS & 2021 &  & 77.90\duta & 30.50\duta & 41.40\duta &  & 77.30\duta & 30.50\duta & 42.50\duta & - & - &  & 79.50\duta & 44.00\duta & 58.40\duta & - & - \\
\rowcolor{rowbackground}
E-BigBiGAN & 2021 &  & 86.00 & 46.40 & 56.30 &  & 88.20 & 51.10 & 62.40 & - & - &  & 90.60 & 68.40 & 79.70 & - & - \\
UVISAM & 2021 &  & 89.90 & 46.00 & - &  & 92.70 & 59.70 & - & - & - &  & 90.40 & 65.70 & - & - & - \\
\rowcolor{rowbackground}
DSM & 2022 &  & - & 56.70 & - &  & - & 51.40 & - & - & - &  & - & 73.30 & - & - & - \\
SegSwap & 2022 &  & 81.80 & 48.90 & 57.80 &  & 88.70 & 57.20 & 69.70 & - & - &  & 91.60 & 72.30 & 83.70 & - & - \\
\rowcolor{rowbackground}
SelfMask & 2022 &  & 91.90 & 65.50 & 85.20 &  & - & - & - & - & - &  & 95.50 & 81.80 & 95.60 & - & - \\
TokenCut & 2022 &  & 89.70 & 61.80 & 69.70 &  & 91.40 & 62.40 & 75.50 & - & - &  & 93.40 & 77.20 & 87.40 & - & - \\
\bottomrule
\end{tabular}
}

\caption{Performance comparison on three popular benchmark datasets for segmentation tasks. {\dutasymbol \ Indicates results reported by E-BigBiGAN\cite{voynov2021object}. \dutbsymbol \ Denotes results reported by USD\cite{zhang2018deep}.}}

\label{tab:comp_dut}
\end{table}

\hyperref[tab:comp_dut]{Table~\ref{tab:comp_dut}} illustrates the performance of various object segmentation methods in the DUT OMRON, DUTS, and ECSSD datasets. Not surprisingly, more recent methods such as SelfMask and TokenCut (\textit{Acc.} 91.90 and 89.70 respectively) have obtained important improvements in metrics such as Accuracy and F-Score compared to early methods such as HS or wCTR (\textit{Acc.} 84.30 and 83.80 respectively). However, the improvement in IoU has been much more significative from HS 43.40 to SelfMask 65.50 in DUT OMRON, and HS 50.80 to SelfMask 81.80 in ECSSD. This table, also reveals that metric such as Acc, IoU and max F-Score have been constantly evaluated and are still commonly use to measure performance of object segmentation. However, metrics such as MAE in the DUTS dataset have not been employed for evaluating more recent methods. 

\hyperref[tab:seg_comp_coco]{Table~\ref{tab:seg_comp_coco}} presents a comparison of evaluations in MS COCO and its variants. This table reveals that the most frequently used variant is COCO-Stuff. According to results reported by the method SegDiscover, there has been an important performance jump in more recently years, coming from an \textit{Acc.}  33.91 to 56.53. It is also clear that this dataset is slowly becoming more widely employed in unsupervised segmentation tasks.

\begin{table}[ht]
\centering


\resizebox{\columnwidth}{!}{%
\begin{tabular}{lcccccccccccccc} 
\toprule
~ & \multicolumn{1}{l}{~} & \multicolumn{4}{c}{{\cellcolor{colorseg3}}MS COCO} & \multicolumn{1}{l}{} & \multicolumn{2}{c}{{\cellcolor{colorseg3}}\shortstack{COCO \\ Things}} & \multicolumn{1}{c}{} & \multicolumn{3}{c}{{\cellcolor{colorseg3}}\shortstack{COCO \\ Stuff}} & \multicolumn{1}{l}{} & \multicolumn{1}{c}{{\cellcolor{colorseg3}}\shortstack{COCO \\ Stuff-3}} \\ 
\cmidrule{3-6}\cmidrule{8-9}\cmidrule{11-13}\cmidrule{15-15} Method & Year & $Acc$ & $mIoU$ & $ABO^c$ & $OR$ &  & $Acc$ & $mIoU$ &  & $Acc$ & $mIoU$ & $wIoU$ &  & $Acc$ \\ 
\midrule
DeepCO3 & 2019 & \nuh & \nuh & \nuh & \nuh &  & \nuh & \nuh &  & \nuh & \nuh & \nuh &  & \nuh \\
\rowcolor{rowbackground}
IIC-seg & 2019 & 27.7 & 6.71{\textsuperscript{$\dagger \star$}} & \nuh & \nuh &  & 43.93{\textsuperscript{$\dagger \star$}} & 13.64{\textsuperscript{$\dagger \star$}} &  & 33.91{\textsuperscript{$\dagger \star$}} & 12.00{\textsuperscript{$\dagger \star$}} & \nuh &  & 72.3 \\
MaskContrast & 2021 & \nuh & \nuh & \nuh & \nuh &  & \nuh & \nuh &  & 23.03{\textsuperscript{$\sharp$}} & 8.86{\textsuperscript{$\sharp$}} & 7.80{\textsuperscript{$\sharp$}} &  & \nuh \\
\rowcolor{rowbackground}
PiCIE & 2021 & 31.48 & 14.36{\textsuperscript{$\star$}} & \nuh & \nuh &  & 69.39{\textsuperscript{$\star$}} & 23.83{\textsuperscript{$\star$}} &  & 50.67{\textsuperscript{$\sharp$}} & 11.79{\textsuperscript{$\sharp$}} & 32.70{\textsuperscript{$\sharp$}} &  & \nuh \\
Odin & 2022 & \nuh & \nuh & 53.0 & 42.3 &  & \nuh & \nuh &  & \nuh & \nuh & \nuh &  & \nuh \\
\rowcolor{rowbackground}
SegDiscover & 2022 & \nuh & \nuh & \nuh & \nuh &  & \nuh & \nuh &  & 56.53 & 14.34 & 40.11 &  & \nuh \\
\bottomrule
\end{tabular}
}

\caption{Performance comparison on COCO Dataset and its variants for segmentation tasks.  {$\dagger$}\ indicates pretraining on ImageNet with supervision. {$\star$}\ Denotes results reported by PiCIE \cite{cho2021picie}. {$\sharp$}\ Denotes results reported by SegDiscover\cite{huang2022segdiscover}}
\vspace{0.8cm} 

\label{tab:seg_comp_coco}
\end{table}

In \hyperref[tab:seg_comp_msrc]{Table~\ref{tab:seg_comp_msrc}}, we observe that the MSRC dataset was steadily employed in evaluations, from early works such as DC (\textit{Acc.} 70.5, \textit{mIoU} 45.0) and CoSand (\textit{Acc.} 54.4, \textit{mIoU} 34.0) to NPG (\textit{Acc.} 90.9, \textit{mIoU} 73.0) in 2017. However, more recent works have not been evaluated in this dataset. Note also that variants such as the 14-class and 8-class subsets have been continuously used. However, other variants stopped being employed several years ago.  

In another comparison, experiments in the Object Discovery dataset have been steadily undertaken by object segmentation methods. In \hyperref[tab:seg_comp_od]{Table~\ref{tab:seg_comp_od}}, we observe neither of both metrics Precision and mIoU, have a clear trend of improvement. Of course, latest methods such as SwgSwap have obtained higher results (\textit{Precision} 91.6 and \textit{mIoU 68.0}), but there are earlier works such as CoAttnCNN that have obtained an even higher Precision and mIoU(92.29 and 69.8, respectively). It is important to note that even when SegSwap compute the same metrics utilizing the same dataset as CoAttnCNN, they report different results. According to results reported by SegSwap, CoAttnCNN obtains a Prcision of 90.9 and mIoU of 0.68 (not shown in the table). In situations like this, to enforce a fair comparison the code of each method should be analyzed in order to verify the actual metric computations. However, this is not always possible and depend on code availability.

The comparison of methods evaluated on the iCoseg dataset, presented in \hyperref[tab:seg_comp_icoseg]{Table~\ref{tab:seg_comp_icoseg}}, reveal that alternative metrics for evaluating the co-segmentation task, such as the F-measure, are not commonly employed. In the last decade, most of the methods focused on evaluating performance using mIoU and Precision. In addition, although earlier methods like DC \cite{joulin2010discriminative} and CoSand \cite{gunheekim2011distributed} started tackling the Co-Segmentation task over a decade ago, this task has kept driving attention in this line of research, as evidenced by recent works such as UnsupOD\cite{zhao2020unsupervised} and CS-OHS\cite{shoitan2021unsupervised}.

Finally, a comparison of methods is presented in \hyperref[tab:seg_comp_pascal]{Table~\ref{tab:seg_comp_pascal}}, separated by the subtasks co-segmentation and semantic segmentation. The former essentially seeks to segment out a common object whose class is shared across the dataset. Most of these methods evaluate performance based on Precision and mIoU on PASCAL VOC 2010, while only a few methods evaluate the PASCAL-S subset. It is clear to see that, in recent years, the focus of segmentation methods has started to switch to the semantic segmentation setting. \newline

\begin{figure}[htpb]
    \centering
   
    \begin{minipage}[t]{0.55\textwidth}
        \centering
        \begin{table}[H]
\centering


\resizebox{\columnwidth}{!}{%

\begin{tabular}{lccccccc} 
\toprule
\textbf{ ~ } & \textbf{ ~ } & \multicolumn{2}{c}{{\cellcolor{colorseg4}}\shortstack{Subset \\ 14-class}} & \multicolumn{1}{l}{} & \multicolumn{1}{c}{{\cellcolor{colorseg4}}\shortstack{Subset \\ 8-class}} & \multicolumn{1}{l}{} & \multicolumn{1}{c}{{\cellcolor{colorseg4}}\shortstack{MSRC (7 cls) \\ \& W. Horses}} \\ 
\cmidrule{3-4}\cmidrule{6-6}\cmidrule{8-8} Method & Year & $Acc.$ & $mIoU$ &  & $mIoU$ &  & $Acc.$ \\ 
\midrule
DC & 2010 & 70.5{\textsuperscript{$\dagger$}}  & 45.0{\textsuperscript{$\dagger$}}  &  & 43.7{\textsuperscript{$\sharp$}} &  & 78.78 \\
\rowcolor{rowbackground}
CoSand & 2011 & 54.4{\textsuperscript{$\dagger$}}  & 34.0{\textsuperscript{$\dagger$}}  &  & 37.15{\textsuperscript{$\sharp$}} &  & \nuh \\
IFBM & 2012 & \nuh & \nuh &  & \nuh &  & 70.76 \\
\rowcolor{rowbackground}
MCC & 2012 & 73.3{\textsuperscript{$\dagger$}}  & 51.0{\textsuperscript{$\dagger$}}  &  & 49.65{\textsuperscript{$\sharp$}} &  & \nuh \\
COMP & 2013 & 89.2 & 73.0 &  & 55.53{\textsuperscript{$\star$}} &  & \nuh \\
\rowcolor{rowbackground}
CST & 2013 & \nuh & 63.0 &  & 64.59{\textsuperscript{$\star$}} &  & \nuh \\
OD & 2013 & 87.7{\textsuperscript{$\dagger$}}  & 68.0{\textsuperscript{$\dagger$}}  &  & \nuh &  & \nuh \\
\rowcolor{rowbackground}
GMS & 2014 & 88.4 & 70.0 &  & 65.54{\textsuperscript{$\star$}} &  & \nuh \\
MRW & 2015 & 84.4{\textsuperscript{$\dagger$}}  & 60.0{\textsuperscript{$\dagger$}}  &  & 63.04{\textsuperscript{$\star$}} &  & \nuh \\
\rowcolor{rowbackground}
INCT & 2016 & \nuh & \nuh &  & 63.25 &  & \nuh \\
SCF & 2016 & 88.7 & 71.0 &  & 63.37{\textsuperscript{$\star$}} &  & \nuh \\
\rowcolor{rowbackground}
NPG & 2017 & 90.9 & 73.0 &  & 74.0 &  & \nuh \\
UnsupOD & 2020 & \nuh & \nuh &  & \nuh &  & \nuh \\
\rowcolor{rowbackground}
CS-OHS & 2021 & \nuh & \nuh &  & 61.26 &  & \nuh \\
\bottomrule
\end{tabular}
}

\caption{Performance comparison on different variants of the MSRC dataset in segmentation tasks. MSRC Subset with 14 classes proposed in experiments undertaken by MCC\cite{joulin2012multiclass}. MSRC Subset with 8 classes proposed in experiments undertaken by INCT\cite{li2016unsupervised}. It is also important to note that, certain methods such as DC\cite{joulin2010discriminative} and IFBM\cite{rubio2012unsupervised} combined the Weizman Horses dataset with a subset of the MSRC datasets with 7 classes.  
{
{$\dagger$}\ Indicates results reported by NPG\cite{wang2017multiple}. {$\star$}\ Denotes results reported by CS-OHS\cite{shoitan2021unsupervised}. {$\sharp$}\ Indicates results reported by INCT\cite{li2016unsupervised}.}
}

\label{tab:seg_comp_msrc}
\end{table}
    \end{minipage}\hfill
    \begin{minipage}[t]{0.37\textwidth}
        \centering
        \begin{table}[H]
\centering


\resizebox{0.9\columnwidth}{!}{%

\begin{tabular}{lccc} 
\toprule
~ & \multicolumn{1}{l}{~} & \multicolumn{2}{c}{{\cellcolor{colorseg5}}\shortstack{Object \\ Discovery}} \\ 
\cmidrule{3-4}
Method & Year & $Precision$ & $mIoU$ \\ 
\midrule
DC & 2010 & 57.26{\textsuperscript{$\dagger$}} & 27.6{\textsuperscript{$\dagger$}} \\
\rowcolor{rowbackground}
CoSand & 2011 & 74.72{\textsuperscript{$\dagger$}} & 4.8{\textsuperscript{$\dagger$}} \\
IFBM & 2012 & \nuh & \nuh \\
\rowcolor{rowbackground}
MCC & 2012 & 56.97{\textsuperscript{$\dagger$}}~ & 25.5{\textsuperscript{$\dagger$}} \\
COMP & 2013 & \nuh & \nuh \\
\rowcolor{rowbackground}
CST & 2013 & \nuh & \nuh \\
OD & 2013 & 85.41 & 57.3 \\
\rowcolor{rowbackground}
GMS & 2014 & \nuh & \nuh \\
CS-ECEM & 2014 & \nuh & \nuh \\
\rowcolor{rowbackground}
DMFC & 2015 & 76.07{\textsuperscript{$\sharp$}} & 33.0{\textsuperscript{$\sharp$}} \\
MRW & 2015 & 62.53{\textsuperscript{$\sharp$}} & 39.2{\textsuperscript{$\sharp$}} \\
\rowcolor{rowbackground}
SCF & 2016 & 88.93{\textsuperscript{$\sharp$}} & 64.3{\textsuperscript{$\sharp$}} \\
CS-GSI & 2017 & \nuh & \nuh \\
\rowcolor{rowbackground}
NPG & 2017 & \nuh & \nuh \\
SGC3 & 2017 & 83.43 & 54.8 \\
\rowcolor{rowbackground}
CoAttnCNN & 2018 & 92.29 & 69.8 \\
DFF & 2018 & \nuh & \nuh \\
\rowcolor{rowbackground}
UnsupOD & 2020 & 88.50 & 62.2 \\
CS-OHS & 2021 & \nuh & \nuh \\
\rowcolor{rowbackground}
SegSwap & 2022 & 91.6 & 68.0 \\
\bottomrule
\end{tabular}

}
\caption{Performance comparison on the Object Discovery dataset (also known as \textit{Internet Dataset}) in segmentation tasks.
\small{{$\dagger$}\ Indicates results reported by OD\cite{rubinstein2013unsupervised}. {$\sharp$}\ Denotes results reported by CoAttnCNN\cite{hsu2018coattention}.}
}
\label{tab:seg_comp_od}
\end{table}
    \end{minipage}
    
\end{figure}

\begin{figure}
    \centering
    \begin{minipage}[t]{\textwidth}
        \centering
        \begin{table}[H]
\centering
\resizebox{0.55\columnwidth}{!}{%

\begin{tabular}{lccccccc} 
\toprule
~ & \multicolumn{1}{l}{~} & \multicolumn{4}{c}{{\cellcolor{colorseg2}}\shortstack{iCoseg \\ All}}
 & \multicolumn{1}{l}{} & {\cellcolor{colorseg2}}\shortstack{Subset \\ 26-class} \\ 
\cmidrule{3-6}\cmidrule{8-8}
 Method & Year & $Pr$ & $mIoU$ & $WF_\beta$ & $F_\beta$~ &  & \shortstack{$Error$} \\ 
\midrule
DC & 2010 & 61.0{\textsuperscript{$\dagger$}}  & 39.0{\textsuperscript{$\dagger$}}  & \nuh & \nuh &  & 37.11{\textsuperscript{$\ddagger$}} \\
\rowcolor{rowbackground}
CoSand & 2011 & 66.6{\textsuperscript{$\dagger$}}  & 38.0{\textsuperscript{$\dagger$}}  & \nuh & \nuh &  & 21.82{\textsuperscript{$\ddagger$}} \\
IFBM & 2012 & 75.9 & ~ & \nuh & \nuh &  & \nuh \\
\rowcolor{rowbackground}
MCC & 2012 & 70.2{\textsuperscript{$\dagger$}}  & 43.0{\textsuperscript{$\dagger$}}  & \nuh & \nuh &  & 28.72{\textsuperscript{$\ddagger$}} \\
COMP & 2013 & 92.8 & 73.0 & \nuh & \nuh &  & \nuh \\
\rowcolor{rowbackground}
CST & 2013 & 89.5 & 48.9{\textsuperscript{$\star$}} & \nuh & \nuh &  & \nuh \\
HS & 2013 & \nuh & \nuh & 53.6{\textsuperscript{$\sharp$}} & \nuh &  & \nuh \\
\rowcolor{rowbackground}
OD & 2013 & 89.9{\textsuperscript{$\dagger$}}  & 69.0{\textsuperscript{$\dagger$}}  & \nuh & \nuh &  & \nuh \\
CS-ECEM & 2014 & \nuh & \nuh & \nuh & \nuh &  & \nuh \\
\rowcolor{rowbackground}
GMS & 2014 & 91.6 & 72.0 & \nuh & \nuh &  & \nuh \\
RC & 2014 & \nuh & \nuh & 39.5{\textsuperscript{$\sharp$}} & \nuh &  & \nuh \\
\rowcolor{rowbackground}
SMD & 2014 & \nuh & \nuh & 61.1 & \nuh &  & \nuh \\
DMFC & 2015 & 90.0 & 64.2 & \nuh & \nuh &  & \nuh \\
\rowcolor{rowbackground}
MRW & 2015 & 91.2 & 70.0 & \nuh & \nuh &  & \nuh \\
INCT & 2016 & \nuh & \nuh & \nuh & \nuh &  & 7.775 \\
\rowcolor{rowbackground}
SCF & 2016 & 91.9 & 72.0 & \nuh & \nuh &  & \nuh \\
CS-GSI & 2017 & \nuh & 70.6 & 67.0 & 82.1 &  & \nuh \\
\rowcolor{rowbackground}
NPG & 2017 & 93.8 & 77.0 & \nuh & \nuh &  & \nuh \\
SGC3 & 2017 & 90.8 & 70.4 & \nuh & \nuh &  & \nuh \\
\rowcolor{rowbackground}
CoAttnCNN & 2018 & 96.5 & 84.0 & \nuh & \nuh &  & \nuh \\
DFF & 2018 & \nuh & \nuh & \nuh & \nuh &  & \nuh \\
\rowcolor{rowbackground}
UnsupOD & 2020 & \nuh & \nuh & \nuh & \nuh &  & \nuh \\
CS-OHS & 2021 & \nuh & 75.36 & \nuh & \nuh &  & \nuh \\
\bottomrule

\end{tabular}
}

\caption{
Performance comparison on iCoseg in segmentation tasks. iCoseg subset with 26 classes proposed by INCT\cite{li2016unsupervised}. Precision is abbreviated $Pr$. Weighted $F$ measure is abbreviated $WF_\beta$. $Error$ refers to Error Rate (\%).
{
{$\dagger$}\ Indicates results reported by NPG\cite{wang2017multiple}. {$\star$}\ Denotes results reported by CS-OHS\cite{shoitan2021unsupervised}.  {$\ddagger$}\ Indicates results reported by INCT\cite{li2016unsupervised} (lower is better). {$\sharp$}\ Denotes results reported by SMD\cite{peng2017salient}. $F_\beta$ evaluated with $\beta^2=0.3$.
}
}

\label{tab:seg_comp_icoseg}
\end{table}
    \end{minipage}
    \begin{minipage}[t]{\textwidth}
        \centering
        \begin{table}[H]
\centering

\footnotesize{

\begin{tabular}{lcccccccc} 
\toprule
\textbf{ ~ } & \textbf{ ~ } & \multicolumn{2}{c}{{\cellcolor{colorseg6}}\shortstack{VOC \\ 2010}} & \multicolumn{1}{l}{} & \multicolumn{1}{c}{{\cellcolor{colorseg6}}\shortstack{VOC \\ 2012}} & \multicolumn{1}{c}{} & \multicolumn{2}{c}{{\cellcolor{colorseg6}}\shortstack{ PASCAL-S}} \\ 
\cmidrule{3-4}\cmidrule{6-6}\cmidrule{8-9}
Method & Year & $Pr$ & $mIoU$ &  & $mIoU$ &  & $F_\beta$ & $MAE$ \\ 
\midrule
\multicolumn{9}{c}{\textit{Co-Segmentation Setting}} \\ 
\midrule
DC & 2010 & 59.5{\textsuperscript{$\S$}} & 23.0{\textsuperscript{$\S$}} &  & \nuh &  & \nuh & \nuh \\
\rowcolor{rowbackground}
CoSand & 2011 & \nuh & \nuh &  & \nuh &  & \nuh & \nuh \\
COMP & 2013 & 84.0 & 46.0 &  & \nuh &  & \nuh & \nuh \\
\rowcolor{rowbackground}
\rowcolor{rowbackground}
wCTR & 2014 & \nuh & \nuh &  & \nuh &  & 65.81{\textsuperscript{$\star$}} & 0.2418{\textsuperscript{$\star$}} \\
DMFC & 2015 & 82.4{\textsuperscript{$\sharp$}} & 29.0{\textsuperscript{$\sharp$}} &  & \nuh &  & \nuh & \nuh \\
\rowcolor{rowbackground}
MRW & 2015 & 69.8{\textsuperscript{$\sharp$}} & 33.0{\textsuperscript{$\sharp$}} &  & \nuh &  & \nuh & \nuh \\
SCF & 2016 & 85.2{\textsuperscript{$\sharp$}} & 45.0{\textsuperscript{$\sharp$}} &  & \nuh &  & \nuh & \nuh \\
\rowcolor{rowbackground}
NPG & 2017 & 84.3 & 52.2 &  & \nuh &  & \nuh & \nuh \\
SBF & 2017 & \nuh & \nuh &  & \nuh &  & 77.80{\textsuperscript{$\star$}} & 0.1669{\textsuperscript{$\star$}} \\
\rowcolor{rowbackground}
CoAttnCNN & 2018 & 91.0 & 60.0 &  & \nuh &  & \nuh & \nuh \\
DFF & 2018 & \nuh & \nuh &  & \nuh &  & \nuh & \nuh \\
\rowcolor{rowbackground}
USDNL & 2018 & \nuh & \nuh &  & \nuh &  & 84.22 & 0.1391 \\
DeepCO3 & 2019 & \nuh & \nuh &  & \nuh &  & \nuh & \nuh \\ 
\midrule
\multicolumn{9}{c}{\textit{Semantic Segmentation Setting}} \\ 
\midrule
IIC-seg & 2019 & \nuh & 9.80{\textsuperscript{$\ddagger$}} &  & \nuh &  & \nuh & \nuh \\
\rowcolor{rowbackground}
MaskContrast{\textsuperscript{$\dagger$}} & 2021 & \nuh & 35.0{\textsuperscript{$\ddagger$}} &  & \nuh &  & \nuh & \nuh \\
DSM & 2022 & \nuh & \nuh &  & 37.2 ± 3.8 &  & \nuh & \nuh \\
\bottomrule
\end{tabular}
}

\caption{Performance comparison on PASCAL datasets in segmentation tasks. In the co-segmentation setting, methods are run on each class separately and metrics are averaged for the entire dataset. In the semantic segmentation setting, methods are run in the entire dataset. Precision is abbreviated $Pr$. {{$\ddagger$}\ Indicates results reported by MaskContrast\cite{melas-kyriazi2022deep}. {$\star$}\ Denotes results reported by USDNL\cite{zhang2018deep}. {$\ddagger$}\ Denotes initialized with supervision. {$\sharp$}\ Denotes results reported by CoAttnCNN\cite{hsu2018coattention}. {$\S$}\ Indicates results reported by COMP\cite{faktor2013cosegmentation}. $F_\beta$ evaluated with $\beta^2=0.3$.
}}

\label{tab:seg_comp_pascal}
\end{table}
    \end{minipage}
\end{figure}

\clearpage
\subsubsection{Comparison of Decomposition Methods}
The following tables present diverse comparisons among the main object-centric learning methods, which correspond to the class of decomposition methods of this survey. We recall that this task is also known as scene decomposition and its goal is to produce segmentation masks that correspond to objects. These comparisons include results of evaluations reported by various comparable methods. The datasets included are CLEVR and Multi-dSprites, and the evaluation metrics are ARI and mIoU. We chose a table disposition where a cell represents the results reported by the method in the row for the method in the column. This disposition allows for a better understanding of the comparison, as evaluation setups might differ on the selected metric and/or dataset.
\begin{table}[H]
\centering

\begin{threeparttable}
\resizebox{\columnwidth}{!}{%

\begin{tabular}{lllccccccc} 
\toprule
 &  &  & \multicolumn{7}{c}{{\cellcolor{colordec1}}Compared Methods in CLEVR} \\ 
\cmidrule{4-10}
Method & Dataset
variant & Metric & MONet & IODINE & Slot Attn. & DTI-Sprites & EfficientMORL & MulMON & uORF \\ 
\midrule
MONet & ~ & ~ & - & - & - & - & - & - & - \\
\arrayrulecolor{graydivider}\hline
IODINE & CLEVR6 & ARI-FG & 96.2 ± 0.6 & 98.8 ± 0.0 & - & - & - & - & - \\
\arrayrulecolor{graydivider}\hline
Slot Attn. & CLEVR6 & ARI & 96.2 ± 0.6 & 98.8 ± 0.0 & 98.8 ± 0.3 & - & - & - & - \\
\arrayrulecolor{graydivider}\hline
\multirow{2}{*}{DTI-Sprites} & \multirow{2}{*}{CLEVR6} & ARI-FG  & 96.2 ± 0.6 & 98.8 ± 0.0 & 98.8 ± 0.3 & 97.2 ± 0.2  & - & - & - \\
 &  & ARI & - & - & - & 90.7 ± 0.1 & - & - & - \\
\arrayrulecolor{graydivider}\hline
EfficientMORL & CLEVR6 & ARI & 96.2 ± 0.6 & 98.8 ± 0.0 & 98.8 ± 0.3 & - & 96.2 ± 1.6 & - & - \\
\arrayrulecolor{graydivider}\hline
\multirow{2}{*}{MulMON} & CLE-MV & mIoU & - & 18.91 ± 0.00  & - & - & - & 78.52 ± 0.08  & - \\
 & CLE-Aug & mIoU & - & 51.37 ± 0.07 & - & - & - & 70.76 ± 0.04 & - \\
 \arrayrulecolor{graydivider}\hline
\multirow{2}{*}{uORF} & \multirow{2}{*}{CLEVR-567} & ARI-FG  & - & - & 93.2 ± 1.5  & - & - & - & 87.4±0.8  \\
 &  & ARI & - & - & 3.5 ± 0.7 & - & - & - & 86.3±0.1 \\
\bottomrule
\end{tabular}

}
\caption{Performance comparison on the CLEVR dataset and some of its variants for the scene decomposition task. FG-ARI excludes the background label.}
\label{tab:comp_clevr}
\end{threeparttable}
\end{table}

As shown in \hyperref[tab:comp_clevr]{Table~\ref{tab:comp_clevr}}, these seemingly comparable methods---they all seek to produce object segmentation masks by deriving them from learned object representations---have slightly different evaluation setups. In this table, each row represents the experimental results from the corresponding method as compared to the competing methods shown as columns. As seen in the table, MONet has not reported results on these metrics since its evaluation is mostly qualitative. In contrast, IODINE reports an ARI-FG of 98.8 on CLEVR6, as compared to MONet that obtained 96.2 in the same evaluation. SLOT ATTN matches IODINE's ARI-FG on CLEVR6 with a higher standard deviation. DTI-Sprites, reported an ARI-FG of 97.2 and a lower ARI of 90.7, highlighting the significance od adjusting ARI for foregrouns pixels. EfficientMORL obtanied an ARI of 96.2. MulMON's performance on multi-view (CLE-MV) and augmented (CLE-Aug) versions of CLEVR, measured by mIoU, further reveals an improvement from 18.91 (IODINE) to 78.52 in CLE-MV and 70.76 on the augmented datasets while the baseline IODINE obtained 51.37. As well, uORF a somewhat distinct methodology based on radiance fields shows comparable performance with an ARI-FG of 87.4 and ARI of 86.3 on CLEVR-567, revelaing difficulties of SLOT ATTN when evaluating ARI with the same settings.

\begin{table}[H]
\centering

\begin{threeparttable}
\resizebox{\columnwidth}{!}{%

\begin{tabular}{lllccccccc} 
\toprule
 &  &  & \multicolumn{7}{c}{{\cellcolor{colordec2}}Compared Methods in Multi-dSprites} \\ 
\cmidrule{4-10}
Method & Metric &  & MONet & GENESIS & IODINE & LMIO & Slot Attn. & DTI-Sprites & EfficientMORL \\ 
\midrule
MONet & ~ &  & - & - & - & - & - & - & - \\
\arrayrulecolor{graydivider}\hline
GENESIS & FID &  & 92.7 & 24.9 & - & - & - & - & - \\
\arrayrulecolor{graydivider}\hline
IODINE & ARI-FG &  & 90.4 ± 0.8 & - & 76.7 ± 5.6 & - & - & - & - \\
\arrayrulecolor{graydivider}\hline
LMIO & mIoU &  & 84.0 ± 0.64 & - & - & 92.0 ± 0.66 & - & - & - \\
\arrayrulecolor{graydivider}\hline
Slot Attn. & ARI-FG &  & 90.4 ± 0.8 & - & 76.7 ± 5.6 & - & 91.3 ± 0.3 & - & - \\
\arrayrulecolor{graydivider}\hline
\multirow{2}{*}{DTI-Sprites} & ARI-FG  &  & 90.4 ± 0.8 & - & 76.7 ± 5.6 & - & 91.3 ± 0.3 & 92.5 ± 0.3  & - \\
 & ARI &  & - & - & - & - & - & 95.1 ± 0.1 & - \\
\arrayrulecolor{graydivider}\hline
EfficientMORL & ARI-FG &  & 90.4 ± 0.8 & - & 76.7 ± 5.6 & - & 91.3 ± 0.3 & ~ - & 91.2 ± 0.4 \\
\bottomrule
\end{tabular}

}
\caption{Performance comparison on Multi-dSprites in scene decomposition tasks.}
\label{tab:comp_dsprites}
\end{threeparttable}
\end{table}

In \hyperref[tab:comp_dsprites]{Table~\ref{tab:comp_dsprites}}, we observe the performance of various object-centric learning methods evaluated on the Multi-dSprites dataset as reported by each method. Again, each row corresponds to experimental results reported by a method and competing methods are shown in columns. In terms of FID, GENESIS demonstrates a notably low score of 24.9 compared to MONet’s 92.7, indicating a significant improvement in the quality of its generated images. SLOT ATTN and DTI-Sprites show strong performance in ARI-FG, both achieving around 91.3, higher than IODINE’s 76.7, suggesting better object segmentation quality. It is also notable that DTI-Sprites is the only method to report the standard ARI (not adjusted for foreground) where it obtains a score of 95.1. On the other hand, LMIO presents results in evaluating mIoU, obtaining an score of 92.0, compared to a mIoU of 84.0 obtanined by MONet in a similar setting. EfficientMORL obtains an ARI-FG of 91.2, comparable to SLOT ATTN and DTI-Sprites.

\section{Discussion}
\label{sec:discussion}
Throughout this survey, we have observed that the landscape of unsupervised object discovery is both extensive and multifaceted; and, it has certainly revealed significant developments over the years, in the quest to emulating human-like object recognition capabilities in machines. However, several critical limitations and challenges persist despite the evidenced progress, pointing to the necessity for further exploration into this field. Indeed, the motivation for creating this survey was driven from the need to provide a unified framework that would bring together the diverse approaches that have emerged within the object discovery realm. As such, this work establishes a comprehensive perspective that can guide and facilitate future advancements. While current methods have improved performance substantially in the tasks of object discovery, a relevant observation is that they often struggle with generalization across diverse unseen datasets. In fact, many methods rely on assumptions such as the prominence or centrality of objects within an image, which limits their applicability in complex, real-world scenarios. To overcome this problem, there is a latent requirement for more robust models that can generalize well across a variety of visual contexts. Moreover, although leveraging self-supervised models has demonstrated significant advantages in the unsupervised domain, it introduces a paradox: the term "unsupervised" becomes somewhat diluted when models benefit from supervised pre-training or other forms of implicit supervision. This reliance raises questions about the true independence of unsupervised methods from annotated data, suggesting that it is not yet clear what constitutes an unsupervised approach. On the other hand, object-centric structured decomposition methods have emerged as a significant advancement in scene understanding, offering more sophisticated means of disentangling complex visual environments into semantically meaningful components. However, these methods are still in their infancy and often rely on synthetic datasets or controlled environments. Expanding these approaches to work effectively in real-world settings remains a significant challenge that fosters ongoing research. As a further point, one particularly promising avenue for continuing research in this field lies in the integration of vision and language models (VLMs) within the unsupervised object discovery landscape. Vision-language models, such as CLIP, have demonstrated impressive capabilities in associating visual content with textual descriptions, even in open-vocabulary settings. These models have primarily been trained in a weakly supervised or supervised fashion, but the potential to align them in a truly unsupervised manner opens up exciting new possibilities. Indeed, if we can design an approach that aligns unsupervised vision models with unsupervised language models, we could facilitate the development of models capable of learning from vast amounts of unlabeled visual and textual data, potentially overcoming the limitations of current methods that struggle with generalization and still rely heavily on annotated data. Additionally, these systems could be more adaptable to new, unseen categories, as they would not be constrained by a fixed vocabulary. In this manner, research could yield new advancements that may be particularly valuable in applications such as autonomous driving and robotics, where encounters with novel objects are commonplace.

\section{Conclusion}
\label{sec:conclusion}
Drawing on the extensive literature reviewed in this survey, we have proposed to categorize the existing methodologies into the four primary classes of methods that have emerged: Clustering, Bounding-Box Localization, Segmentation, and Decomposition. As discussed in various sections of this study, this approach is advantageous because the methods categorized within each of these classes address the task of object discovery from distinct perspectives that align with the proposed frameworks for interpreting object discovery. On the other hand, there is a compelling need for unsupervised object discovery methods that arises from their potential to enhance and expand the range of applications across various industrial and scientific sectors. Therefore, in response to this need, diverse advancements in deep learning---such as variational autoencoders, generative adversarial networks, and transformer networks---have catalyzed the development of numerous methodologies aimed at identifying and localizing objects without explicit human guidance. Nonetheless, despite the advancements achieved, significant challenges persist, including the need for more robust generalization, the capacity to handle varied object scales and appearances, and the computational efficiency necessary for real-time applications. This survey intends to increase awareness of these topics by presenting a structured and comprehensive taxonomy that can serve as a roadmap for researchers navigating the complex landscape of unsupervised object discovery. We hope this study helps identify opportunities for further research in this area. 


\bibliographystyle{unsrt}  
\bibliography{references}

\end{document}